\documentclass[Afour,sageh,times]{sagej}

\usepackage{moreverb,url}
\usepackage[colorlinks,bookmarksopen,bookmarksnumbered,citecolor=blue,urlcolor=blue]{hyperref}
\usepackage{array}
\usepackage[caption=false,font=footnotesize]{subfig}
\usepackage{stfloats} % double column floats at the bottom of the page "\begin{figure}[!b]" 
\usepackage{url}
\usepackage{bm,amssymb,amsmath,color,algorithm,algorithmic,url}%,tabularx,subcaption,float}
\usepackage{multirow}
\usepackage{enumitem}

\newcommand\BibTeX{{\rmfamily B\kern-.05em \textsc{i\kern-.025em b}\kern-.08em
T\kern-.1667em\lower.7ex\hbox{E}\kern-.125emX}}

\def\volumeyear{2025}
\def\journalname{\mbox{The International Journal of} Robotics~Research}

\begin{document}

\runninghead{Lepora}
\title{Tactile Robotics: Past and Future}
\author{Nathan F. Lepora\affilnum{1,2}\vspace{-2em}}
\affiliation{\affilnum{1} School of Engineering Mathematics and Technology, University of Bristol, UK. \affilnum{2} Bristol Robotics Laboratory, University of Bristol, UK.}
\corrauth{Nathan Lepora, University of Bristol, UK.}
\email{n.lepora@bristol.ac.uk}

\begin{abstract}
What is the future of tactile robotics? To help define that future, this article provides a historical perspective on tactile sensing in robotics from the wealth of knowledge and expert opinion in nearly 150 reviews over almost half a century. This history is characterized by a succession of generations: 1965-79 (origins), 1980-94 (foundations and growth), 1995-2009 (tactile winter) and 2010-2024 (expansion and diversification). Recent expansion has led to diverse themes emerging of e-skins, tactile robotic hands, vision-based tactile sensing, soft/biomimetic touch, and the tactile Internet. In the next generation from 2025, tactile robotics could mature to widespread commercial use, with applications in human-like dexterity, understanding human intelligence, and telepresence impacting all robotics and AI. \textcolor{black}{By linking past expert insights to present themes, this article highlights recurring challenges in tactile robotics, showing how the field has evolved, why progress has often stalled, and which opportunities are most likely to define its future.}
\vspace{-1em}
\end{abstract}


\maketitle
\section{Introduction}\label{sec:introduction}
\begin{quote}
``Study the past if you would define the future.'' (quote by Confucius, {\em Kong Fuzi} $\sim$500\,BCE)
\end{quote}
Ever since the modern concept of a `robot' entered popular culture in the early 20th century, anticipation has steadily grown that industrial technology will culminate in machines with human-like abilities. Human dexterity, and arguably human intelligence, originated from the evolution of hands adapted to manipulating tools~(\cite{wilson_hand_1999}). For that manipulation, the senses of touch and proprioception are essential to provide the force and geometric feedback needed for fine sensorimotor control of the fingers. Therefore, robots will also need these senses to attain human-like dexterity, motivating tactile robotics: the study, development, and use of tactile sensing for robotics. 

For nearly half a century, there have been many excellent surveys of progress in tactile robotics, beginning with ``Touch-Sensing Technology: A Review''~(\cite{harmon_touch-sensing_1980}). Subsequently, new reviews were published on average every year or two until 2010, with progress over that period captured in the landmark article ``Tactile Sensing -- From Humans to Humanoids''~(\cite{dahiya_tactile_2010}). Reviewing has since proliferated, with 15 published in 2024 alone. Yet, even though some major progress has been reported, tactile robotics still does not seem to have transformed robot dexterity. Indeed, one authoritative review began with a wry comment: ``Tactile sensing always seems to be {\em a few years away} from widespread utility''~(\cite{cutkosky_force_2008}). Therefore, it seems a good time to ask again: When will tactile sensing transform robotics? 

How do you ask about the future? You can study the~past. When studying tactile robotics, the past views of experts are valuable because many of their original motivations still hold. However, people tend to devalue older knowledge, especially when the technology appears new, such as robotics, and when the pace of change seems rapid. The aim of this article is to give voice to those experts, so that today's researchers may better define the future. 

\begin{figure*}[t!]
	\centering
    \begin{tabular}{@{}c@{}}
    {\bf Bibliometric analysis of review articles on tactile robotics}\\
    \includegraphics[width=2\columnwidth,trim=80 15 90 20,clip]{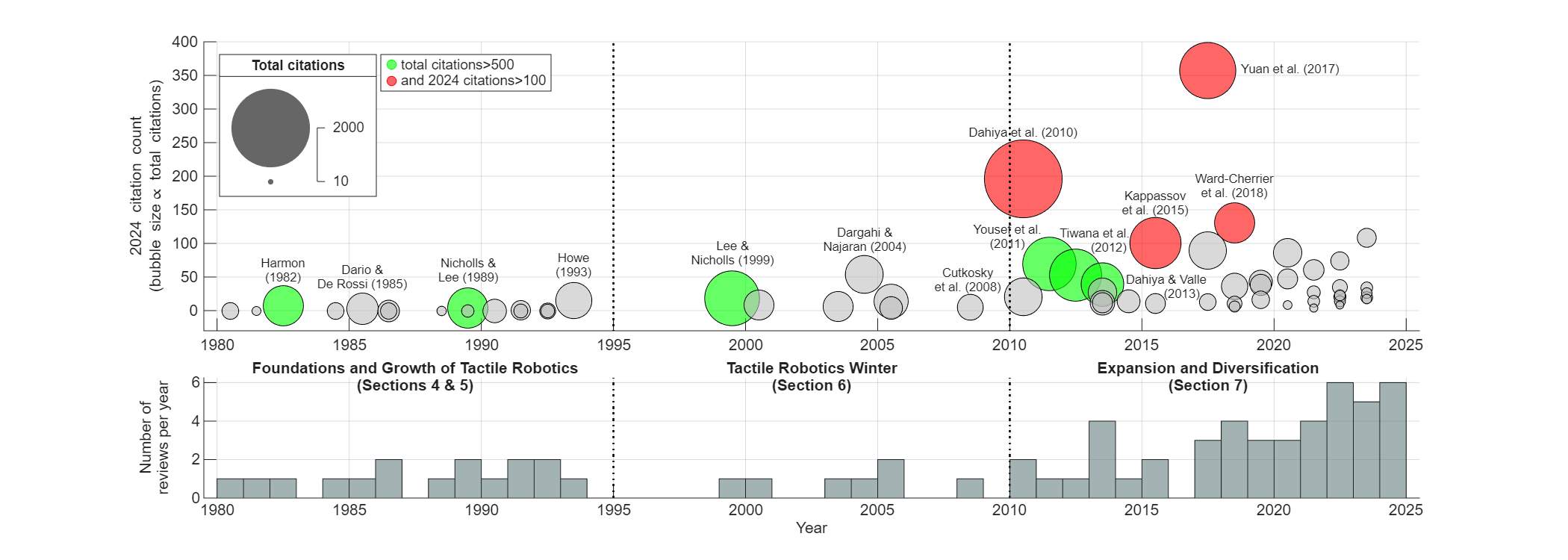} 
    \end{tabular}    
    \vspace{-1em}
	\caption{Bubble plot for the citation counts of review papers in tactile robotics (using the tabularized list of papers in Table~\ref{tab:1}). The bubble height is the number of citations in 2024 and the bubble size is proportional to the citation total up to 2024. Those with more than 500 citations are colored green, and those also with more than 100 citations in 2024 are colored red \textcolor{black}{to indicate past and present impact. The expert opinion in those papers guides this historical perspective, with author names from key papers displayed.}}
    \vspace{-.5em}
	\label{fig:0}
\end{figure*}

This article will provide a historical perspective on tactile robotics using the wealth of knowledge and expert opinion from past review articles: 69 on tactile robotics in Table~\ref{tab:1} and another 54 on e-skins in Table~\ref{tab:5} up to 2024. By visually representing their publication dates (Figure~\ref{fig:0}), one can immediately see trends in the steady growth of the number of tactile reviews from 1980 to the mid-1990s, then a drop in reviewing for about 15 years, and accelerating growth since. These changes in behavior suggest three distinct periods in the 45 years from 1980 to 2024, which we term {\em historical generations} (like societal generations, such as `gen Z'). 

The structure of this article is organized by these historical generations of tactile robotics. Section~\ref{sec:origins} considers the period 1965-1979: the {\em origins of tactile robotics}. Sections~\ref{sec:beginnings} and~\ref{sec:foundations} consider 1980-1984 then 1985-1994: the {\em foundations of a research field} and {\em growth of tactile robotics}. Section~\ref{sec:pessimism} considers 1995-2009, termed: {\em tactile robotics winter}. Section~\ref{sec:renewal} considers 2010-2024: {\em expansion and diversification}. This period separates into five distinct themes covered in Sections~\ref{sec:divergence}-\ref{sec:towards} on e-skins, tactile robotic hands, vision-based tactile sensing, soft/biomimetic technologies, and the tactile Internet. Lastly, Section~\ref{sec:future} considers the {\em future of tactile robotics} in the next generation from 2025 onward. Initially, these five distinct themes may continue, then disruptive technological change affecting all robotics and AI is expected. In this, the original motivations retain their immediacy today: to enable human-like dexterity, better understand human intelligence, and implement telepresence.

\section{Methodology}
\label{sec:comments}

The tactile robotics research surveyed here was found with a review article search in Google Scholar, followed by a manual search of their bibliographies. Given the very large number of papers now being published in tactile robotics, it would be unfeasible for a human to collate information across the entire field by surveying individual research papers. Any such review could cover only a narrow area within the authors' own interests. For that reason, reviews are treated here as the {\em only} source of information in this article. The exception is before the first review in 1980, when original articles and technical reports were used. However, academic publishing was very different in that era, with relatively few articles, many of which are difficult to find.

%The disadvantage of this approach is that only research findings in reviews are surveyed. Therefore, research that has not yet entered review articles will not be covered, and some of this could be influential. In the author's view, this limitation is preferable to introducing bias from selectively including some research papers. It also has the benefit of revealing gaps in the review literature that others can fill.  

The details of the reviews were collated into tables with summary information such as journals, titles, and overall or recent citations (Appendix: Tables~\ref{tab:1}-\ref{tab:7}). This information is shown diagrammatically in Figure~\ref{fig:0} (tactile robotics) \textcolor{black}{and later in Figure~\ref{fig:X} (e-skins) to indicate which have been influential and whose expert opinion should guide this historical perspective. Accordingly, author names of key articles are displayed on the plots, along with selected articles at important times in the evolution of tactile robotics that were close to the criteria we used to identify key articles.} Several other trends became evident from Figure~\ref{fig:0} that also helped guide the structure of this article.

Most prominently, there appear to be three distinct generations of tactile robotics research: an initial sustained period from 1980-1994, then a drop in output over 1995-2009, and rapid growth since 2010. It is interesting that these periods seem analogous to the generations of social change (`gen X', `millenials', `gen Z') that also occur every 15 years. In the author's opinion, this similarity reflects that research is, of course, done by people whose attitudes to research are influenced by their society. In addition, technologies such as ubiquitous access to information on the Internet can underlie changes in both society and research culture. 

Two other trends stood out when considering all tactile robotics reviews (see Figure~\ref{fig:0}). First, a tendency for early (pre-1995) reviews to now be forgotten, and second, the proliferation of reviewing since about 2018. These trends were accompanied by a change in who was writing the reviews. The earliest reviews tended to be written by recognized experts who presented their views with authority. This gradually shifted to students writing with an experienced supervisor, then more recently to new teams entering the research area. %Likely, this reflects a change in the priorities for writing a review: originally, an individual had gained enough experience to form a perspective on the field that they felt valuable to share; now, there is pressure on early career researchers to gather citations, and new surveys of topical subjects tend to be cited far more than research papers.

The problem with these trends is that it is now very difficult to gain a complete perspective over the field in a way that was possible in the past. This can lead to important motivations for progressing tactile robotics being forgotten and overcapacity in popular areas of research, while fewer researchers work on the topics that may actually drive progress. By giving voice to past experts and placing those voices within both a historical and a modern context, this article aims to help improve those imbalances as the field enters a new generation of tactile robotics.

\section{1965-79: Origins of Tactile Robotics}
\label{sec:origins}

Like modern robotics that has many origin stories~(\cite{mason_creation_2012}), so does tactile robotics. One origin story is shared with robotics: the development of teleoperator systems that began in the late 1940s, which led to servo-driven telemanipulators and the first tactile sensors to relay a sense of touch. Another origin is within 1960's AI, where early pioneers such as Marvin Minsky recognized the importance of human dexterity for understanding intelligence but faced insurmountable challenges from the technology of that time. The motivations for these early attempts remain insightful today and give a context for later research in tactile robotics. 

\begin{figure}[t!]
	\centering
	\includegraphics[width=0.47\columnwidth,trim=0 0 0 0,clip]{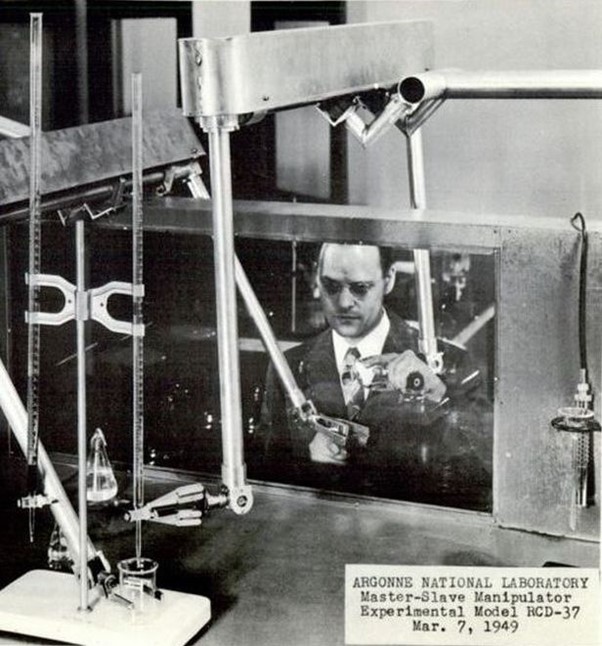}
	\includegraphics[width=0.51\columnwidth,trim=40 80 40 10,clip]{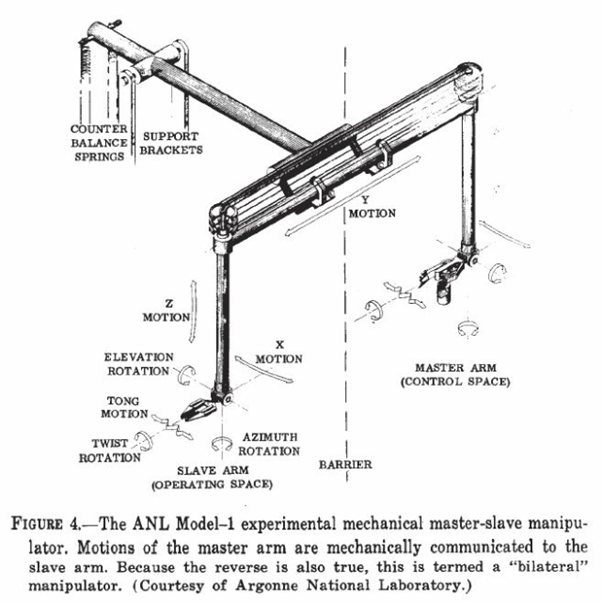}
    \vspace{-.5em}
	\caption{Left: Raymond Goertz demonstrating his mechanical telemanipulator. Right: Design of his ANL Model-1 manipulator: motions of the controlling and operating arms follow each other bilaterally. (Images from  \href{https://en.wikipedia.org/wiki/Raymond_Goertz\#/media/File:Apf1-06395t.jpg}{University of Chicago Library}.)}
    \vspace{-1em} 
	\label{fig:1}
\end{figure}

The Atomic Age following World War II created a need for teleoperator-controlled systems to handle radioactive materials. These early telemanipulators were purely mechananical, beginning with Raymond Goertz's Model-1 Manipulator (1948) from Argonne National~Laboratory, Illinois (Figure~\ref{fig:1}). A pair of mechanical arms with pinch grippers were coupled bilaterally through metal tape cabling so that a human operator could guide all 6 degrees of freedom of the gripper pose along with its closure. Due to mechanical coupling, these telemanipulators provided excellent force feedback to the operator through the cabling, and are also an early example of haptic feedback devices~(\cite{sheridan_human_1997}). Such telemanipulators achieved feats of physical dexterity that remain impressive a human lifetime later; for example, a film reel from the 1950s shows a manipulator placing a cigarette in a secretary's mouth, striking a match, and lighting the cigarette~(\cite{british_pathe_archives_atomic_nodate}). Although the choice of task has not dated well, it shows a level of dexterity that remains impressive today. 

Progress in the use of electric servomotors in the early 1950s led to telemanipulators in which the controlling and operating arms were connected electrically, rather than mechanically. The first of these was Goertz's E-1 Manipulator (1954), which was also bilateral in that any difference in pose between the controlling and operating grippers drove opposing electric motor forces on the two arms. This mechanism gave force feedback on the operator’s hand that mirrored the forces exerted on the remote gripper~(\cite{sheridan_human_1997}). Although this ingenious device initiated much of modern telerobotics and haptics, electric motor feedback could not transmit forces to the operator as effectively as direct mechanical coupling~(\cite{minsky_telepresence_1980}). Nevertheless, the advantages of electric servomotor-based systems for precision and control are the basis of almost all industrial robotics today~(\cite{gasparetto_brief_2019}).

% L: https://en.wikipedia.org/wiki/Raymond_Goertz#/media/File:Apf1-06395t.jpg
% R: https://cyberneticzoo.com/teleoperators/1954-electromechanical-manipulator-ray-goertz-american/
% CC BY-SA 4.0 license 

% example attribution: (A) Reproduced with permission from Ref. [39], ©IEEE, 2016. (B) Reproduced with permission from Ref. [41], ©IEEE, 2019. (C) Reproduced with permission from Ref. [42], ©the authors, 2019

The arrival of the space age with the 1957 launch of Sputnik and the 1958 formation of the North American Space Agency (NASA) gave a need for teleoperation over long distances. However, an issue was that the teleoperators would adopt a slow and cumbersome ``move-and-wait'' behavior to avoid instabilities in control with time delays. \cite{ferrell_supervisory_1967-2} addressed this problem by developing {\em supervisory control} where a human operator guides a semi-autonomous robotic system, later called a telerobot~(\cite{sheridan_telerobotics_1989,hokayem_bilateral_2006}). Supervisory control made robot teleoperation useful in practice, by helping resolve issues such as limited sensory feedback and enabling faster-than-human reactions to enable smooth contact. 

\begin{figure}[t!]
	\centering
    \begin{tabular}{@{}c@{}c@{}}
    \small{\bf (a) MIT optical touch sensor (1966)} & \small{\bf (b) Tactile images}\\
	\includegraphics[width=0.52\columnwidth,trim=50 45 50 10,clip]{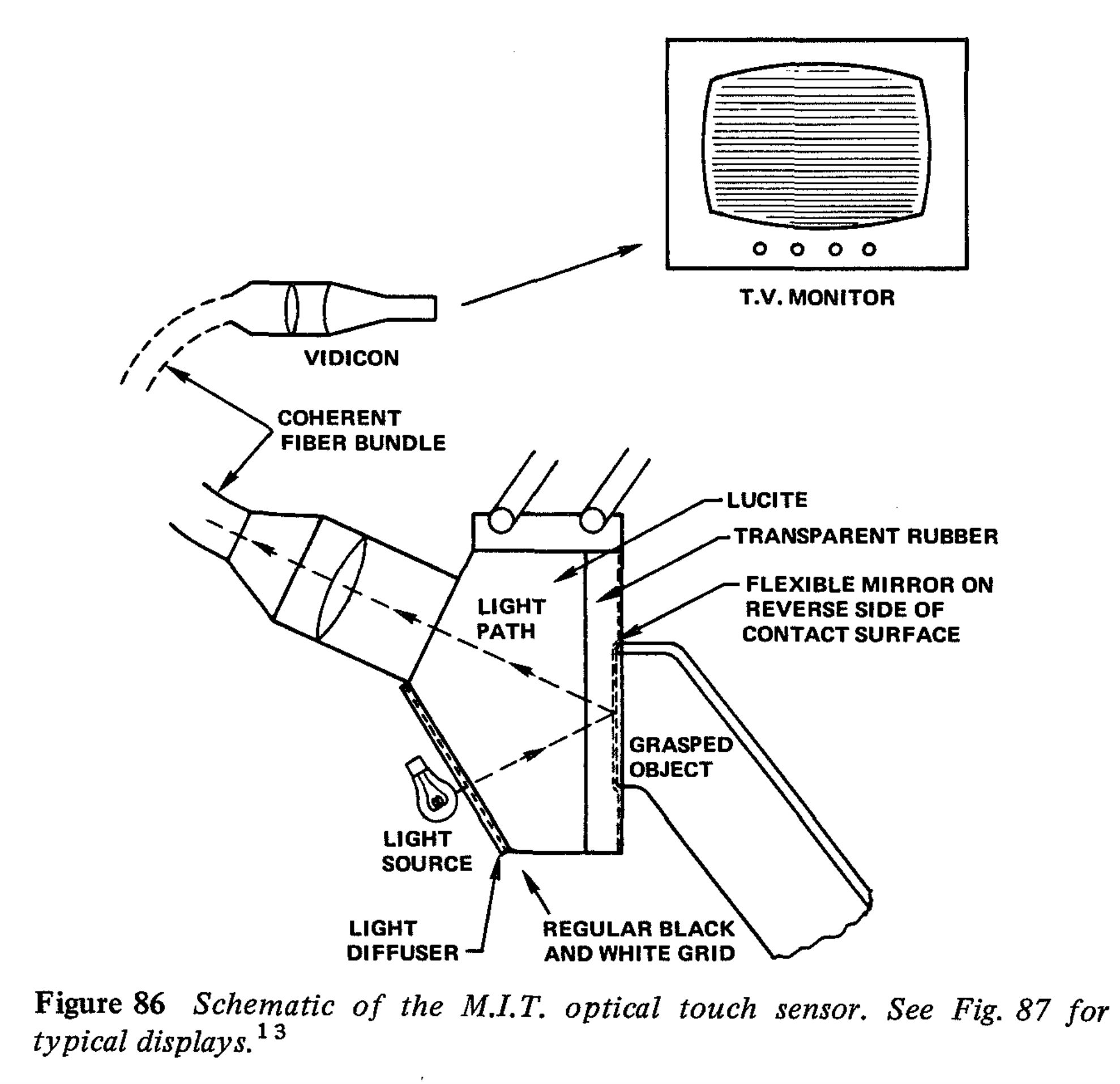} &
	\includegraphics[width=0.46\columnwidth,trim=0 0 0 0,clip]{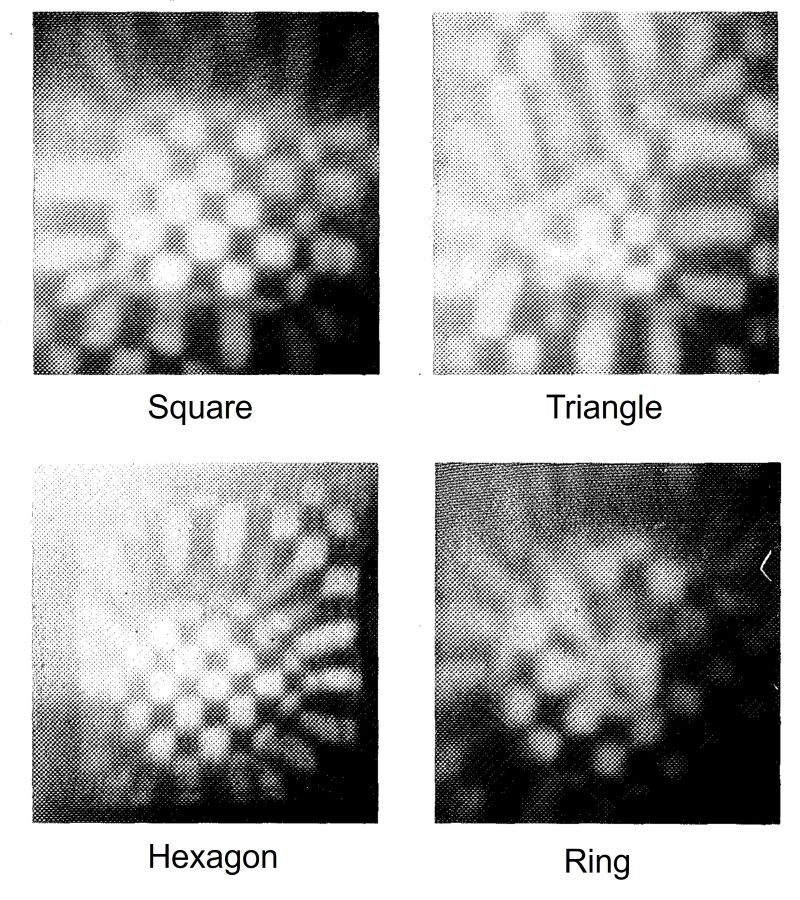} \\
    \end{tabular}
    \vspace{-.75em}
	\caption{MIT optical touch sensing system~(\cite{strickler_design_1966-1}). Left: the imprint on the tactile skin is imaged via a fiber-optic bundle viewing a chequerboard pattern on a flexible mirror. Right: tactile images of contacts with various planar shapes. The system transduces touch into a picture on a TV for a human operator. (Images from report by \cite{corliss_teleoperator_1968}.)}
    \vspace{-1em}
	\label{fig:2}
\end{figure}
% images from Corliss-Johnson TELEOPERATOR CONTROLS P132, 133
% Work of the US Gov. Public Use Permitted: https://ntrs.nasa.gov/citations/19690012116

In an attempt to relay tactile sensation during teleoperation, the first artificial tactile sensors to transmit spatial patterns of contact were invented in Thomas Sheridan's Man-Machine Systems Laboratory~(\cite{strickler_design_1966-1}). The {\em MIT optical touch sensing system} used a TV camera fitted to a fiber-optic bundle to send detailed images from the inner surface of a deformable skin to a TV monitor (Figure~\ref{fig:2}). Several skin designs were explored to visually represent the contact, {\em e.g.,} a photoelastic material~(\cite{kappl_sense_1963}) and moir\'e interference patterns~(\cite{strickler_design_1966-1,strickler_development_1966}). Thomas Strickler finalized the tactile sensor with a checkerboard pattern on a flexible mirror to give a high-contrast image of the deformation relayed to a human operator using a television monitor (example tactile images in Figure~\ref{fig:2}). However,  this early type of `teletouch' did not attract much attention~(\cite{sheridan_telerobotics_1989}) and its research was abandoned. 

\begin{figure}[t!]
	\centering
	\includegraphics[width=0.9\columnwidth,trim=0 120 0 100,clip]{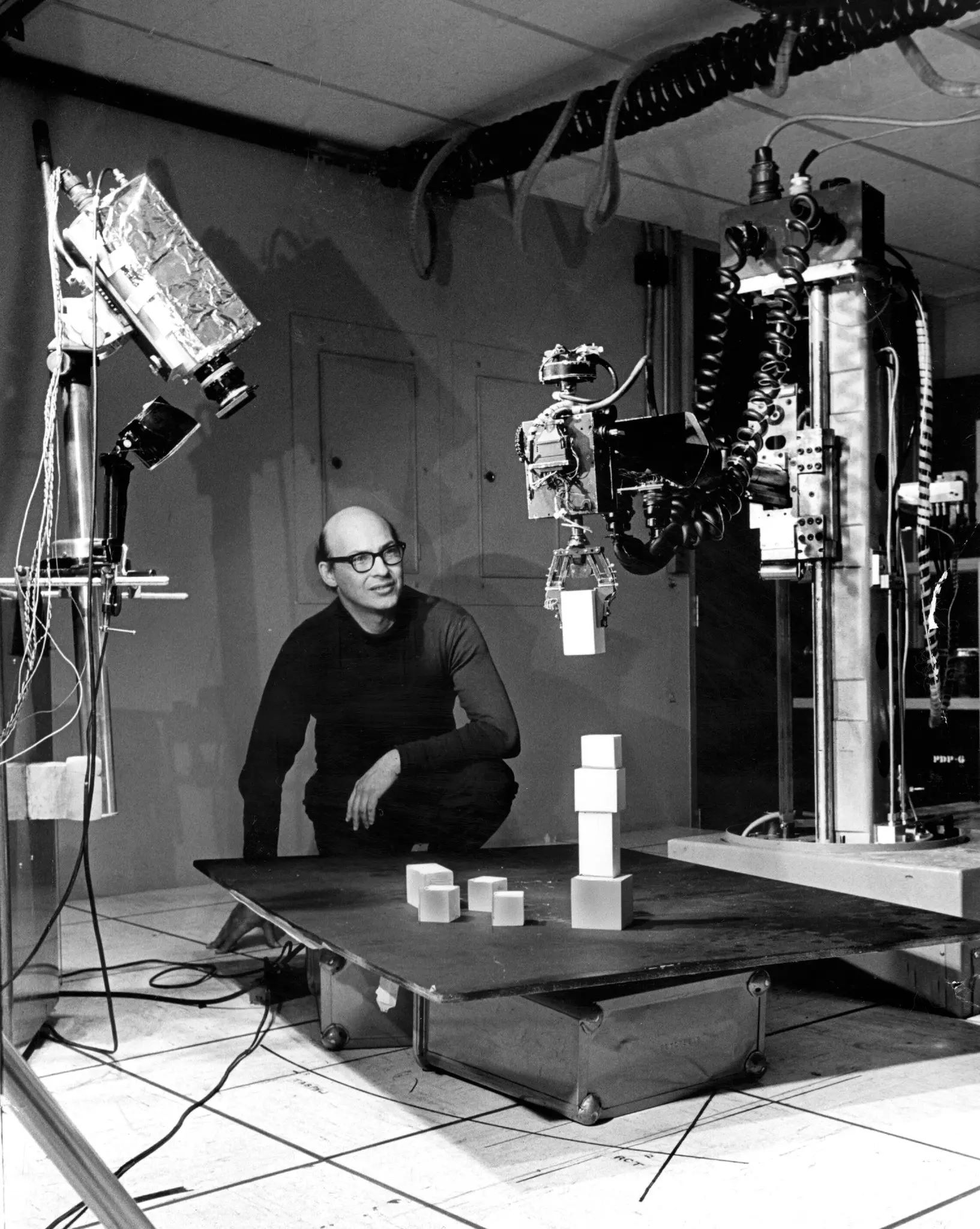}
    \vspace{-0.5em}
	\caption{Marvin Minsky with a robot arm and gripper. The robotic system was constructed with Seymour Papert in Minsky's Artificial Intelligence (AI) Laboratory in MIT. It was originally intended to instantiate a ``blocks world'' task where cubes could be stacked. (Image from the \href{https://mitmuseum.mit.edu/collections/object/GCP-00017594}{MIT Museum}.)}
    \vspace{-1em}
	\label{fig:3}
\end{figure}

The earliest investigation of tactile sensing for robotics was in Marvin Minsky's Artificial Intelligence (AI) group, also at MIT. A pioneer of AI, Minsky recognized the importance of human dexterity for human intelligence and proposed a project to develop ``techniques for machine perception, motor control and coordination that are applicable to performing real-world tasks of object recognition and manipulation'' such that ``the system needs visual and tactile input devices capable of unusual discrimination ability"~(\cite{minsky_autonomous_1966}). This project resulted in a mechanical hand and arm that Minsky wanted to autonomously stack blocks~(Figure~\ref{fig:3}). Only hints now remain of this research, such as a 1969 report that describes how a graduate student, David Waltz, ``engineered a grasping surface with good pressure sensitivity at a great many points by wrapping a coaxial cable connected to a time-domain reflectometer''~(\cite{minsky_mechanical_1969}). However, practical issues in creating a tactile robot manipulator proved insurmountable, motivating a change to purely computational studies of intelligence. 

Thus, in the early 1970s, Minsky's `blocks world' became a purely computational construct of a robot and its environment~(\cite{minsky_winograds_1971}). Nevertheless, this led to seminal research on understanding natural language~(\cite{winograd_understanding_1972}) and Minsky's acclaimed theory of natural intelligence in his book {\em The Society of Mind}~(\cite{minsky_society_1988}). In the chapter ``Easy Things Are Hard,'' Minsky  reflects that he and Seymour Papert had ``long desired to combine a mechanical hand, a television eye, and a computer into a robot that could build with children's building-blocks,'' and to do that they ``had to build a mechanical hand, equipped with sensors for pressure and touch at its fingertips''. But he said they needed to keep adding more programs in an ever-increasing universe of complications, so ``in attempting to make our robot work, we found that many everyday problems were much more complicated than the sorts of problems, puzzles and games adults considered hard.'' This observation is similar to its contemporary the {\em Moravec Paradox} that ``it is comparatively easy to make computers exhibit adult-level performance on intelligence tests or playing checkers, and difficult or impossible to give them the skills of a one-year-old when it comes to perception and mobility''~(\cite{moravec_mind_1988}).

%https://mitmuseum.mit.edu/collections/object/GCP-00017594
% I have emailed about copyright to use this image.

\begin{figure}[t]
	\centering
    \reflectbox{\includegraphics[width=\columnwidth,trim=0 0 5 0,clip]{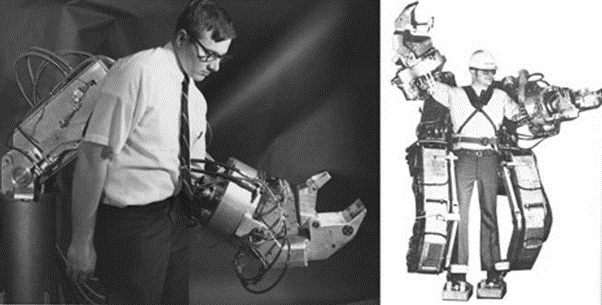}}
    \vspace{-1.5em}
	\caption{Ralph Mosher in the G. E. HARDIMAN Exoskeleton (Human Augmentation Research and Development Investigation MANipulation). Right: one arm of the exoskeleton showing the powered gripper. (Images from the \href{https://misci.org/collections-and-research/}{miSci, Museum of Innovation and Science}; reprinted in \cite{minsky_telepresence_1980}.)}
    \vspace{-.5em}
	\label{fig:4}
\end{figure}

At the end of that decade, Minsky's last contribution to the story of tactile robotics was in bringing teletouch and telemanipulation together as {\em telepresence}. He popularized this term in his so-named article in {\em Omni Magazine}~(\cite{minsky_telepresence_1980}): ``Telepresence emphasizes the importance of high‑quality sensory feedback and suggests future instruments that will feel and work so much like our own hands that we won't notice any significant difference." Minsky had a revolutionary view of seeing a transformation of energy production, manufacturing, medicine, and creating entire new industries for a telerobot-based economy. Alongside human augmentation (Figure~\ref{fig:4}), he called for investment to solve the engineering challenges, including: ``very little is known about tactile sensations. It seems quite ironic that we already have a device that can translate print into feel, but we have nothing that can translate {\em feel} into feel." Yet, despite progress in tactile robotics and haptics, Minsky's manifesto remains a compelling vision of a future that did not occur. 

It is hard to assess how much these original studies directly influenced later research in tactile robotics. Not least, most were reported in internal memos rather than in journals, and so may not have been known to many researchers. However, their importance is to show that tactile robotics was deeply intertwined with the roots of artificial intelligence and robotics a human lifetime ago. Also, the original motivations for these fields of study retain their immediacy today: to enable human-like dexterity, better understand human intelligence, and implement telepresence. The difference now is that these aims may soon be realizable given the progress over three intervening generations (45 years) of tactile robotics research.   
 % also on Cyberzoo.com)
% have emailed about reuse

 \vspace{-1em}
\section{1980-84: Foundations of a Research Field}
\label{sec:beginnings}

In the early 1980s, tactile robotics was formed into a distinctive research discipline through a series of four reviews by~\cite{harmon_touch-sensing_1980,harmon_sense_1981,harmon_automated_1982,harmon_tactile_1984}. Of these, the 1982 article on ``Automated Tactile Sensing'' was published in the first volume of the {\em International Journal of Robotics Research~(IJRR)}. As the first robotics journal, it marks a beginning of robotics as a research field~(\cite{mason_creation_2012}), with a new working definition: ``Robotics is {\em the intelligent connection of perception to action}'' (\cite{brady_artificial_1985}).

Their author, Leon Harmon, had insight into artificially recreating the human senses as a pioneer in computer vision. He was a free thinker, inspiring Salvador Dali's ``Lincoln in Dalivision'' (1977) from a mosaic image of Abraham Lincoln that Harmon published in {\em Scientific American} (he also created the ``Computer Nude'' (1966) that was famously reprinted in the {\em New York Times}). In the last few years of his life, Harmon became convinced of a need for a robotic sense of touch in manufacturing and related industries, taking inspiration from the human tactile sense. He demonstrated this insight to others with four papers: ``Touch-sensing Technology: A Review'' (\cite{harmon_touch-sensing_1980}), ``A Sense of Touch Begins to Gather Momentum''~(\cite{harmon_sense_1981}), ``Automated Tactile Sensing''~(\cite{harmon_automated_1982}), and ``Tactile Sensing for Robots''~(\cite{harmon_tactile_1984}). 

%, which Dali appears to have paid homage to by combining Lincoln with a nude portrait of his wife Gala

Harmon was motivated by his view that tactile sensing could transform industrial robotics. He sought authority from influential industry leaders of the time, such as Unimation co-founder Joseph Engelberger who ``laid heavy stress on tactile sensing'' for robot physical interactions~(\cite{harmon_sense_1981}). His views were also based on findings from teleoperation research, quoting Antal Bejczy that: ``Manipulation-related key events are not contained in visual data,'' specifically the ``geometric and dynamic reference data for the control of [pose], adaptive and soft grasping of objects, [and] gentle load transfer.'' This led Harmon to conclude that the technology of that time was too crude for robotic manipulation and that a new generation of sophistication was beginning to emerge.

%He recognized that the ``general needs for sensing in manipulator control are proximity, touch/slip, and force/torque''. 

\begin{figure}[t]
	\centering
	\includegraphics[width=0.71\columnwidth,trim=20 95 20 160,clip]{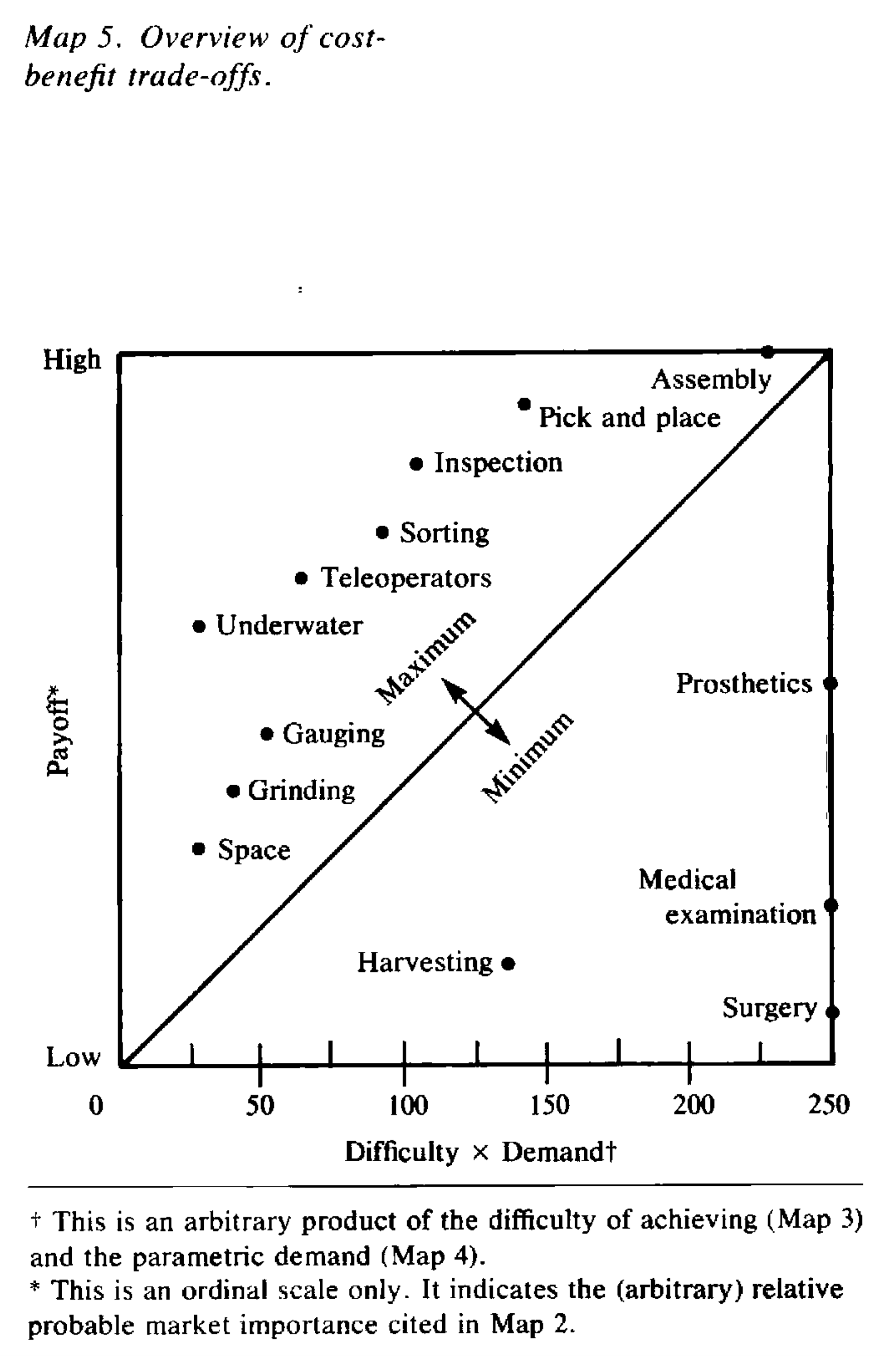}
    \vspace{-0.25em}
	\caption{Harmon's cost-benefit analysis to characterize touch-sensing commercial feasibility. The payoffs were estimated from 1979 robotics sales, if tactile sensing had been integrated, along with commercial need and technical maturity. The costs in terms of difficulty $\times$ demand were likewise estimated by scoring these aspcts of the tasks. Full details are given by \cite{harmon_automated_1982} (Figure~5 reprinted above).}
    \vspace{-1em}
	\label{fig:5}
\end{figure}

For context, industrial robotics became well-established in the 1970s with several leading companies (Unimation in the US, Hitachi in Japan, and ASEA/ABB in Europe). These were experiencing a surge in demand for advanced robotic arms from the automotive and other manufacturing industries~(\cite{gasparetto_brief_2019}), which created a demand for better interaction with the operator and the environment to perform more complex tasks. The companies responded by making robots easier to program with improved sensing of themselves and their surroundings. However, these advances were insufficient for many commercially valuable tasks (see, {\em e.g.}, Figure~\ref{fig:5}), motivating Harmon to survey artificial tactile sensing for robotics. 
% , such as using advances in force/torque (FT) sensing

Like many later reviews of the field that followed, Harmon's papers surveyed the main transduction technologies being explored for tactile sensing. Examples include piezo-resistive semiconductors, electrode arrays in conductive elastomers, piezo-diodes, ``pressistor'' compounds, arrays of piezo-electric transducers, electromagnetic, hydraulic, optical and capacitive touch sensing. He drew particular attention to arrays of solid-state transducers with attached computing elements, typified by the research of Marc Raibert at CalTech's JPL (later published in \textit{IJRR} as a ``VLSI tactile sensing computer''~(\cite{raibert_design_1982}). More than four decades later, this list remains current and many of these technologies have since been commercialized (see Table~\ref{table:2} and Table~\ref{table:X} later for lists of sensors in the 1980s and 2000s).  

%, by distinguishing between sensing contact through a point or spatially distributed across a sensing area. 
%Also, in earlier comments, \cite{harmon_sense_1981} emphasized  {\em deformation %sensitivity}, rather than force sensitivity, which could be substituted in his definition. 

\cite{harmon_automated_1982} also made the first working definition for a robot sense of touch: {``{\em Tactile sensing refers to skin-like properties, with which areas of force-sensitive surfaces are capable of reporting graded signals and parallel patterns of touching}.''} He further clarified a distinction between simple contact and tactile sensing: ``{\em When reference is made to simpler contact sensing [at one or a few points], the term simple touch will be used.}'' Although this term is no longer in use, it describes well the spatial aspects of tactile sensing. 

His clarification of simple contact remains insightful in distinguishing between force-torque and tactile sensing~(\cite{harmon_sense_1981,harmon_automated_1982}). Force-torque sensors measure contact through a point, with six axes matching the degrees of articulation of a robot arm. However, force sensing is subject to the ``unwelcome fact'' that force cannot be measured directly, but only from displacement, so its measurement is subject to errors, hysteresis, and delays from its transmission through compliant structures. Tactile measurements that relate directly to skin surface displacement do not suffer from those limitations of indirect force-torque measurements. 

Another way that Harmon sought to bring coherence to tactile robotics was to gather responses to questionnaires answered by over 50 academic, industrial, and other researchers~(\cite{harmon_automated_1982}). He asked about the need for next-generation robots, finding that ``90\% of the respondents felt strongly that tactile sensing is needed.'' Views spanned from one industry executive feeling that tactile sensing ``would open the manipulator market by perhaps an order of magnitude'' to a chief engineer who said that ``sensors are not the key element in [the near future]'' but that the need was for improved electronic control using existing technology such as strain gauges and limit switches. 

% The questions covered many topics, whose ``most realistic and imaginative'' answers are summarized and best read in Harmon's article. The robot hands envisioned most frequently were ``similar to human hands so as to permit similar dexterity and flexibility'' with two hands working cooperatively most desirable (referring also to Minsky~(\cite{minsky_telepresence_1980}). Tasks included arc-weld tracking, bin picking, adaptive grasping, batch assembly, acquisition/manipulation underwater and in space, rehabilitation/prosthetics, inspection, measurement of shape/weight/centre-of-mass, very fine assembly, handling of delicate/soft/flexible parts and electronics-component insertion such as automated wiring/cutting/crimping. Clearly, these tasks remain current (and unsolved) today. 

%One said ``it is difficult to be optimistic about agriculture as a promising domain for industrial robots, particularly in relation to the promise of more machine-oriented industrial operations such as automated assembly.'' Today, agriculture remains highly human-intensive, particularly for manual tasks such as harvesting fruits and vegetables, although there is a promise of more automation in the expanding areas of vertical farming and hydroponics. Another

Some of the respondents gave cautionary quotes that remain pertinent today. For example, one responder said ``household robotics probably would not be enthusiastically received [because]  people did not want competition from machines at home (neither did they want to be watched or followed by robots).'' To this, Harmon retorted that ``this respondent may not have been familiar with \mbox{toilet-}, window- or oven-cleaning; floor scrubbing, dish washing, snow shoveling, or diaper changing.'' The outlook was that robots would ultimately replace humans in ``grungy, hazardous, dull, repetitive tasks,'' a view that has become a cliche as the 3 Ds (dirty, dangerous, and dull) of robotics.

Harmon also surveyed the tactile sensor needs for robotic tasks and their probable market importance along with the technological developments required to achieve these functions for that decade 1980-1989. He synthesized the data as a cost-benefit analysis (reprinted in Figure~\ref{fig:5}). Conclusions included: (1) 9 applications (to the top left) seem commercially feasible, with the really big returns being in factory automation; (2) the 4 applications (bottom right) that seem unlikely to return the investment all concern living systems; and (3) if ``technological problems are solved for one particular task, those to the left of it [in difficulty] will also be solved implicitly.'' Thus, making an all-out effort to solve assembly would likely solve all the other applications.  

%, to ``assess the research and development exposure necessary before payoff.''

Why, four decades later, has none of this come to pass? There is still minimal use of tactile sensing in industrial automation, and many of these tasks remain unsolved ({\em e.g.,} assembly) or have partial solutions with force-torque sensing or computer vision ({\em e.g.,} inspection or pick and place). In part, the issue is that the fabrication of practical high-performance tactile skins has proven to be far more challenging than first envisioned. Even today, there is no agreement on tactile sensing technology. Also, the mundane answer from one engineer on the need for improved electronic control using existing technology transpired to be correct in predicting the improvements required for industrial robots in the 1980s. However, given the arguments in Harmon's paper, it is more surprising that these tactile robotic applications were not just delayed by a decade or two but remain unsolved today. 

%: e-skin fabrication still remains an active research area with little agreement around any a technology (unlike the convergence of image sensors around CCD/CMOS semi-conductor technology).  
%``A good production engineer can solve almost any specific problem by using a special jig, fabricating a new part, developing a special tool for the hand, or replacing the hand with a special tool. Each problem gets solved, to be sure, but the overall technology becomes antiquated.''

% \clearpage

\begin{table*}[b!]
\resizebox{\textwidth}{!}{%
	\renewcommand{\arraystretch}{1}
	\centering
	\begin{tabular}{@{}cccccc@{}}	
		\textbf{Type} & \textbf{Principle of operation} & \textbf{Developer or manufacturer} & \textbf{Status} & \textbf{Comments}  \\
        \hline
    	Capacitive-based & Change in electrical properties by pressure & Bell Labs, N.J. (Bioe) & Experimental & 8$\times$8 array; displayed with video output \\
        & & Stanford Univ. (Fearing) & Experimental & 7$\times$8 hemispherical array \\
        & & Sussex Univ., U.K. (Jayawant) & Experimental & depressed pin-mechanism; 4$\times$4 flat array \\
		\hline
		Conductive/resistive & Change in electrical properties by pressure & Barry Wright Corp., Mass. & Commercial & `Sensorflex' (128 or 257 element versions) \\
        &  & JPL, Calif. & Experimental & 4$\times$8 element tactile array \\
        &  & MIT, Mass. (Hillis) & Experimental & 16$\times$16 array with 1-100\,g range \\
        &  & CMU, Penn. (Raibert) & Experimental & Resistive 4$\times$3 array with VLSI processing \\
        &  & Univ. of Pennsylvania (GRASP lab) & Experimental & Resistive 133-taxel (binary) probe \\
        \hline
		Magnetic-based & Change in magnetic field by pressure & Chuo Univ., Japan (Kinoshita/Mori) & Experimental & 20 Hall effect sensors each below a magnet \\
        &  & N. Carolina State Univ. (Luo) & Experimental & 16$\times$16 array; magnetic induction; video display\\
		\hline
        Optoelectric-based & Mechanically-induced light-intensity modulation & Tactile Robotic Systems, Calif. & Commercial & 16$\times$16 array, optomechanical cells \\
        & & Lord Corp., N. Carolina & Commercial & 4$\times$8 array; LED/photodiode mechanism \\ 
        & fiber-optic based & MIT, Mass. (Man-Machine Lab.) & Experimental & fiber-optic bundle; visual display; $>$330$/$cm$^2$ \\
        & & JPL, Calif. & Experimental & fiber-optic array captured with sensing cells \\
        & Tactile image captured with CCD camera & Aber Intell. Systems (AIS), Wales & Commercial & Videos of tactile images; $>$28000 elements$/$cm$^2$ \\
        \hline
    	Piezoelectric & Generation of electric charges by deformation & Univ. of Pisa, Italy (Dario/De Rossi) & Experimental & 8$\times$16 array; 2 active layers imitating human skin \\
        & & Univ. of Florida (Nevill/Paterson) & Experimental & Vibration sensitive; based on fingertip ridges \\
        \hline
		Strain gages & Change in electrical resistance by pressure & Transensory Devices Inc., Calif. & Commercial & 3$\times$3 cell array (also as individual cells) \\
         & & Flexigage Ltd, Scotland & Commercial & Used in several later tactile sensors \\
		\hline
	\end{tabular}}
	\caption{Representative commercial and experimental tactile sensors reviewed by \cite{dario_tactile_1985}, including the details of other tactile sensors of that time reviewed by~\cite{pennywitt_robotic_1986} and~\cite{nicholls_survey_1989}.}
	\label{table:2}
\end{table*}

\section{1985-94 Growth of Tactile Robotics}
\label{sec:foundations} 

By the mid-1980s, the sense of touch had gathered momentum. A new generation of researchers were drawn to tactile robotics with optimism about enabling human-like capabilities. This growth resulted in many new biomimetic and engineered tactile sensor designs, with some becoming commercial products. Meanwhile, the nature and complexity of a tactile sense for robotics was beginning to be understood, with insights entering the field from physiology and neuroscience. This progress was captured in a succession of review articles, including ``Tactile Sensing and the Gripping Challenge'' (\cite{dario_tactile_1985}), ``Tactile Sensing in Robotics'' (\cite{nicholls_survey_1989}), and ``Tactile Sensing and Control of Robotic Manipulation''~(\cite{howe_tactile_1993-1}).

Two early researchers of that generation, Paulo Dario and Danillo De Rossi, were inspired by ``increasing the performance of [tactile sensors as] a first step towards robotry that can hold and manipulate objects as humans do''~(\cite{dario_tactile_1985}). They formed a cluster of interdisciplinary research in Pisa, Italy, where progress in robotics combined knowledge from the biological sciences with mechanical engineering (research now known as {\em biorobotics} or {\em neurobotics}). The goal of their research was to create an artificial sense of touch integrated with its motor control, extending from a tactile fingertip to an actuated finger to a functional robotic hand as a complete system. 

Motivated by the functionality of human skin, \cite{dario_tactile_1985} placed emphasis on grasping and manipulating objects safely: ``the primary goal of an ideal tactile sensor is to measure variable contact forces on a sensing area'', including to ``control the delicate manipulation of objects by providing information on the tangential forces.'' Other useful measurements included ``the hardness and thermal properties of different materials'' and ``whether an object has been grasped safely and when slippage occurs.'' 

They surveyed the main types of tactile sensor of that time with an emphasis on commercialization (see Table~\ref{table:2}). Here, their table has been extended to include other contemporary tactile sensors of the 1980s (Table~\ref{table:2}, below), using the review articles by~\cite{pennywitt_robotic_1986} and~\cite{nicholls_survey_1989}. As part of their survey, Dario and De Rossi also listed the prices of commercial tactile sensors, which were \mbox{\$2\,000-\$4\,000} (equivalent to \$6\,000-\$12\,000 in 2025). This range is not dissimilar to that for the later tactile sensors in Table~\ref{table:X}. 

\begin{figure}[t!]
	\centering
    \begin{tabular}{@{}c@{}}
    \small{\bf (a) Biomimetic piezoelectric tactile sensor (1985)}\\
	\includegraphics[width=0.75\columnwidth,trim=0 0 0 0,clip]{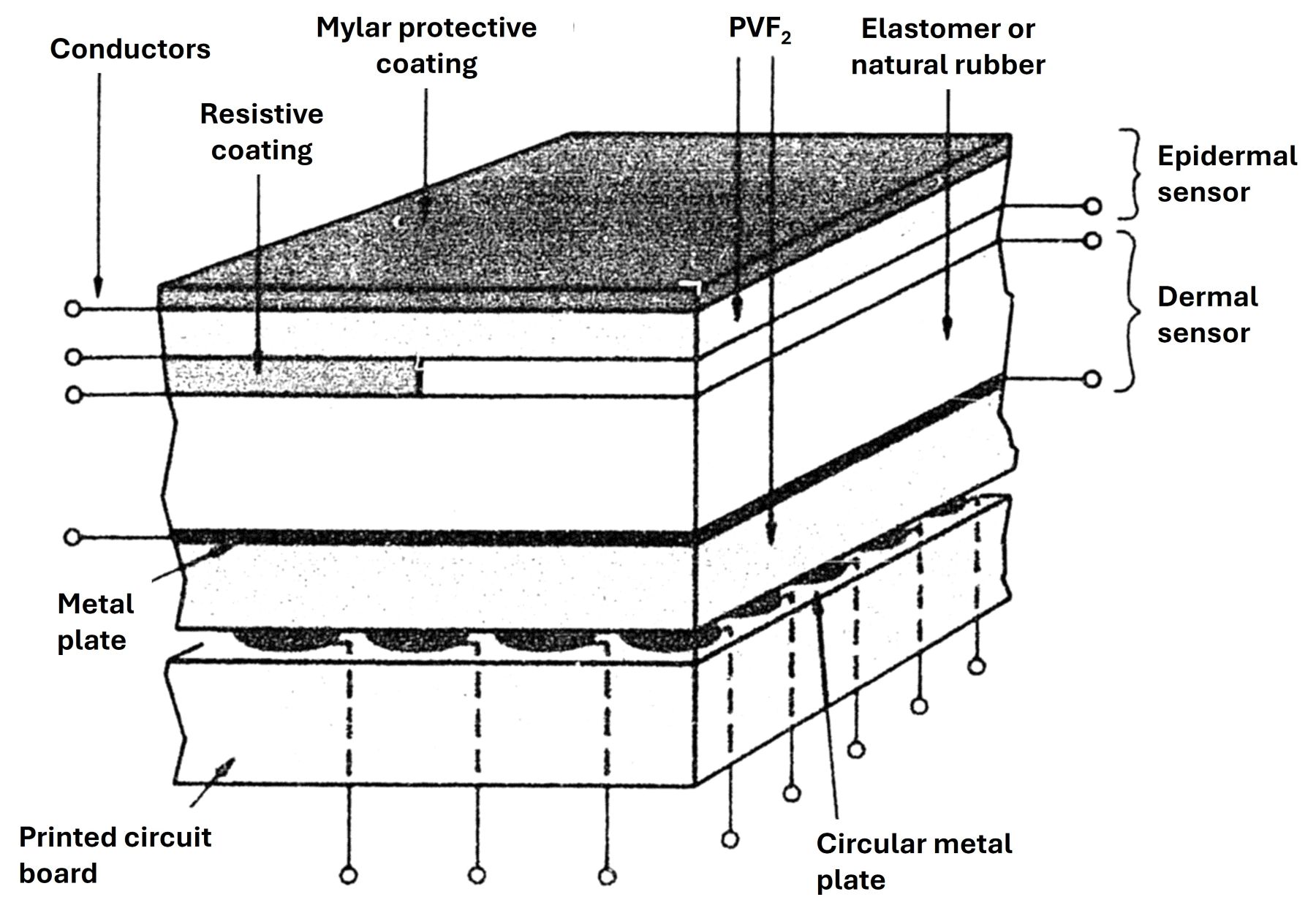}\\
    \end{tabular}    
    \begin{tabular}{@{}c@{}c@{}}
    \small{\bf (b) Anthropomorphic finger (1986)}\ \ \ & \small{\bf \ \ \ (c) Tactile fingertip (1986)} \\  
	\includegraphics[width=0.45\columnwidth,trim=20 0 10 0,clip]{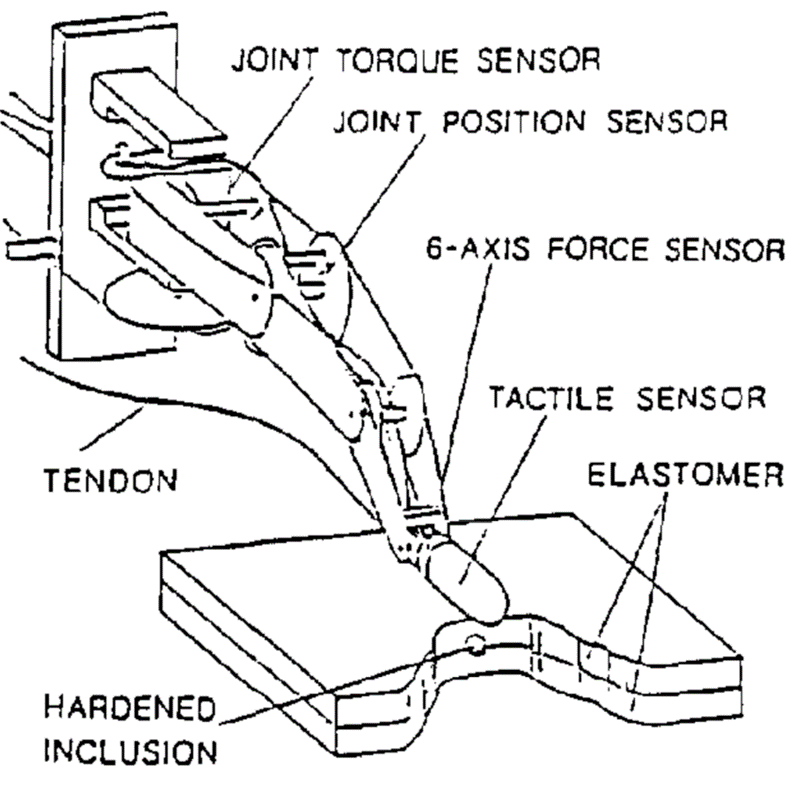} &
	\includegraphics[width=0.4\columnwidth,trim=0 -100 10 0,clip]{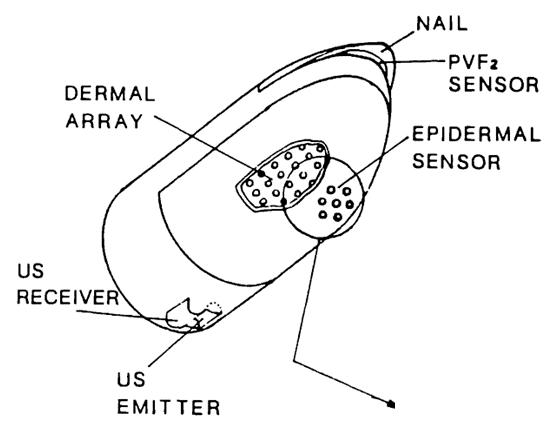}
    \end{tabular}
	\caption{Biomimetic tactile skin and fingertip from the mid-1980s. (a)~Layered epidermal (outer) and dermal (inner) tactile skin using a piezoelectric material (PVF$_2$). (b)~An anthropomorphic finger with (c) a biomimetic tactile fingertip using a modified version of the piezoelectric skin. (Images from~\cite{dario_tactile_1985} and~\cite{dario_tactile_1991}.)}
    \vspace{-1em}
	\label{fig:7}
\end{figure}
% (a) Fig 7 of https://ieeexplore.ieee.org/abstract/document/6370785?casa_token=WOjzUpcRWjIAAAAA:n03rZXhZ14OnPsIKUa2FNIcVYZNCPDPy9czRmP_JqSWU1Zb9WAlMZWdgIUDQo0duCv5O9wdK
% (b) from https://www.sciencedirect.com/science/article/abs/pii/092442479187001J Elsevier

An incentive to write a review is when your research gives you insight into the promises and challenges of a research area. Dario and colleagues were developing a biomimetic tactile sensor based on the layered structure of human skin~(Figure~\ref{fig:7}a), using piezoelectric material that generates local changes in electrical properties under contact pressure. Their design was one of the first multi-modal tactile sensors, combining a deeper dermal layer using an array of pressure-sensitive conductive plates with a shallower epidermal layer composed of a thin Mylar film sensitive to gross deformation changes and temperature. A later fingertip design included a tactile ``fovea'' sensitive to rapidly-changing contacts at the epidermis (Figure~\ref{fig:7}c; \cite{dario_force_1988}). This biomimetic fingertip was part of a body of research from the University of Pisa on the creation and control of an anthropomorphic finger (Figure~\ref{fig:7}b; \cite{dario_anthropomorphic_1987}) to mimic aspects of how humans explore and interact with objects~(see reviews by \cite{dario_force_1988,dario_tactile_1991}).

\begin{table*}[b!]
\resizebox{\textwidth}{!}{%
	\renewcommand{\arraystretch}{1}
	\centering
	\begin{tabular}{@{}ccc@{}}
    {\bf Transduction} & {\bf Advantages} & {\bf Disadvantages} \\
    \hline
    \multirow{2}{*}{\shortstack{{\bf Resistive \& conductive}}} 
    & Wide dynamic range; Durability; Good overload tolerance;  & Hysteresis; Highly dependent on elastomer properties; \\
    & Compatibility with ICs, e.g. VLSI & Limited spatial resolution; Lot of wiring; Non-linear response \\
    \hline
    \multirow{2}{*}{\shortstack{{\bf Piezoelectric \& } \\ {\bf pyroelectric effects}}} 
    & Wide dynamic range; Durability; Good material properties; & Difficult separating piezo- \& pyroelectric effects; Complex Designs; \\
    & Temperature and force sensing & Inherently dynamic transient response$^\dagger$; Difficult to scan elements \\
    \hline
    \multirow{1}{*}{\shortstack{{\bf Capacitive techniques}}} 
    & Wide dynamic range; Linear response; Robust & Noise susceptible; Temperature sensitive; Limited spatial resolution \\
    \hline
    \multirow{1}{*}{\shortstack{{\bf Magnetic transduction}}} 
    & Wide dynamic range; Large displacements possible; Simple & Poor spatial resolution; Mechanical problems \\
    \hline  
    \multirow{2}{*}{\shortstack{{\bf Magnetoelastic \& } \\ {\bf magnetoresistive}}} 
    & Wide dynamic range; Low hysteresis; Linear response; & Susceptible to stray fields \& noise; AC circuitry required  \\
    & Robustness; Normal force, shear \& torque sensitive & \\
    \hline 
    \multirow{1}{*}{\shortstack{{\bf Mechanical methods}}} 
    & Well-known tech; Good for probe applications & Limited spatial resolution; Complex for arrays \\
    \hline 
    \multirow{2}{*}{\shortstack{{\bf Optical methods}}} 
    & Very high resolution; Compatible with vision technology; & Dependence on elastomer; Some hysteresis \\
    & Can be robust to electrical interference; Low cabling &  \\
    \hline 
 \end{tabular}}
    \vspace{0em}
	\caption{Pros and cons of various transduction technologies for tactile sensing (summarized from \cite{nicholls_survey_1989}).}
    \vspace{0em}
	\label{tab:2}
\end{table*} 

\begin{figure}[t!]
	\centering
	\begin{tabular}{@{}c@{}}
    \small{\bf (a) Fibre optic-based tactile sensor (1984)}\\
	\includegraphics[width=\columnwidth,trim=0 0 0 0,clip]{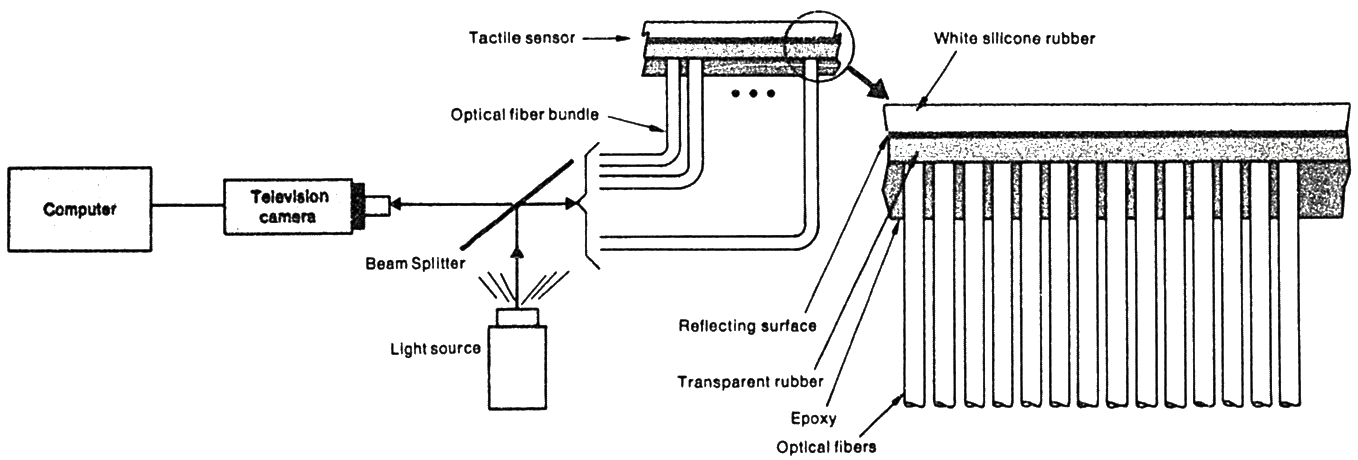}\\
    \small{\bf (b) Commercial optoelectronic tactile sensor (1985)}\\
    \includegraphics[width=0.45\columnwidth,trim=0 150 0 90,clip]{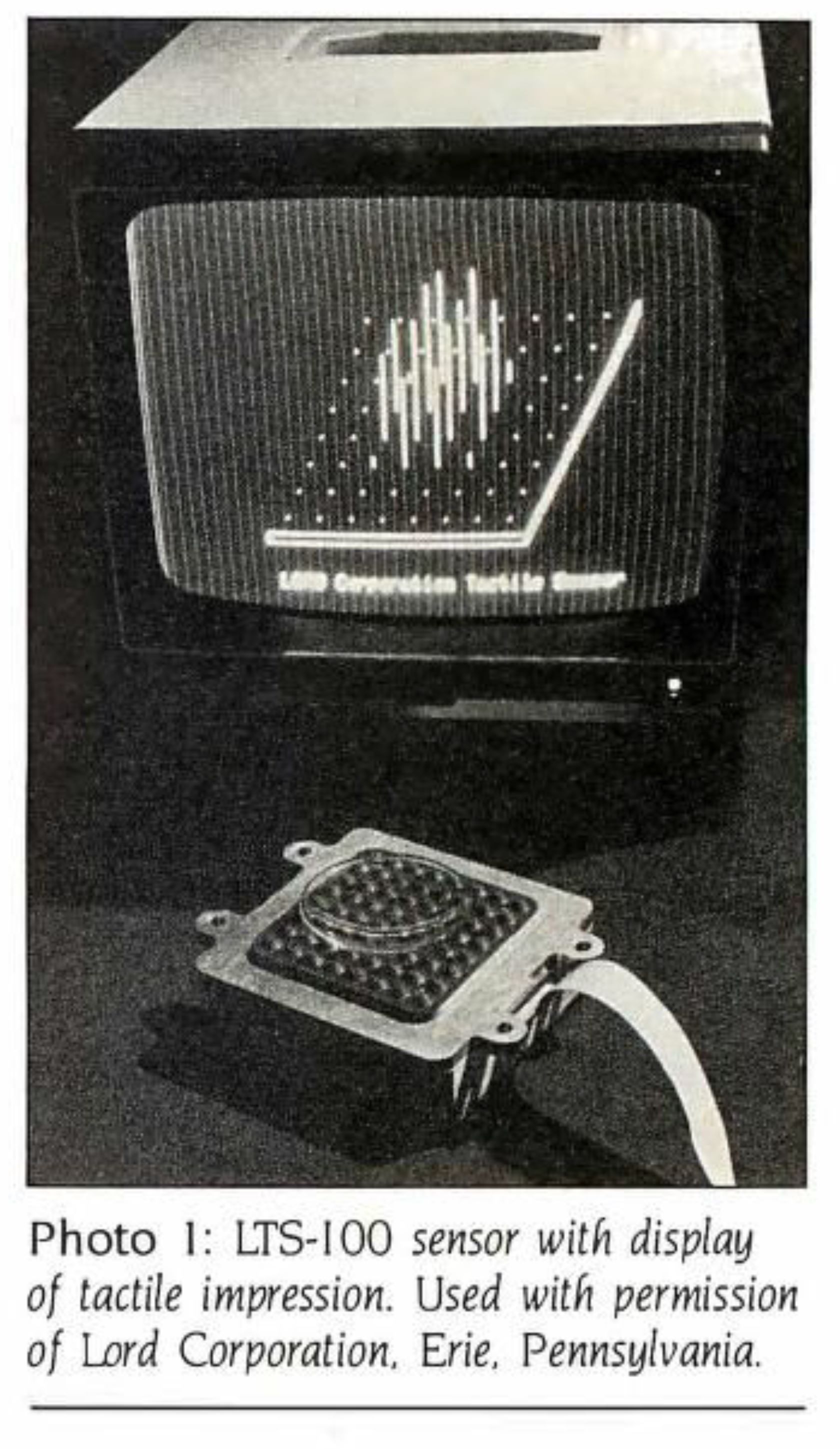}
    \includegraphics[width=0.53\columnwidth,trim=10 40 248 40,clip]{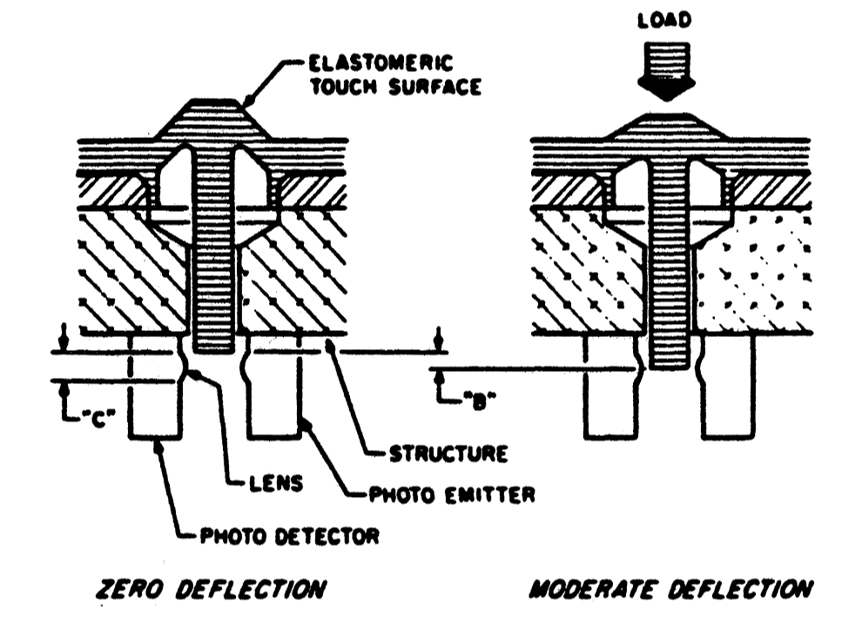}\\
    {\small{\bf (c) Commercial vision-based tactile sensor (1987)}}\\
    \includegraphics[width=0.8\columnwidth,trim=-20 0 -20 0,clip]{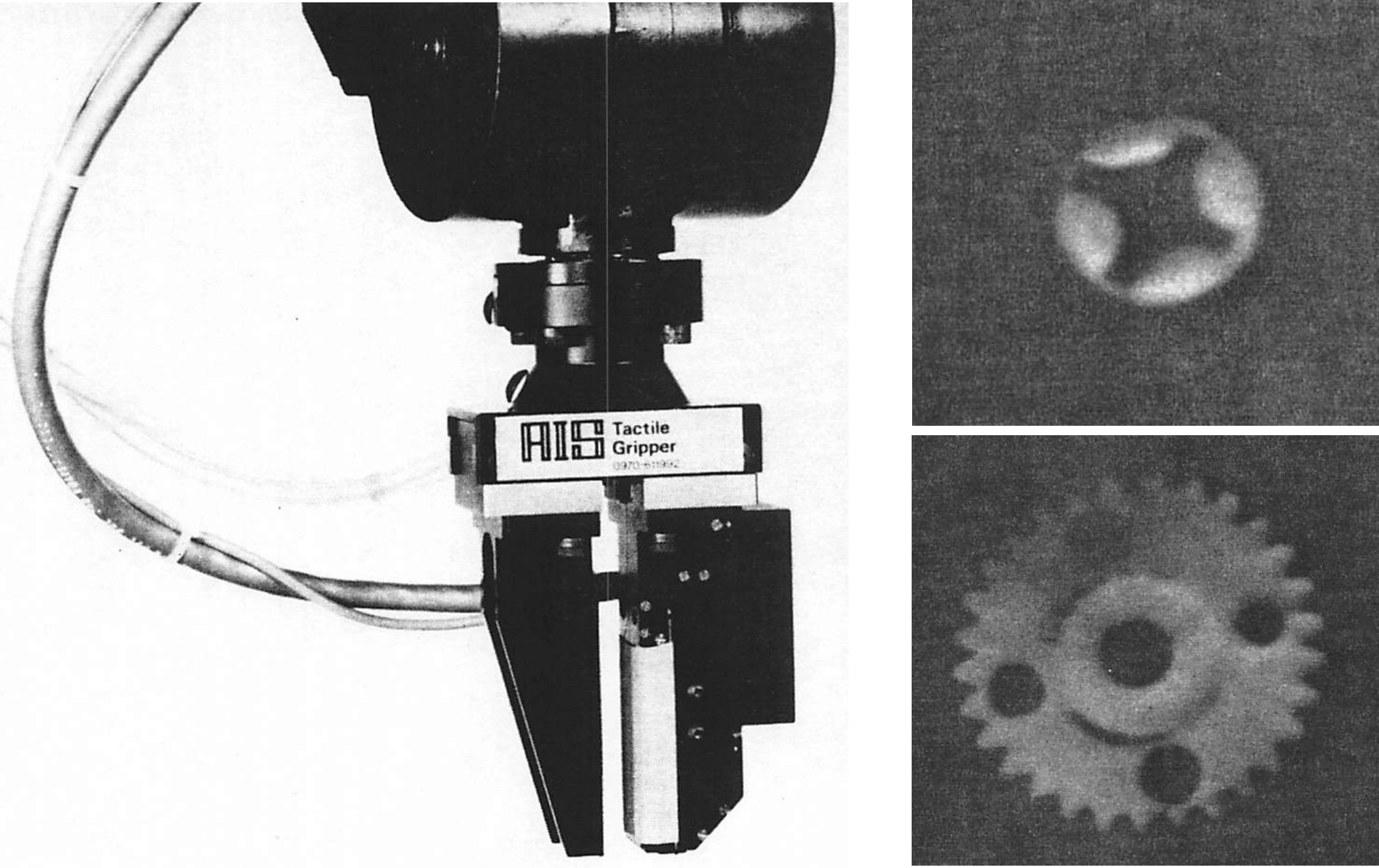}
    \vspace{-1em}
	\end{tabular}
     \caption{Optical tactile sensors from the mid-1980s. (a)~Fiber optic tactile sensor (MIT Man-machine lab, 1984);~(b)~Opto-\\electronic tactile sensor (Lord Corporation, 1986) with display and mechanism; (c)~Vision-based tactile sensor mounted on a gripper (Aber Intelligent Systems, 1987) with high-resolution tactile images. (Images from \citep{dario_tactile_1985}, \citep{agrawal_overview_1986}, and \citep{mcclelland_giving_1987}.) }
	\label{fig:6}
\end{figure}
% (a) Fig 3 of https://ieeexplore.ieee.org/abstract/document/6370785
% (b) Fig 4 of https://ieeexplore.ieee.org/abstract/document/6370785
% can replace with agrawal and jain fig 12 or fig 3 of jayawant
% (c) figure from https://www.emerald.com/sr/article-abstract/7/4/203/346775/Giving-tactile-sensors-a-good-image
% also tactile images from Fig 8 of https://iopscience.iop.org/article/10.1088/0022-3735/22/9/002 (Jayawant)

Another tactile sensing technology that was well covered in reviews at that time was optoelectronics~(\cite{dario_tactile_1985,pennywitt_robotic_1986,nicholls_survey_1989}). Optical tactile methods were considered promising because of their high sensitivity in that a ``small change in force results in a relatively large change in light intensity''. Some examples are shown in Figure~\ref{fig:6}, including: (a)~a~fiber optic tactile sensor from MIT~(\cite{schneiter_optical_1984-1}), which both emitted and received light down the same fibers to image a deformable reflective layer underneath an opaque outer rubber surface; (b)~a~commercial optoelectronic tactile sensor with an array of pin-like tactile elements connected to an elastomeric pad, where depressed pins partially block light between paired LEDs and photodetectors; and (c)~an early vision-based tactile sensor using a CCD camera to capture a high-resolution  (128$\times$256 pixels) tactile image of light internally refracted from a deformable membrane~(\cite{mott_experimental_1984}). This latter work, by Mott, Lee and Nicholls from Aberystwyth, Wales, pioneered vision-based tactile sensing 30 years before it became a main theme of tactile robotics (see Section~\ref{sec:growth}).

\cite{nicholls_survey_1989} also wrote an authoritative review in {\em IJRR} of the tactile sensing technology of that time as a successor to the review by~\cite{harmon_automated_1982}. They extended the definition of tactile sensing to transduce five main categories of data: {\em simple contact, force, shape, slip, and thermal properties}; and so defined that ``{\em the acquisition, processing and manipulation of this data constitutes tactile sensing.}'' Their definition has a wider coverage of tactile sensors than Harmon's original ``graded signals across force-sensitive surfaces'', which had become too narrow for the advances that they surveyed over that decade. 

\begin{figure}[t!]
	\centering
	\begin{tabular}{@{}c@{}}
    \small{\bf (a) Capacitive tactile sensor (1985)}\\
	\scalebox{1}[0.75]{\includegraphics[width=0.8\columnwidth,trim=0 0 0 60,clip]{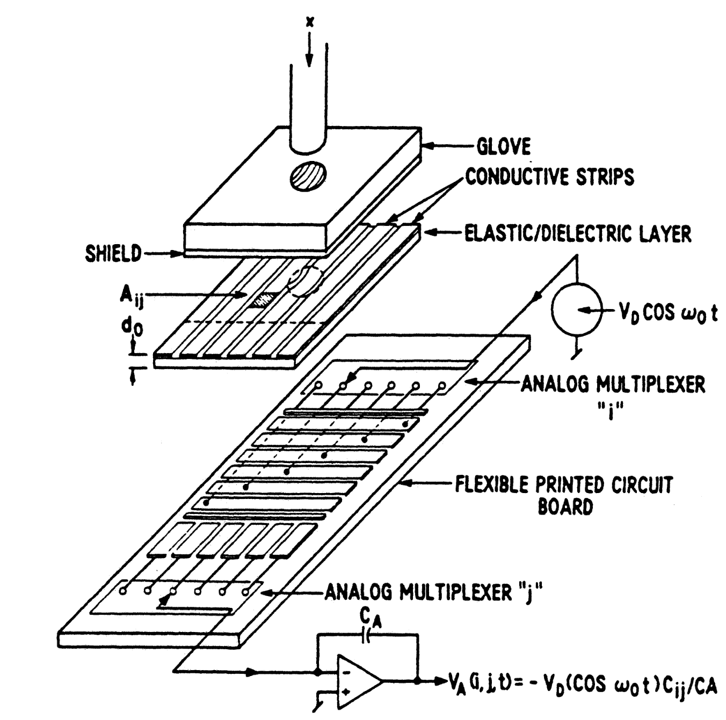}}\\
    \small{\bf (b) Strain gage-based tactile sensor (1985)}\\
	\includegraphics[width=\columnwidth,trim=0 0 0 0,clip]{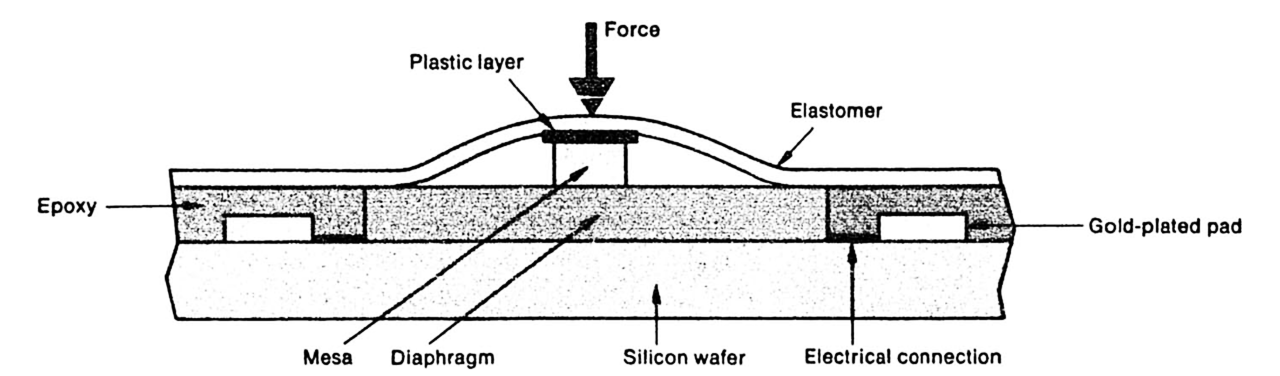}\\
    \small{\bf (c) Capacitive pin-based tactile sensor (1985)}\\
	\includegraphics[width=\columnwidth,trim=0 25 0 0,clip]{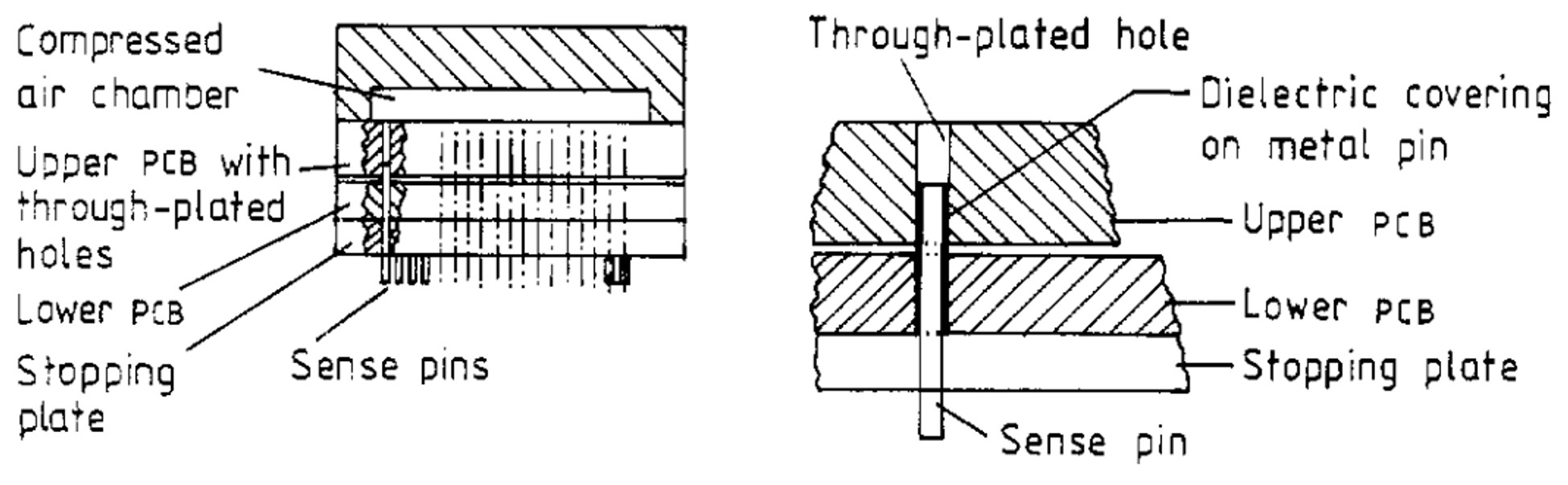}\\
    \vspace{-1em}
	\end{tabular}
     \caption{Electromechanical tactile arrays from the mid-1980s. (a) Capacitive tactile 6$\times$6 array. (b) Strain gage deployed as a tactile sensor. (c) Capacitive transducer using a pin mechanism to measure normal contact. (Images from \citep{agrawal_overview_1986}, \citep{dario_tactile_1985}, and~\citep{jayawant_tactile_1989}.)}        
     \vspace{-1em}
	\label{fig:6A}
\end{figure}

By the late 1980s, the methods for tactile transduction had separated into distinct classes that remain the main methods in development and use today. These include piezoelectric sensors ({\em e.g.,} Figure~\ref{fig:7}) and various types of optical tactile sensor (Figure~\ref{fig:6}), alongside other technologies such as capacitive and strain gage-based (examples shown in Figure~\ref{fig:6A}). After surveying the technologies being considered for tactile sensing,~\cite{nicholls_survey_1989} usefully distilled the pros and cons of each transduction technology (summarized here in Table~\ref{tab:2}). These benefits and limitations seem current today; for example, many of the mechanisms are limited by having low-resolution or complex designs.

% (a) Fig 7 of Agrawal and Jain https://deepblue.lib.umich.edu/bitstream/handle/2027.42/3180/and6076.0001.001.pdf
%no copyright https://ntrs.nasa.gov/citations/19870007116
% (b) Fig 5 of https://ieeexplore.ieee.org/abstract/document/6370785
% (c) Fig 13 of https://iopscience.iop.org/article/10.1088/0022-3735/22/9/002 

The 1980s was still a time when topical reviews of leading research could be published in general-interest magazines. In 1986, {\em Byte}, a highly successful microcomputer magazine, ran a special issue on robotics (with accompanying artwork, still available online). Its content was guided by the questions: ``What makes robotics so hard? Why is it taking so long to develop this technology?'' The editors commissioned articles on: ``Machine vision'', ``Robotic tactile sensing'', ``Multiple robotic manipulators'', ``Autonomous robot navigation'', ``AI in computer vision'', and ``Automation in organic synthesis.'' Among these, the article by~\cite{pennywitt_robotic_1986} provided a high-quality early review of tactile sensing up to the mid-1980s. Although much of the material overlaps with other reviews at that time, it stands out in its quality and professionalism.

\begin{table*}[b!]
\vspace{0em}
\resizebox{\textwidth}{!}{%
	\renewcommand{\arraystretch}{1}
	\centering
	\begin{tabular}{@{}cccccccc@{}c@{}}	
		 & \textbf{Type} &\textbf{Adaptation} & \textbf{Location} & \textbf{Endings} & \textbf{Spatial} & \textbf{Frequency} & \textbf{Receptor} & \textbf{Stimulus sensitivity} \\
		 & & \textbf{rate} & \textbf{of endings} & \textbf{per unit} & \textbf{acuity} (mm) & \textbf{sensitivity} (Hz) & \textbf{density} (cm$^{-2}$) &  \\
		\hline
		Merkel disks & SA-I & Slow & Shallow & 4-7 (clustered) & 0.5 & 0.4-3 & 70 & Sustained pressure; spatial deformation \\
		Meissner's corpuscles & FA-I & Fast & Shallow & 12-17 & 3-4 & 3-40 & 140 & Temporal changes in skin deformation; slip \\
		Ruffini endings & SA-II & Slow & Deep & 1  & 7+ & 100-500+ & 49 & Lateral skin stretch \\
		Pacinian corpuscles & FA-II & Fast & Deep & 1 & 10+ & 40-500+ & 70 & High-frequency skin deformation; vibration \\
		\hline
	\end{tabular}}
	\caption{Mechanoreceptor properties at the human fingertip. Based on~(\cite{jayawant_tactile_1989}, Table 1), combined with details from (\cite{dargahi_human_2004}, Table 2) and (\cite{dahiya_tactile_2010}, Figure 3). See Figure~\ref{fig:9A} for a diagram of human tactile skin.}
    % \vspace{-1.5em}
	\label{tab:4}
\end{table*}
% From Jayawant - All requests should be sent by email to: permissions@ioppublishing.org

% \begin{figure}[t!]
% 	\centering
%     \begin{tabular}{@{}c@{}c@{}}
% 	\includegraphics[width=0.495\columnwidth,trim=0 0 0 0,clip]{figures/1986_01_BYTE_11-01_Robotics_0000.jpg} &
% 	\includegraphics[width=0.49\columnwidth,trim=0 0 0 0,clip]{figures/1986_01_BYTE_11-01_Robotics_0169.jpg}
%     \end{tabular}
% 	\caption{Left: cover from the January 1986 Issue of {\em Byte} Magazine. Right: introductory image for the editorial on robotics. At this time, {\em Byte} had a circulation of 400\,000, just behind its newer rivals PC World and PC Magazine, but with a broader coverage of computing rather than just products.}
% 	\label{fig:8}
% \end{figure}
% % ~\cite[page 159]{noauthor_byte_1986})

From the mid-1980s to mid-1990s, a new review of tactile sensing was published every year or so. Their content overlapped, particularly when surveying tactile technologies, but they all contributed new perspectives. In a NASA-commissioned report, \cite{agrawal_overview_1986} gave an account of tactile and vision integration, proposing that ``if the object recognition system makes a hypothesis about [an] object, tactile sensing can be used to verify model features.'' In a book on ``Robot Tactile Sensing'',~\cite{russell_robot_1990} surveyed artificial tactile sensing for undergraduate students. In a review of ``Artificial Tactile Sensing and Haptic Perception'', \cite{de_rossi_artificial_1991} gave a new treatment of tactile transduction and introduced models of contact mechanics, friction, texture, and rheology.

\begin{figure}[t!]
    % \vspace{-1em}
	\centering
    \begin{tabular}{@{}c@{}}
	\includegraphics[width=0.85\columnwidth,trim=0 0 0 0,clip]{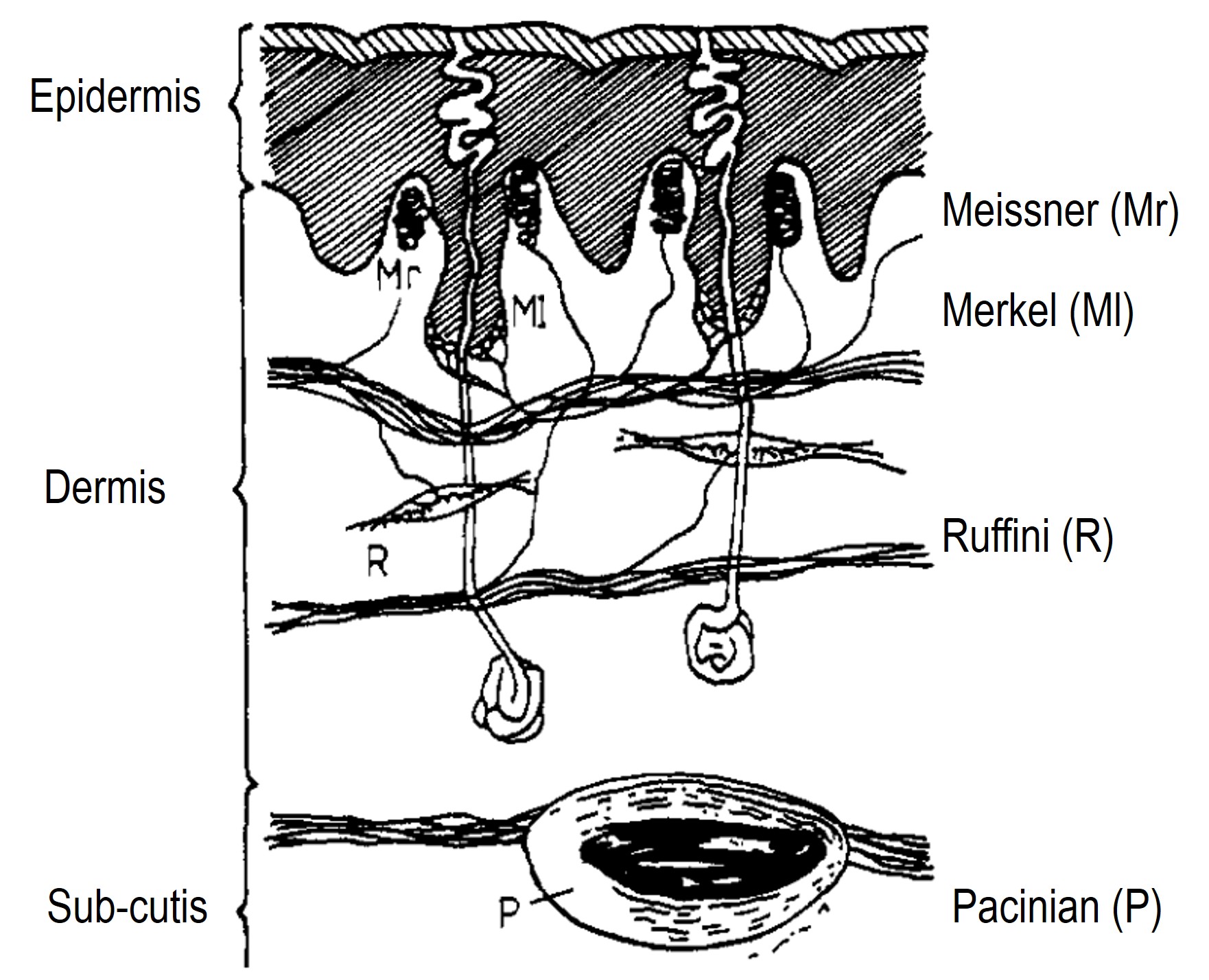} 
    \end{tabular}
	\caption{The layered structure of fingertip skin. MI, Merkels neurite  complex; Mr, Meissner corpuscle; R, Ruffini cylinder; P, Pacinian corpuscle. (Image from (\cite{jayawant_tactile_1989}, Figure 1).)}
	\label{fig:9A}
\end{figure}
% Fig 1 of https://iopscience.iop.org/article/10.1088/0022-3735/22/9/002 (Jayawant)

A central theme that inspired artificial tactile sensing in this period and beyond was the physiology and function of the human sense of touch. \cite{harmon_tactile_1984}, in his last article ``Tactile Sensing for Robots,'' covered the physiology of the human tactile sense in some detail, including the types and properties of tactile mechanoreceptors. In his view, ``consideration of some of what we know about ourselves could illuminate or at least stimulate our machine design.'' He also conveyed that human touch is fundamentally an {\em active} sense, in that ``the hand should properly be considered the sense organ for touch rather than the mechanoreceptors, and... emphasis ought to be on the active seeking of information by the exploring hand.'' At that time, these aspects of human touch were being explored in the life sciences, some of which  translated to robotics~({\em e.g.,} \cite{lederman_lessons_1992} in a book by~\cite{nicholls_advanced_1992}).

A detailed survey of human tactile neurophysiology for roboticists was given in ``Tactile Sensing in Robotics'' by  \cite{jayawant_tactile_1989}, including full coverage of mechanoreceptor properties (see Table~\ref{tab:4} and Figure~\ref{fig:9A}). Like Harmon, he became a researcher in tactile robotics after an established career in another area, in his case as a pioneer in magnetic levitation applied to high-speed trains. Using his engineering experience, he proposed capacitance transducers built on arrays of movable pins as a robust and scaleable technology for tactile sensing (Figure~\ref{fig:6A}c).

\cite{howe_tactile_1993-1} also described some lessons from human tactile sensing for robot dexterity, including that a diversity of receptors in skin and muscle respond to a variety of stimuli. He emphasized that these receptors can be highly sensitive ({\em e.g.,} $\sim$1\,$\rm\mu$m perceivable static height change), but are also hysteretic, non-linear, time-varying and respond to many physical parameters simultaneously. In addition, nerve conduction is slow (usually $<$60\,m/s), so animals use feedforward control for voluntary actions in tasks requiring quick reactions. Although robots can use faster signaling, anticipatory control may likewise help perform such tasks

His focus was on ``tactile sensing and control of robot manipulation''~(\cite{howe_tactile_1993-1}) (related to a chapter by \cite{howe_touch_1992}, now out of print). Specifically, dexterous manipulation ``requires control of forces and motions at the contact [with] the environment, which can only be accomplished through touch''. This perspective contrasted with earlier perspectives that focused mainly on hardware and applications but omitted how tactile sensors could be controlled to do those tasks. The motivation that ``we lack a comprehensive theory that defines sensing requirements for various manipulation tasks'' remains current today, as do the sources of knowledge: ``investigations of human sensing and manipulation, and mechanical analyses of grasping and manipulation.''.   %The last influential review of this period was on ``tactile sensing and control of robot manipulation'' by Robert Howe from Harvard University.

\begin{table*}[b!]
% \vspace{-1em}
\resizebox{\textwidth}{!}{%
	\renewcommand{\arraystretch}{1}
	\centering   
	\begin{tabular}{@{}l@{}ll@{}ll@{}l@{}}
    \multicolumn{2}{c}{\bf Surgery \& medicine} & \multicolumn{2}{c}{\bf Healthcare \& service robotics} & \multicolumn{2}{c}{\bf Natural produce processing} \\
    \hline
    \ \ \ {\bf Feature} & \ \ \ {\bf Need} & \ \ \ {\bf Feature} & \ \ \ {\bf Need} & \ \ \ {\bf Feature} & \ \ \ {\bf Need} \\
    Very rapid take-up & Restore taction in MIS & Demographic projections & Personal space manipulation & High volume & High speed \\
    Disposable equipment & Laparoscopy improvements & Enormous demand imminent\ \ & Mobility aids & Human excluded environment\ \ & Inspection function \\
    Sophisticated users & Remote palpation & Cost reduction essential & Automated household tools & Versatile, product changes & Consistency \\[0.2em]
    \ \ \ {\bf Technical issue} & \ \ \ {\bf Challenge} & \ \ \ {\bf Technical issue} & \ \ \ {\bf Challenge} & \ \ \ {\bf Technical issue} & \ \ \ {\bf Challenge} \\
    Telepresence & Force and tactile feedback & Haptic exploration, dexterity & Safety & Soft, delicate items & Active handling control \\
    Teletaction & Mobility, fine control & Adaptation, customization & Reliability & Irregular objects & Reliability \\
    Soft tissue discrimination\ \  & Hardness/softness sensing & Low costs & User acceptance & Long run-times & Hygiene \\
    \hline 
 \end{tabular}}
 \vspace{-0.2em}
	\caption{A summary of future application areas of tactile sensing and their needs and challenges. Based on \cite[Table 4]{lee_tactile_2000}.}
	\label{tab:3}
    % \vspace{1em}
\end{table*}

\begin{figure}[t!]
    % \vspace{-1em}
	\centering
    \begin{tabular}{@{}c@{}}
	\includegraphics[width=\columnwidth,trim=0 0 0 0,clip]{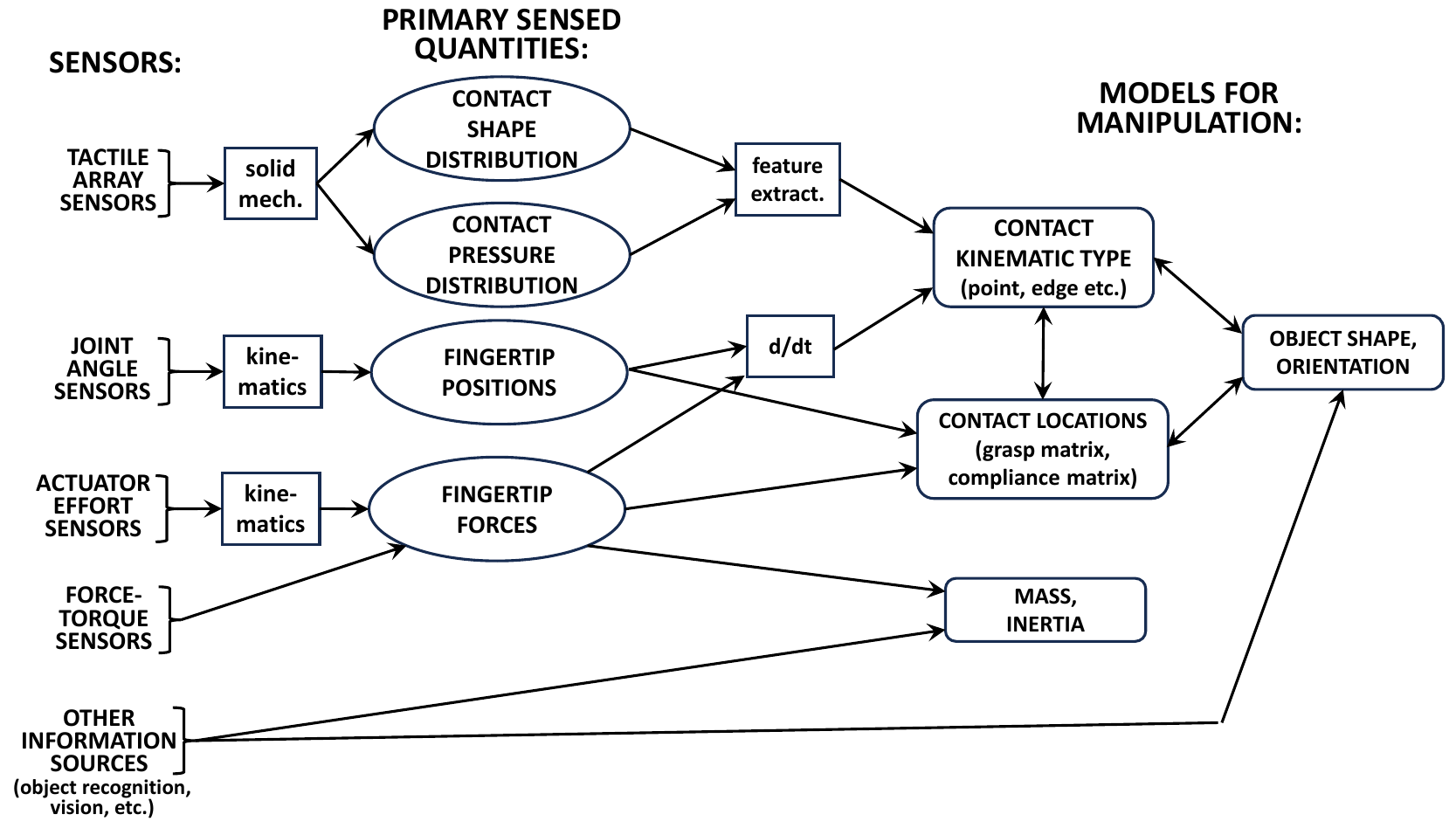} 
    \end{tabular}
    \vspace{-1em}
	\caption{Use of geometric touch information in manipulation. The diagram links types of sensor (left) to their primary sensed quantities (middle) and how they update models for manipulation (right). Based on (\cite{howe_tactile_1993-1}, Figure 2).}
    \vspace{-1em}
	\label{fig:9}
\end{figure}

In considering control as the primary use of tactile sensing,~\cite{howe_tactile_1993-1} distinguished two ways by which tactile sensing can provide useful information for manipulation: (1)~{\em geometric features}, including contact shapes and forces; and (2)~{\em contact-condition features}, including local friction, slip, and transitions between different stages of the task ({\em e.g.,} making contact or losing grip). Touch-derived quantities such as contact shape/pressure and fingertip positions/forces can then be input to models for use in controlling the manipulation of object shape and pose (see Figure~\ref{fig:9}). Likewise, similar connections can be made to models for detecting task phases, using dynamic information such as contact vibration and slip detection~(\cite{howe_tactile_1993-1}, Figure 4). %Howe comment that it is not known ``how to structure a control system to use this information'' and the greatest task is to experimentally evaluate ``the role of touch sensing in manipulation.'' 

% https://www.tandfonline.com/doi/abs/10.1163/156855394X00356

However, these reviews by~\cite{howe_tactile_1993-1, nicholls_survey_1989} and others led to a more realistic view of tactile sensing technology. The final theme of this foundational generation of tactile robotics was a dawning realization of the major challenges to make these systems a practical reality, displacing the early optimistic forecasts of the first reviews. This sentiment is expressed in the closing comments of the last few surveys of this generation, which are quoted below.

``Despite the findings of the~\cite{harmon_automated_1982} survey, tactile sensing has not made any significant contribution to real applications in factory systems''~(\cite{nicholls_survey_1989}).

``Although the rapid growth in interest in the field of tactile sensing has initiated significant progress in [tactile] hardware, very little fall-out in real applications has occurred... Future widespread use of [tactile] systems is still foreseen, but the time scale for these events to occur should be realistically correlated with the great theoretical and technical difficulties associated with this field, and the economic factors that ultimately drive the pace of its development''~(\cite{de_rossi_artificial_1991}).

``Ten years ago, a commonly cited impediment to progress in tactile sensing was the lack of suitable [devices] and algorithms... The primary issues in touch sensing are [now] the integration of these devices and algorithms into practical manipulation systems.'' Also, ``contact tasks where small forces and displacements must be controlled'' are ``beyond the capability of even laboratory robots, and understandably industry avoids these tasks''~(\cite{howe_tactile_1993-1}).

\section{1995-2010: Tactile Robotics Winter}
\label{sec:pessimism}
 
There followed a long `tactile robotics winter' from the mid-1990s to the late 2000s when academic activity dropped compared to the earlier foundational generation (Table~\ref{table:2}). This slowdown was against a background of growing robotics research ({\em e.g.,} the primary robotics conference, {\it ICRA}, grew from under 900 to over 2100 submissions). It is difficult to attribute the cause, but perhaps a realistic assessment of the challenges dispelled the early optimism that attracted researchers to tactile robotics. Also, the major research efforts to explore tactile sensing technologies (see Tables~\ref{table:2} and \ref{tab:2}) had not resulted in a clear leader, but rather many candidates that all had problems.   

These sentiments were expressed by the authors of the few influential reviews written at that time, which were successors to those of the previous generation: ``Tactile~sensing for mechatronics''~(\cite{lee_review_1999}), ``Tactile sensing -- New directions, new challenges''~(\cite{lee_tactile_2000}), and ``Force and tactile sensors''~(\cite{cutkosky_force_2008}). Alongside this realism was a renewed focus on applications, with surgery, healthcare, and food production considered the most promising (Table~\ref{tab:3}), which attracted some new researchers to tactile robotics. Moreover, even though academic research slowed, there remained steady progress in commercial tactile sensing, particularly for robotic hands~(Table~\ref{table:X}).

The slow progress in tactile robotics was well described by~\cite{cutkosky_force_2008}, who said: ``Tactile sensing has been a component of robotics for roughly as long as vision. However, [unlike] vision, ... tactile sensing always seems to be a few years away from widespread utility.'' Considering why: ``one reason for the slow development of tactile sensing technology, compared to vision, is that there is no tactile analog of the CCD or CMOS optical array.'' This remains the case, although perhaps with vision-based tactile sensing, the tactile analog of the CCD or CMOS array {\em is} that array.

%Maybe this drop in activity reflected that early optimism had been replaced by a sober realization of the challenges to be faced. Perhaps it was also due to the rapid progress in vision-based robotics, which would have attracted more funding and researchers to areas such as visual navigation (SLAM) and autonomous vehicles. 
%Whatever the reason, a 5-year hiatus followed with no influential reviews up to 1999, and then a new article only every couple of years. 

%The leading review from this period was ``Tactile Sensing for Mechatronics -- A State-of-the-Art Survey'' by~\cite{lee_review_1999} in the journal {\em Mechatronics}. This article built on their earlier survey in {\em IJRR} of robot tactile sensing technology~(\cite{nicholls_survey_1989}), updated with the literature of the 1990s and assessing ``near-future'' developments. In the 2000s, six influential reviews of tactile sensing stemmed from this survey, beginning with an assessment of new developments by~\cite{lee_tactile_2000} in {\em IJRR}. %The last article of this period, an authoritative survey of ``force and tactile sensors'' by~\cite{cutkosky_force_2008} in the {\em Springer Handbook of Robotics}, \ followed an earlier review~(\cite{howe_touch_1992}).

% \cite{lee_review_1999} noted that all the other basic human senses of sight, sound, taste and smell have well-developed electronic analogues available commercially. 

Similarly,~\cite{lee_review_1999} said that ``despite repeated suggestions in many papers regarding the potential importance of tactile sensing to industrial automation and other application areas, the literature reported very few applications beyond the experimental prototype stage and hardly any in serious regular use.'' Reflecting on the ``reasons for this slow commercial exploitation,'' they observed that:\\
\noindent 1) {\em There is no localized sensory organ}: transduction of tactile signals is distributed over a wide area, so the creation of an artificial tactile skin is more difficult than a discrete sensor.\\
\noindent 2) {\em The sensing is complex}: tactile sensing does not simply transduce one physical property into an electronic signal but has many forms; {\em e.g.,} shape, texture, friction, force, pain and temperature. It is not well understood how these are related and not easy to find suitable analogues within a single device.\\
\noindent 3) {\em Difficult to imitate}: it is unclear which are the best tactile signals for applications to develop further. In contrast, computer vision research can focus on understanding images because visual data capture is solved. Tactile sensing has not yet reached this point and is still concerned with data capture.\\
\noindent 4) {\em Lack of availability}: In addition to and because of these difficulties, there is a ``lack of availability of commercial sensors with suitable configurations and characteristics.'' 

\begin{table*}[b!]
\resizebox{\textwidth}{!}{%
	\renewcommand{\arraystretch}{1}
	\centering
	\begin{tabular}{@{}cccccc@{}}	
		\textbf{Type} & \textbf{Principle of operation} & \textbf{Developer or manufacturer} & \textbf{Status} & \textbf{Comments}  \\
		\hline
        Conductivity/resistivity & Change in electrical resistance & NASA/DARPA & Commercial & QTC for RoboNaut gloves: 33 taxels per hand \\
        & with pressure & Peratech, UK & Commercial & QTC for skin for Shadow Hand: 36 taxels over hand\\
        & & Weiss Robotics, Germany & Commercial & 78/84-taxel fingertips/phalange of Schunk SDH Hand\\
        & & Interlink Electronics, Calif. & Commercial & FSR for RobotNaut gloves: 19 taxels over hand\\
        & & TekScan, Mass. & Commercial & FSR to attach to Shadow Hand: 349 taxels over hand\\
        & & Gifu Univ. (Mouri) & Experimental & FSR to attach to Gifu hand: 624 taxels over hand\\
        & & Korea Univ. (Kim) & Experimental & MEMS-based 32$\times$32 array\\
        & & Taiwan Univ. (Cheng) & Experimental & PDMS-based 8$\times$8 flat array\\
        & & Univ. Tokyo (Someya) & Experimental & OFE transistor-based 32$\times$32 flat array\\
		\hline
    	Capacitive-based & Change in electrical capacitance & Italian Institute of Technology & Commercial & iCub tactile sensors: 12-taxel fingertip; 45 taxel palm \\
        & with pressure & Pressure Profile Systems, US/UK & Commercial & 22/24-taxel fingertips/palm (PR2 \& Barrett Hands) \\
        & & Minesota Univ. (Lee) & Experimental & MEMS-based 64-taxel flat array \\
        \hline
    	Impedance Tomography & Change in electrical impedance & Univ. Tokyo (Alirezaei) & Experimental & 16 electrodes on boundary of stretchable membrane \\
        \hline
    	Impedance-based & Change in electrical impedence & SynTouch Ltd, Calif. & Commercial & BioTac: 19-taxel fingertip; vibration \& temperature \\
        \hline
    	Piezoelectric & Generation of electric charge & Karlsruhe Univ. (Goger) & Experimental & PVDF-based 28-taxel flat tactile array \\  
        & & Sungkyunkwan University (Choi) & Experimental & PVDF 24-taxel skin for SKKU Hand-II fingertips \\
		\hline
        Optical-based & Mechanical light-intensity change & Opto-force, Hungary & Commercial & LED/photo-diode reflections on membrane \\
        & Image captured with CCD camera & Univ. Tokyo (Kamiyama) & Experimental & GelForce: Markers in transparent elastomer \\
        & & MIT \& GelSight Inc, US & Commercial & GelSight: Reflective membrane, colored LEDs \\
        & & Bristol Robotics Lab, UK & Experimental & TacTip: Markers on biomimetic pins \\
        \hline
	\end{tabular}}
	\caption{Representative commercial and experimental tactile sensors from circa. 2000-2010. Abbreviations: Quantum Tunneling Composite (QTC), Force Sensitive Resistor (FSR), Polyvinylidene Fluoride (PVDF), Micro-electromechanical Systems (MEMS), Organic Field-effect (OFE). {Original references in papers by~\cite{yousef_tactile_2011} and~\cite{kappassov_tactile_2015-1}}.}
	\label{table:X}
\end{table*}

%Overall, Lee and Nicholls' (1999) review is a broad and thorough survey of 1990s tactile sensing. Topics included cutaneous sensors ({\em e.g.,} artificial skin), fingers/grippers/multi-finger hands, soft materials, probes/whiskers, haptic/active perception, data processing and new applications. We refer the reader to their article for full coverage of these topics, and comment here only on applications.

%\footnote{Minimally invasive surgery is also called minimal access surgery (MAS), keyhole surgery, endoscopic surgery and laparoscopic surgery}

Even though they recognized that ``the predicted growth of applications in industrial automation has not eventuated,'' \cite{lee_review_1999} were optimistic about the future of tactile sensing. They saw promising new applications in three main areas: {\em medical procedures}, especially `keyhole' or minimal access surgery; {\em rehabilitation/health care} and service robotics; and {\em food processing and agriculture}. Likewise, \cite{lee_tactile_2000} saw that ``tactile sensing has undergone a major change of direction'' and that it will ``soon play a major role in unstructured environments'' for those same three application areas. For concreteness, he listed the main characteristics and needs to be satisfied by appropriate technological developments~(Table~\ref{tab:3}), concluding that ``we can look forward to the realization of this potential.''  

% \begin{figure}[b!]
%     \vspace{-1em}
% 	\centering
%     \begin{tabular}{@{}c@{}c@{}}
%     \small{\bf (a) MAS Procedure} & \small{\bf (b) Tactile system for MAS}\\
%     \includegraphics[width=0.38\columnwidth,trim=0 0 100 0,clip]{figures/eltaib_MAS_2.png} &
% 	\includegraphics[width=0.62\columnwidth,trim=0 0 0 0,clip]{figures/eltaib_MAS_1.png} 
%     \end{tabular}
%     \begin{tabular}{@{}c@{}c@{}}
%     \small{\hspace{-1.5em}\bf (c) Arthroscopic Procedure} & \small{\hspace{-1em}\bf (d)\,Tactile\,endoscopic\,grasper (2004)}\\
%     \includegraphics[width=0.5\columnwidth,trim=0 50 -10 0,clip]{figures/dargahi_MAS_2.png} &
% 	\includegraphics[width=0.5\columnwidth,trim=0 0 0 0,clip]{figures/dargahi_MAS_1.png} 
%     \end{tabular}
% 	\caption{Tactile sensing for minimal access surgery. (a) MAS procedure where a surgeon operates by slender instruments through small access wounds with a view from an endoscopic camera. (b) Proposed system to give surgeons a remote sense of touch~(from \cite{eltaib_tactile_2003}, Figs 1,2). (c)~Endoscopic image of a surgical grasping tool removing a loose body from a joint. (d) Prototype endoscopic grasper with piezoelectric tactile sensor~(from \cite{dargahi_human_2004}, Figs 3,5). } 
% 	\label{fig:10}
% \end{figure}

Interest followed in applications to medicine, inspiring focused reviews of ``Tactile sensing technology for minimal access surgery''~(\cite{eltaib_tactile_2003}) in {\em Mechatronics} and ``Human tactile perception as a standard for artificial tactile sensing''~(\cite{dargahi_human_2004}) in {\em Medical Robotics and Computer Assisted Surgery}. These were motivated by ``restoring a tactile capability to MAS \textcolor{black}{[minimal access surgery]} surgeons by artificial means would bring immense benefits in patient welfare and safety'' and ``tactile feedback to medical robotic systems [will help] approach the process demonstrated by bare-handed operations.'' The latter article also renewed interest in the biomimetics of artificial touch, in which it is still influencing new audiences~(Table~\ref{table:2}). (See Section~\ref{sec:foundations} for prior surveys of this topic.) 

Similar interest followed in applications to industry, inspiring two reviews in the journal {\em Industrial Robotics} on ``Tactile sensing in intelligent robotic manipulation''~(\cite{tegin_tactile_2005}) and ``Advances in tactile sensors design/manufacturing and its impact on robotics applications''~(\cite{dargahi_advances_2005}). The latter article extended their prior treatment to applications in farming, the food industry, service robotics, and the environment.

\begin{figure*}[t!]
    % \vspace{-1em}
    \resizebox{\textwidth}{!}{%
	\centering
    \begin{tabular}{@{}ccccccc@{}}
    {\bf (a)} & {\bf (b)} & {\bf (c)} & {\bf (d)} & {\bf (e)} & {\bf (f)} & {\bf (g)}\\
    \includegraphics[width=0.41\columnwidth,trim=0 -60 0 0,clip]{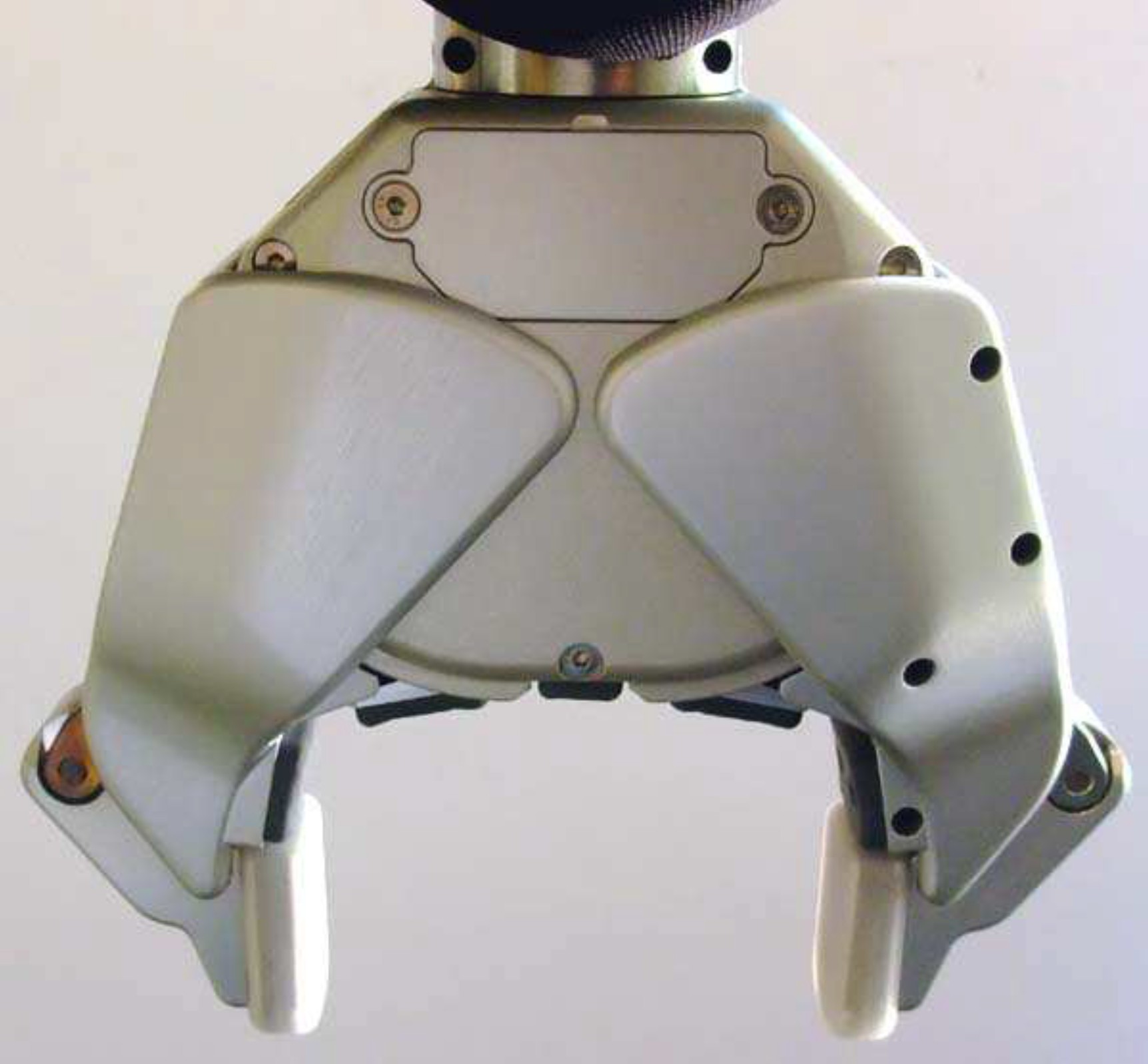} &
    \includegraphics[width=0.35\columnwidth,trim=25 0 170 140,clip]{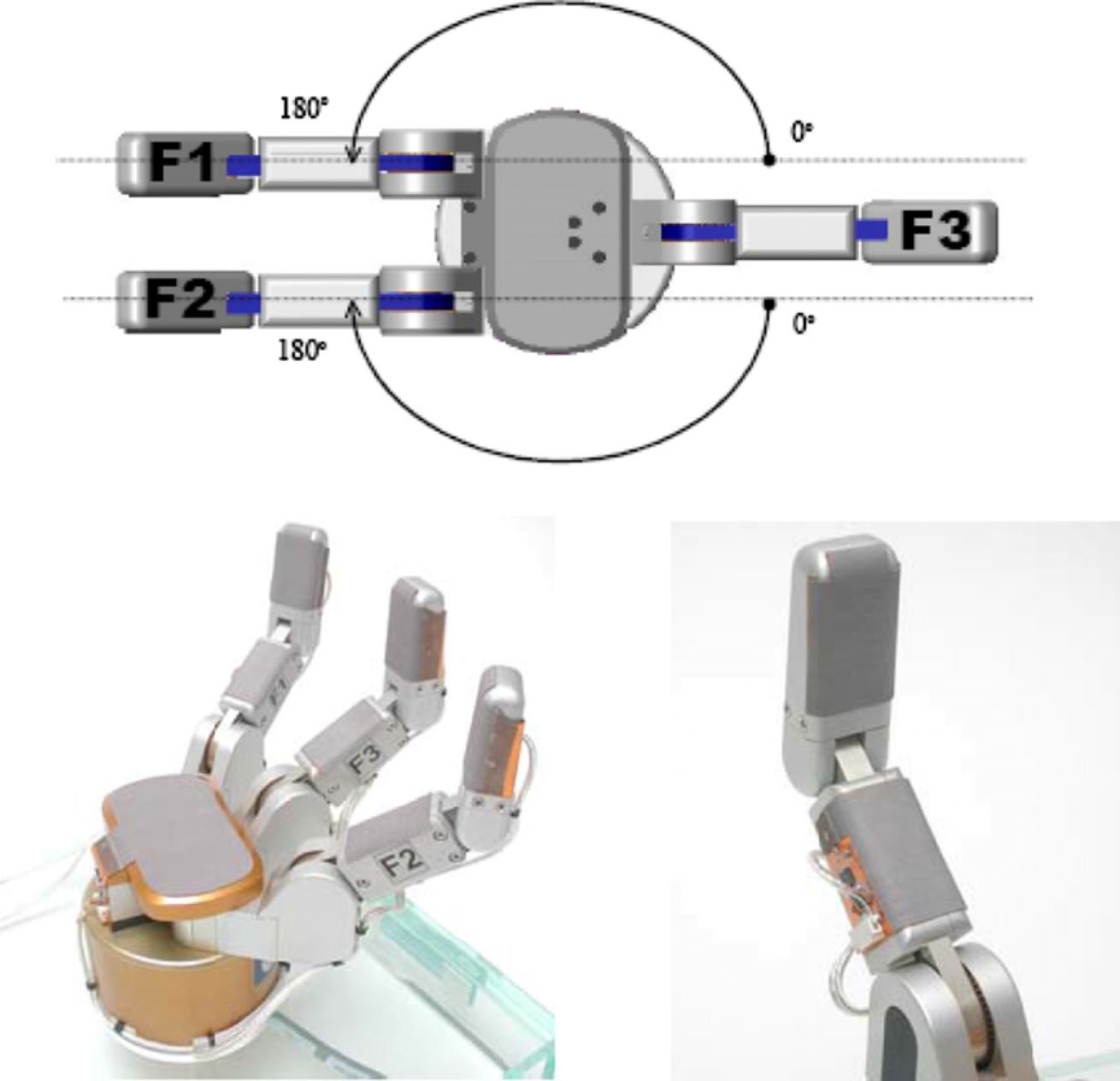} &
    \includegraphics[width=0.35\columnwidth,trim=120 20 0 0,clip]{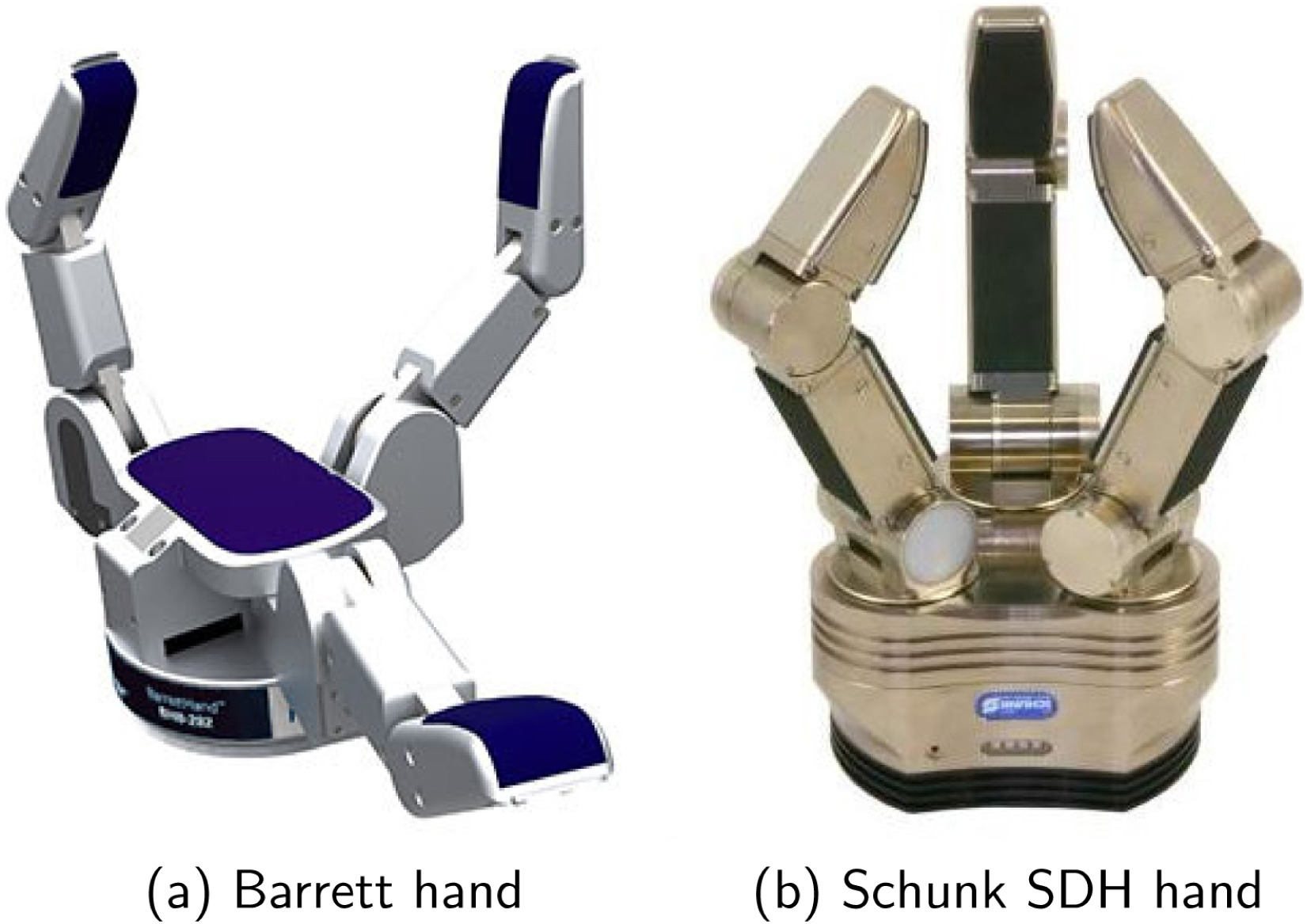} &
    \includegraphics[width=0.33\columnwidth,trim=0 0 40 0,clip]{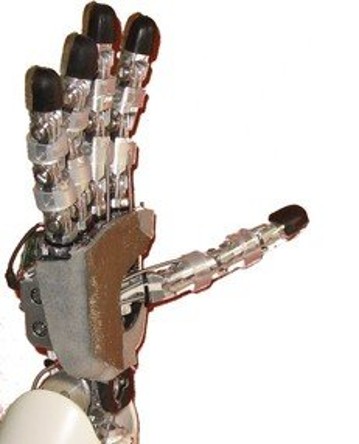} &
	\includegraphics[width=0.35\columnwidth,trim=170 0 0 0,clip]{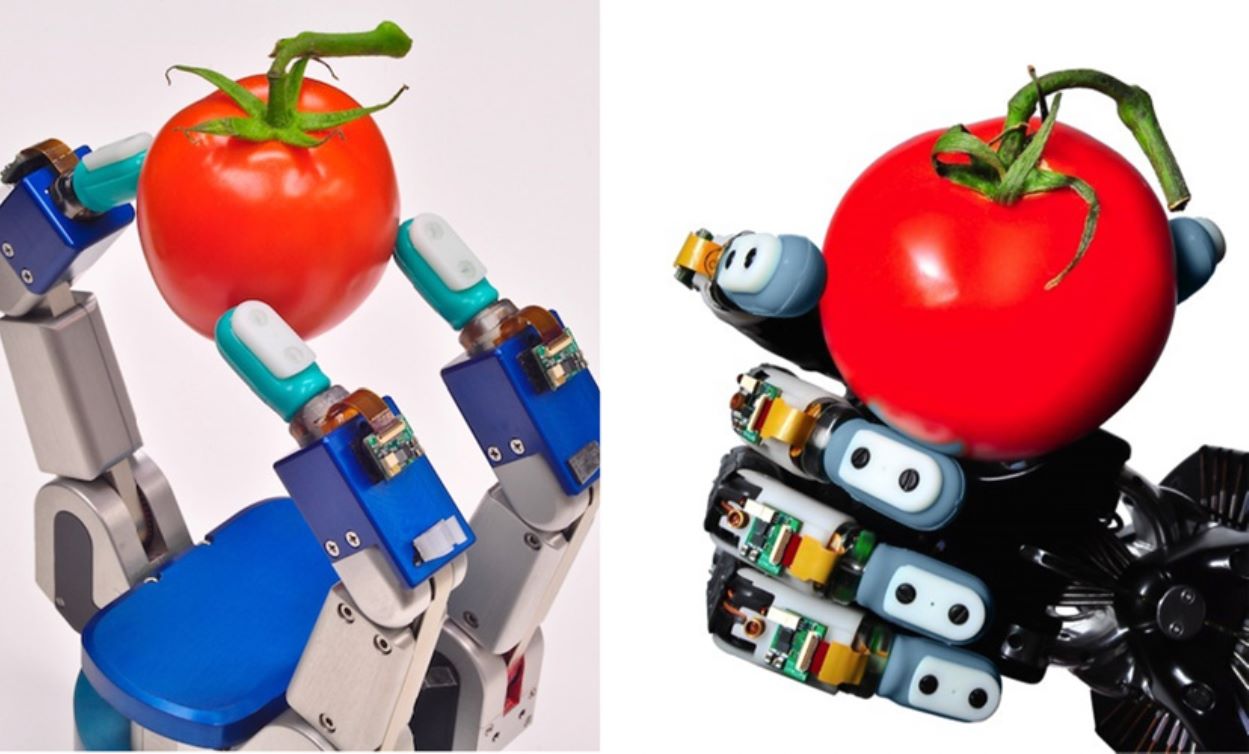} &
	\includegraphics[width=0.35\columnwidth,trim=2 0 166 0,clip]{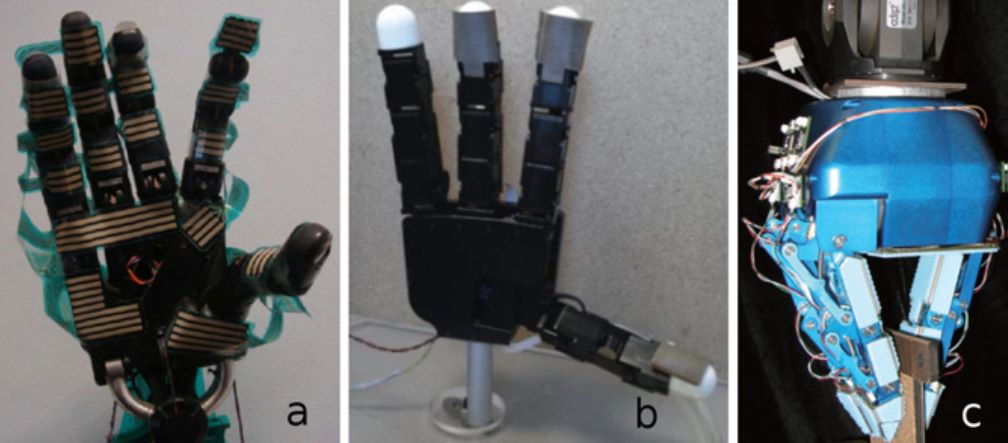} & 
    \includegraphics[width=0.38\columnwidth,trim=30 0 20 0,clip]{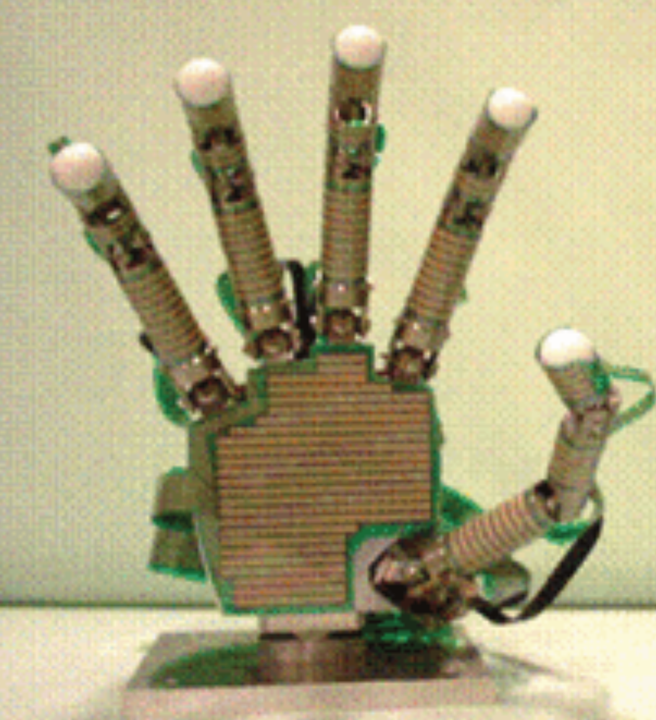} \\    
    {\bf (h)} & {\bf (i)} & {\bf (j)} & {\bf (k)} & {\bf (l)} & {\bf (m)} & {\bf (n)}\\
    \includegraphics[width=0.41\columnwidth,trim=380 20 10 40,clip]{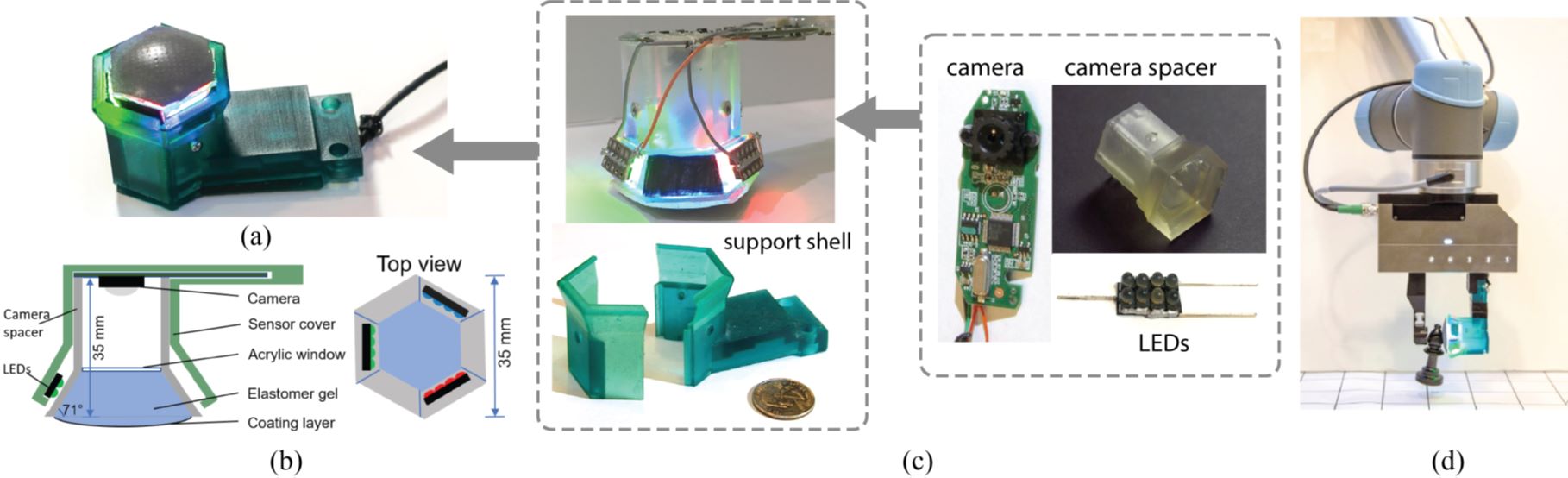} &
    \includegraphics[width=0.41\columnwidth,trim=10 10 10 10,clip]{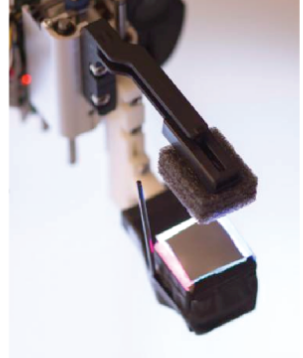} &
    \multicolumn{5}{c}{
    \includegraphics[width=2\columnwidth,trim=0 0 0 15,clip]{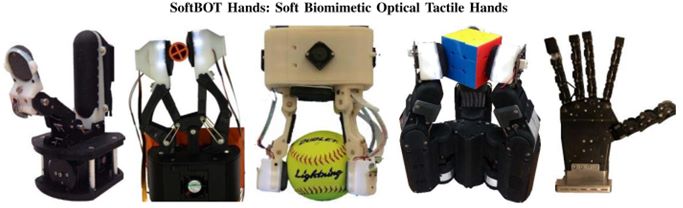}} \\    
    \end{tabular}} 
	\caption{Tactile-sensorized robot grippers and hands. Top row: examples circa. 2000-2010 from Table~\ref{table:X}. (a) PR2 gripper with 2 PPS RoboTouch 22-taxel fingertips (circa. 2009). (b)~Barratt hand with 3 PPS RoboTouch 22-taxel fingertips and 24-taxel phalanges, and a 24-taxel palm (circa. 2009). (c)~Schunk Dexterous Hand with 3 Weiss 78-taxel piezoresistive tactile fingertips and 84-taxel phalanges (circa. 2008). (d)~iCub humanoid robot hand with 5 capacitive 13-taxel fingertips and a 44-taxel palm (circa. 2010). (e) Shadow Hand with 5 SynTouch multimodal 19-taxel BioTac tactile fingertips. (f)~Shadow Dexterous Hand with TekScan 355-taxel FSR tactile skin (circa. 2010). (g)~Gifu hand III with 624-taxel FSR tactile skin (circa. 2002). Bottom row: examples with vision-based tactile sensors, circa. 2015-2020. (h,i) Grippers with a GelSight fingertip. (j) Model-M2 (2016), (k) Model-GR2 (2018), (l) Model-O (2020) Yale Open Hand grippers, (m) Shadow Modular Grasper (2019), and (n) Pisa/IIT SoftHand (2020) with 1,~2, 3, 3 and 5 TacTip fingertips. (Top row: images from~\citep{girao_tactile_2013} and~\citep{saudabayev_sensors_2015}; bottom row: images from~\citep{yuan_gelsight_2017-1} and~\citep{lepora_soft_2021-2}; the original references can be found in those papers.) See Figure~\ref{fig:15} for more examples.}
	\label{fig:12}
\end{figure*}
% (a,b) Fig 15, 14 of https://www.sciencedirect.com/science/article/abs/pii/S0263224112004368 Elsevier
% (c) Fig 5 of https://www.sciencedirect.com/science/article/pii/S0921889019300247
% (d) Fig 3 of https://ieeexplore.ieee.org/abstract/document/5686825 (I can email Alex for permission)
% (e) Fig 19 of https://www.sciencedirect.com/science/article/abs/pii/S0263224112004368 Elsevier
% (f) Fig 10 of https://www.sciencedirect.com/science/article/abs/pii/S0921889015001621 Elsevier
% (g) Fig 13 of https://ieeexplore.ieee.org/abstract/document/7283549
% (h,i) Fig 12d, 11e of https://www.mdpi.com/1424-8220/17/12/2762 free if under cc-by license
% (j-n) mine no permissions

\begin{figure}[b!]
    \vspace{-.5em}
	\centering
    \begin{tabular}{@{}c@{}}
    \small{\bf Intrinsic tactile sensing}\\
    \includegraphics[width=\columnwidth,trim=0 0 0 0,clip]{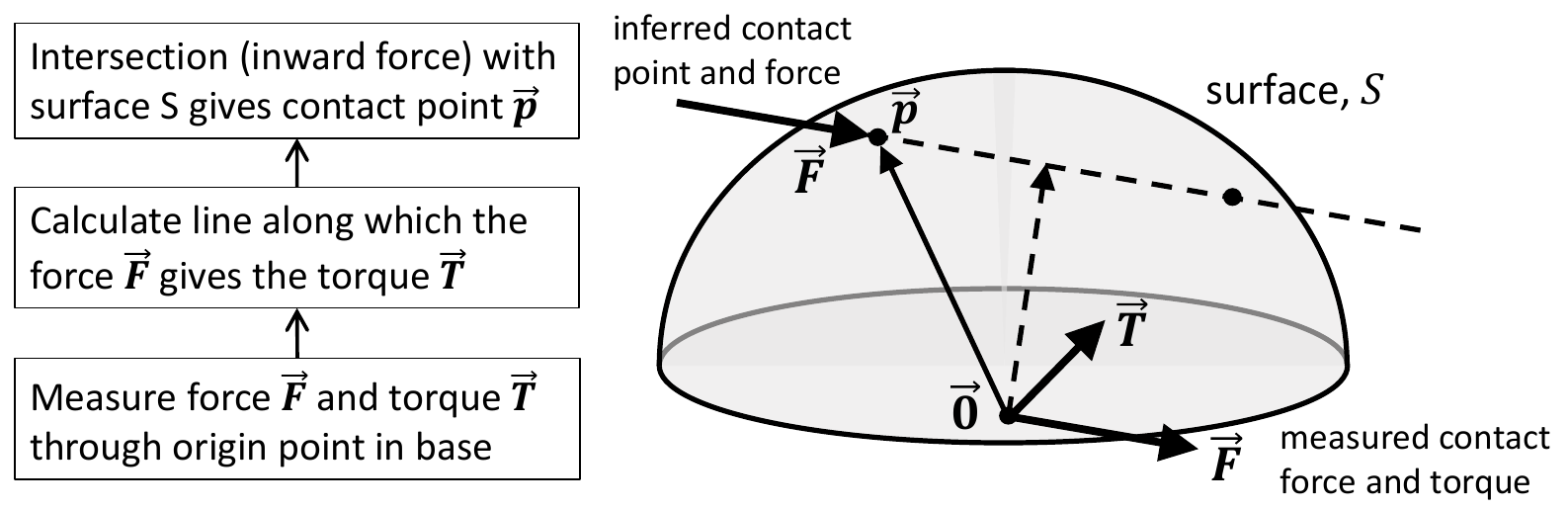} \\
    \small{\bf Extrinsic tactile sensing}\\
    \includegraphics[width=\columnwidth,trim=0 0 0 0,clip]{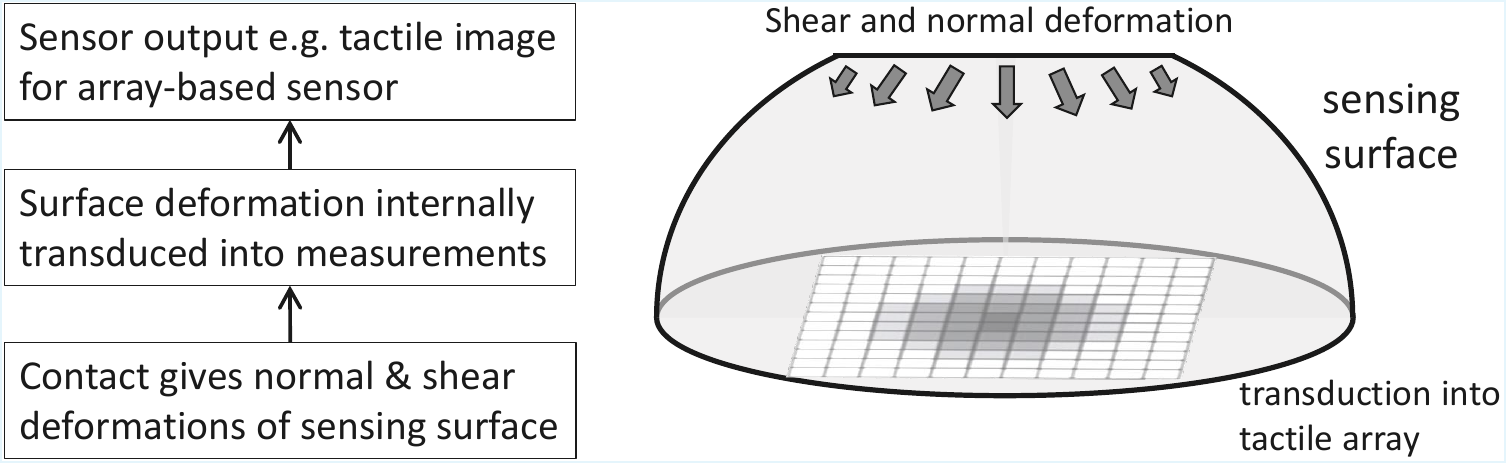} \\
    \end{tabular}
	\caption{Types of tactile sensing. Top: Intrinsic tactile sensing using force and torque measurements, assuming a known sensor surface shape (see~\cite[Figure~19.7]{cutkosky_force_2008}). Bottom: Extrinsic tactile sensing with a tactile array, where contact is transduced into an array output. .}
	\label{fig:10}
\end{figure}

More fundamentally, in this period, there was also progress in understanding the constitution of tactile robotics. The definition of artificial tactile sensing continued to be refined and improved. \cite{lee_tactile_2000} redefined: ``\textit{tactile sensing is a form of sensing that can measure given properties of an object through physical contact between the sensor and the object}'' (based on a similar statement by~\cite{lee_review_1999} a year earlier). Both authors rejected the overly narrow definitions given previously that just considered force sensing over an area, by accepting ``\textit{any property that is measured through contact, including the shape of an object, texture, temperature, hardness, moisture content, etc.}''%~(\cite{lee_review_1999}). % (and also omitting that a ``sensor is a device or system that...'')

\cite{lee_review_1999} also highlighted a theoretical issue for tactile sensing research, the {\em inverse tactile transduction problem}: it can be well-posed to (forward) model a tactile sensor output given an object shape and contact, but the inverse problem of inferring those low-dimensional object features from the high-dimensional tactile data can be poorly posed~(\cite{nowlin_experimental_1991}). Traditionally, this problem motivated the design of sensors with sharp tactile images that have minimal `cross-talk', analogous to CCD imaging arrays or touch-screens. However, such tactile arrays can be low-resolution (Table~\ref{table:X}) or too stiff to use as robotic skins. More recently, the inverse problem has been largely forgotten, since data-driven methods ({\em e.g.,} deep learning) can accurately model complex statistical relationships within tactile data.

Likewise, \cite{cutkosky_force_2008} clarified the foundations of tactile robotics in a thorough treatment of ``Force and tactile sensing.'' In particular, they clarified the relation between force, tactile, and touch sensing, which is still often confused and earlier definitions considered indirectly. They said that the most important quantities measured with touch sensing are {\em shape} and {\em force}, and tactile quantities are averages or spatial distributions over a contact area. Devices that measure an average or resultant quantity are {\em intrinsic tactile sensors} and are based on force sensing~(\cite{bicchi_augmentation_1989}). {\em Extrinsic tactile sensors} derive measurements from the contact interface, most~commonly using one of the many types of tactile array (illustrated in Figure~\ref{fig:10}).
%(as in contact sensing by \cite{lee_tactile_2000})

However, even though academic progress seemed slow in this middle generation of tactile robotics, the period resulted in many new commercial and experimental sensors. Some of these are collected in Table~\ref{table:X}, and are presented in a similar way to Table~\ref{table:2} that showed tactile sensors from the mid-1980s. Many of the commercial tactile sensors of the 2000s were integrated with robot hands, along with some from academic laboratories (see Figure~\ref{fig:12}). %Some later integrations of optical-based sensors in Table~\ref{table:X} are also shown.%, relating to advances in the most recent generation of tactile robotics. 

Comparing tactile sensors in successive generations (Tables~\ref{table:2} and \ref{table:X}), a consolidation occurred that resulted in many commercial products. Most technology types were the same, such as resistive, capacitive, and piezoelectric tactile sensors. Some types became more prevalent; {\em e.g.,} there were many FSR technologies and optical sensing began to become established. A few technology types were new, such as impedance-based tactile sensing. In particular, the BioTac would become influential for the first part of the the next generation of tactile robotics, but ultimately be discontinued by the end. Overall, this diversity of technology types and their compatibility with integrating into robot hands would help drive the expansion in the generation that followed.

\begin{figure*}[t!]
	\centering
    \begin{tabular}{@{}c@{}}
	\includegraphics[width=1.8\columnwidth,trim=0 2 0 0,clip]{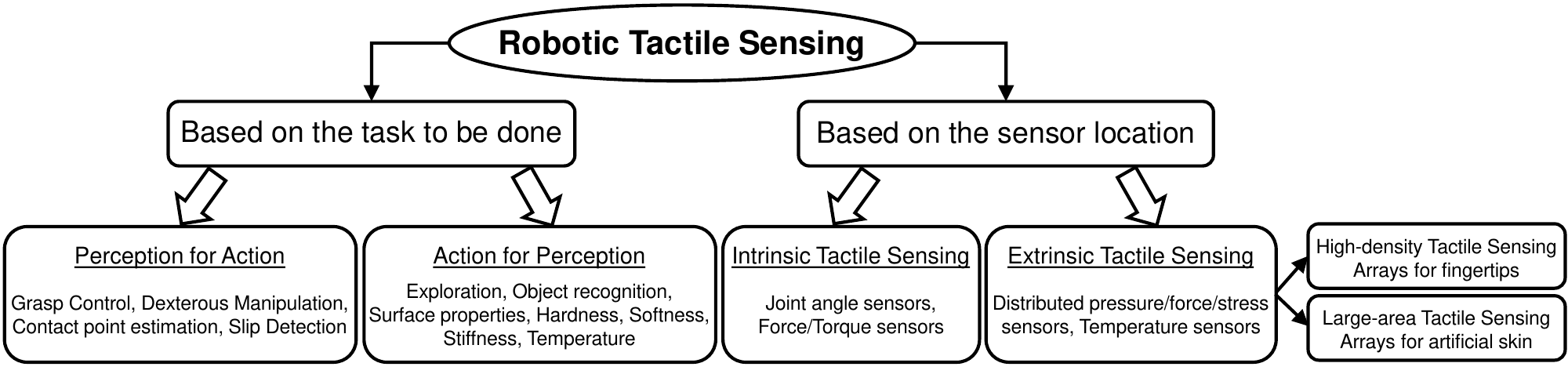} 
    \end{tabular}
    \vspace{-1em}
\caption{\cite{dahiya_tactile_2010} separated robotic tactile sensing by the tasks to be accomplished (left side) and the sensor location within the robot body (right side). {\em Perception for action} covers tasks where tactile sensing is used to control the end effector as it interacts with objects. {\em Action for perception} covers tasks where control is used to move the tactile sensor to explore or better recognize object properties. {\em Intrinsic tactile sensors} are placed within the mechanical structure (body) of the robot, {\em e.g.,} force sensors at the joints or fingertip base, and are akin to kinesthetic or proprioceptive sensing. {\em Extrinsic tactile sensors} are sited at or near the contact interface, {\em e.g.,} high-density tactile fingertips or large-area tactile skin, and are analogous to cutaneous sensing.}
\label{fig:11}
\end{figure*}
% based on Fig 2 of https://ieeexplore.ieee.org/abstract/document/5339133

%{\em Intrinsic} and {\em extrinsic} tactile sensors differ in whether the sensing elements are located within the body structure or near the contact interface: intrinsic tactile sensors are based on force sensing such as from the joints of the robot.

% . {\em Intrinsic sensors} are placed within the mechanical structure (body) of the robot, {\em e.g.,} force sensors at the joints or fingertip base, and are akin to kinesthetic sensing. {\em Extrinsic sensors} such as tactile arrays are sited at or near the contact interface and are analogous to cutaneous sensing (summarized in Figure~\ref{fig:11}).

\section{2010-24: Expansion and Diversification}
\label{sec:renewal}

The recent period from 2010 is characterized by a rapid expansion and diversification of tactile robotics. This change is indicated in part by the large increase in the number of reviews (on average 2--3 per year up to 2018, increasing to $>$5 per year more recently); also, there has been a large increase in the citation counts of articles~(Figure~\ref{fig:0}). Once again, this was against a growing background in robotics (with the largest robotics conference, {\it ICRA}, growing from around 2100 to 4000 submissions per year). However, the expansion of tactile robotics has been greater.

This generation began with several activities that helped cohere and accelerate tactile robotics research. The landmark review by \cite{dahiya_tactile_2010} on ``Tactile sensing -- From humans to humanoids'' authoritatively covered all tactile robotics of its time, from the human sense of touch to the role, importance, and state of tactile sensing in robotics (see Figure~\ref{fig:11} for their categorization of robotic tactile sensing). The authors were motivated by an ambition to provide tactile skins to cover the bodies of humanoids, which were gaining interest with the iCub robot~(\cite{metta_icub_2010}), such as for human-robot interaction~(\cite{argall_survey_2010}). These activities led to a ``2011 Special Issue on a Robotic Sense of Touch'' in {\em Transactions on Robotics}~(\cite{dahiya_guest_2011}), which also drew research attention to tactile robotics. 

\begin{figure}[b!]
	\centering
    \begin{tabular}{@{}c@{}}
    {\bf Bibliometric analysis of review articles on e-skin}\\
    \includegraphics[width=\columnwidth,trim=83 15 380 20,clip]{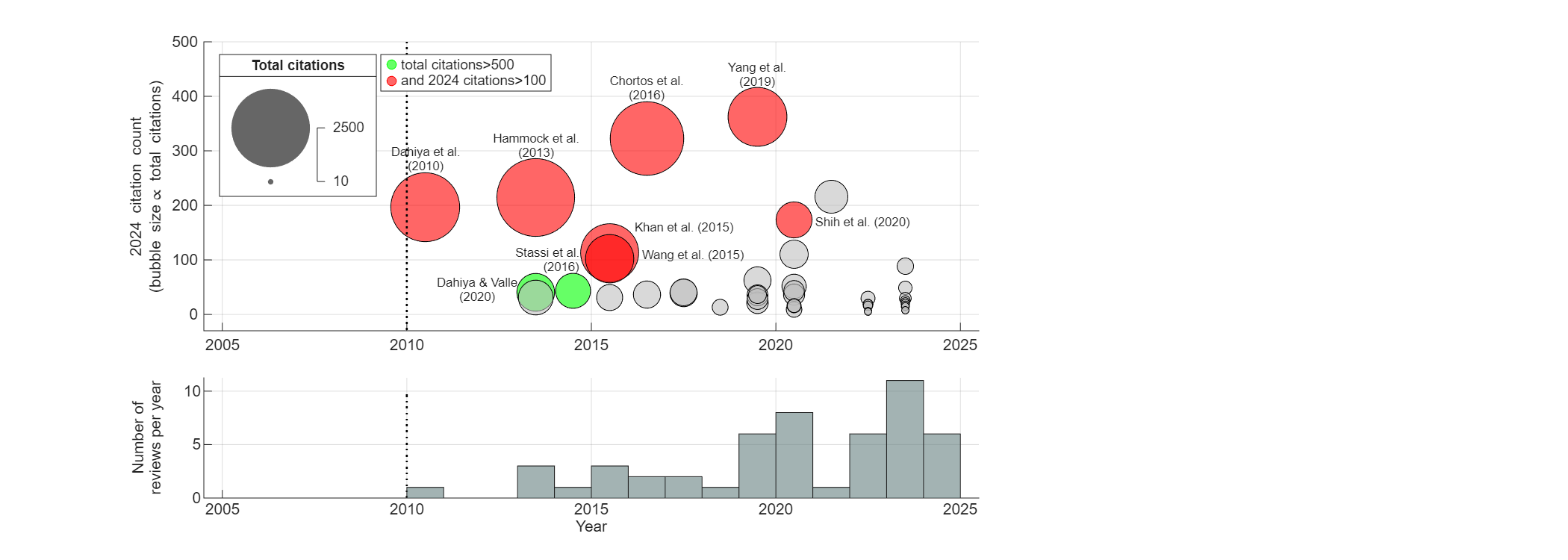} 
    \end{tabular}    
    \vspace{-.5em}
	\caption{Bubble plot for the citation counts of review papers in e-skins. The plot is organized similarly to Figure~\ref{fig:0}, based on a tabularized list of articles on e-skins (Table~\ref{tab:7}). }
	\label{fig:X}
\end{figure}

Since the review by~\cite{dahiya_tactile_2010} until 2024, there have been no other surveys of the entirety of tactile robotics. Instead, there have been influential treatments of themes within tactile robotics. Notable examples from Figures~\ref{fig:0} and \ref{fig:X} (Tables~\ref{tab:1}-\ref{tab:7}) include: ``Evolution of electronic skin (e-skin)''~(\cite{hammock_25th_2013}), ``Tactile sensing in dexterous robot hands''~(\cite{kappassov_tactile_2015-1}), ``GelSight: High-resolution tactile sensors for estimating geometry and force''~(\cite{yuan_gelsight_2017-1}), ``TacTip family: Soft optical tactile sensors with 3D-printed biomimetic morphologies''~(\cite{ward-cherrier_tactip_2018-2}), and ``Tactile Internet: Applications and challenges''~(\cite{fettweis_tactile_2014}).

%It was also the last review to authoritatively cover all the tactile robotics of its time, from the human sense of touch to the role, importance, and state of tactile sensing in robotics.% from the relation between human and robotic tactile sensing to the design of distributed sensor arrays.

The prevalent themes of those reviews appear to describe well the main areas of activity in tactile robotics over this generation: e-skins, tactile robot hands, vision-based tactile sensing, soft/biomimetic technologies, and the tactile Internet. In particular, these five themes usefully segregate the articles listed in Tables~\ref{tab:1}-\ref{tab:7}, and will form the basis for describing the expansion and diversification of this recent generation of tactile robotics.

\subsection{Separation of tactile fingertips and e-skins}
\label{sec:divergence}

An important and timely insight made by \cite{dahiya_tactile_2010} is: ``{\em extrinsic tactile sensing is further categorized in two ways -- first, for highly-sensitive parts ({\em e.g.,} fingertips), and second, for less-sensitive parts ({\em e.g.,} palm)}'' (see Figure~\ref{fig:11}, right). This separation into fingertips and artificial skins has proved insightful for the progress of tactile sensing since. %over the generation that followed. 

%In their review,~\cite{dahiya_tactile_2010} separated their coverage into ``tactile sensing arrays for parts like fingertips with high-density receptors''~(\cite{dahiya_tactile_2010}, Table I) and ``tactile sensing array for parts like large-area skin with low density of receptors''~(\cite{dahiya_tactile_2010}, Table II). 

%To some extent, it appears that the articles and book by~\cite{dahiya_tactile_2010,dahiya_robotic_2013,dahiya_directions_2013} on ``Tactile sensing -- From humans to humanoids,'' ``Directions toward effective utilization of tactile skin,'' and ``Robotic Tactile Sensing'' played a similar role to those by~\cite{harmon_touch-sensing_1980,harmon_sense_1981,harmon_automated_1982,harmon_tactile_1984} in initiating a research community.

During this recent generation, the development of tactile skins has gained a momentum that is distinct from other areas of tactile robotics. More than half of the research activity on tactile sensing has focused on artificial skins or e-skins, with a rapid expansion of the research in those areas~(Figure~\ref{fig:X}). Researchers in materials science have been attracted to tactile robotics, growing the field with publications in influential materials science journals such as {\em Advanced Materials} and {\em Nature Materials} (Table~\ref{tab:5}). This includes ``Evolution of electronic skin (e-skin)''~(\cite{hammock_25th_2013}), ``Pursuing prosthetic electronic skin''~(\cite{chortos_pursuing_2016}) and ``Electronic skin: Recent progress and future prospects...''~(\cite{yang_electronic_2019}).

This diversification of tactile sensing into materials science has resulted in two separate research communities that have distinct priorities and research methods. Although there is some overlap in publishing venues, such as {\em IEEE Sensors}, much of the traditional tactile robotics community has continued with robotics and engineering journals and conferences, while the artificial skin community is based within electronics and materials science. Accordingly, the present article splits contributions in tactile fingertips and e-skins between Figures~\ref{fig:0} and~\ref{fig:X} (Tables~\ref{tab:1} and \ref{tab:5}). Currently, judging by the number of citations, electronic skin research comprises the larger and more active community.  

In the next generation of tactile robotics, the two communities have the potential to achieve a far higher impact if they converge. For example, the potential benefits of \mbox{e-skins}, such as wide skin coverage and seamless integration onto the surface of robots, could be combined with the advances in dexterity from robot learning currently being developed for robotic hands with tactile fingertips.

%For completeness, the articles on artificial skins in Table~\ref{tab:5} are documented in the same way with citation counts ordered by year, to highlight influential reviews. %Currently, judging by the number of citations, the artificial skin community is the largest. 
%, which will be considered further in the latter sections of this review. 

%Meanwhile, reviews of tactile fingertips and related topics, such as tactile hands, have continued to be published in robotics journals such as {\em Transactions on Robotics}, {\em Robotics and Autonomous Systems} and {\em Mechatronics} (Table~\ref{tab:1}). Although there is some overlap in publishing venues, such as {\em IEEE Sensors}, it is clear that two separate communities have formed with different priorities and distinct research methods. 

\subsection{Tactile robotic hands}
\label{sec:development}

Much of the foundational research on tactile sensing was motivated by a desire to create human-like robot hands. For example, in their early review of ``Tactile sensing and the gripping challenge,''~\cite{dario_tactile_1985} concluded that ``at some point in the future, the solution may be an artificial hand with intelligent control, able to process and use tactile information adaptively.'' However, there was little research on multi-fingered tactile robotic hands until the 2000s, after which several commercial options became available (see Figure~\ref{fig:12}, top row). According to the article ``A century of robot hands'' (\cite{piazza_century_2019}), the decade after 2010 saw as many new robotic hands as the 90 years prior, and likewise interest in tactile hands has also increased.

%there is little coverage of tactile robot hands in any survey before 2011, with the most attention on tactile pinch grippers and robotic manipulation~(\cite{howe_touch_1992,howe_tactile_1993-1,cutkosky_force_2008}). %Likewise, there is little coverage of tactile sensing in surveys of robot hands. For example, the single mention in the seminal review of robotic grasping by \cite{cutkosky_grasp_1989} said that while ``there have been significant advances in control strategies and tactile sensing for hands. However, it seems that we are still a long way from building robots that can independently decide how to pick up and manipulate objects to accomplish everyday tasks.'' %Likewise, a later (2000) review of robot hands for dexterous manipulation and robust grasping by Bicchi says: ``many other important aspects could not be discussed, such as tactile sensing''~(\cite{bicchi_hands_2000}).

To show the variety of tactile robotic grippers and hands in use during that period, some photos of tactile sensing technologies collected in Table~\ref{table:X} are presented together in Figure~\ref{fig:12}. Interestingly, most hands (panels (a)-(e)) have dedicated tactile sensors for the fingertips, but some also have them in the phalanges (panels (b) and (c)) and/or palms (panels (c) and (d)). Two of the anthropomorphic hands are covered in tactile skin arrays (panels (f) and (g)). Impressively, the Gifu Hand shown in panel (g) had both the most tactile coverage and was the earliest (from 2002). 

This availability of new tactile robotic hand technologies led to adoption in many laboratories and growth in this research area. This interest was reflected in two influential surveys of dexterous tactile robot hands: ``Tactile sensing for dexterous in-hand manipulation in robotics''~(\cite{yousef_tactile_2011}) and ``Tactile sensing in dexterous robot hands'' (\cite{kappassov_tactile_2015-1}). The expanding activity in this area was reflected in other reviews of tactile hands, such as~(\cite{girao_tactile_2013,saudabayev_sensors_2015}). Since then, the focus on hands has changed to coverage within other topics, first in vision-based tactile sensing~(\cite{yuan_gelsight_2017-1,ward-cherrier_tactip_2018-2}), then soft robotics~(\cite{wang_toward_2018,subad_soft_2021,zhou_three-dimensional_2021,qu_recent_2023})

Reflecting on why progress in the use of tactile robotic hands has been challenging, \cite{yousef_tactile_2011} offered several insights. First, even the simplest tasks performable by humans are difficult to study in practice, {\em e.g.,} multi-finger in-hand manipulation. Consequently, a full appreciation from robotics or neuroscience of the requirements for tactile skins for hands is lacking. Second, tactile sensing technologies are limited in their force range, spatial/temporal resolution, sensing area, and shear force sensing. Therefore, robot hand control has instead relied on traditional kinematic models, complicated offline planning, and overuse of external sensors such as vision. These challenges need overcoming to lead to reactive manipulators that can handle objects with ease under uncertainty, modeling inaccuracies, and nonlinear dynamics. %and unpredictable object properties. 

\cite{kappassov_tactile_2015-1} expressed a complementary view that future research work in dexterous manipulation should focus on the investigation of autonomous control algorithms that apply tactile servoing and force control. Combined with tactile-based object recognition and grasp stabilization, this would allow robot hands to operate in real-world scenarios.

\begin{figure}[t!]
	\centering
    \includegraphics[width=\columnwidth,trim=30 0 30 0,clip]{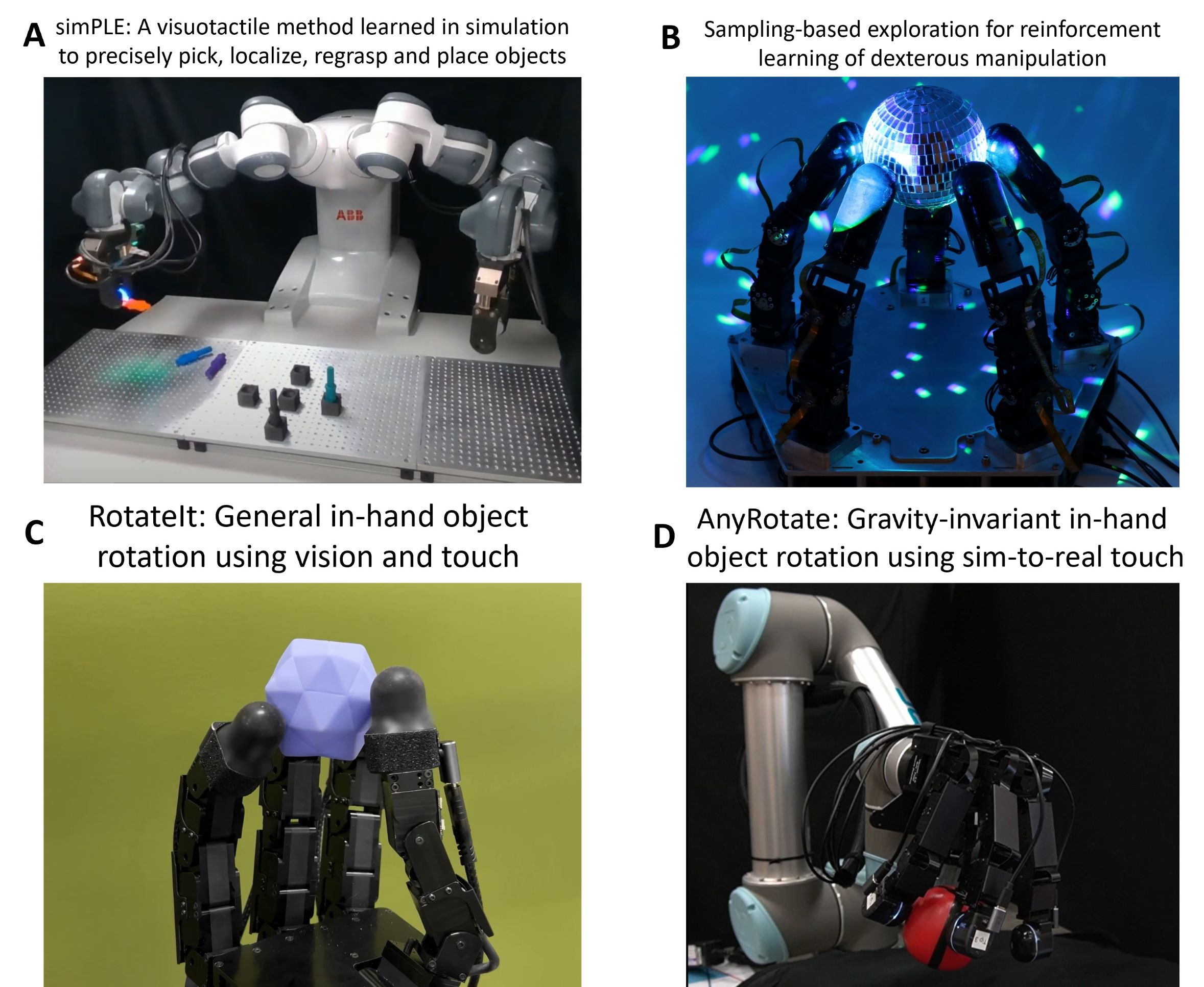} 
	\caption{State of the art in robot dexterity using sim-to-real methods with deep reinforcement learning applied to tactile data from optical tactile sensors. From~\cite{lepora_future_2024-1}, where references to the individual studies can be found.} 
    \vspace{-1em}
	\label{fig:15}
\end{figure}

Very recently, the field has undergone a step change with multiple demonstrations of in-hand object manipulation with multi-fingered robotic hands (Figure~\ref{fig:15}), as surveyed in a commentary in {\em Science Robotics}~(\cite{lepora_future_2024-1}). All of those hands used high-resolution tactile sensing in the fingertips, consistent with comments by~\cite{yousef_tactile_2011}. They also used variants of tactile control, trained via deep reinforcement learning in a physics simulation. By encountering many objects in simulation, they learn how to manipulate new objects while maintaining a stable grasp.

%(see Figure~\ref{fig:15}). That article pointed out that those examples used variants of optical or vision-based tactile sensing in the fingertips along with state-of-the-art AI methods such as sim-to-real deep reinforcement learning. 

% have permissions

% More recently, the focus of new review articles on tactile hands has changed toward technologies for soft tactile grippers and manipulators. Research in soft robotics has expanded greatly over the last decade, and its impact on tactile robotics is considered in Section~\ref{sec:soft}.

\subsection{Vision-based tactile sensing}
\label{sec:growth}

Vision-based tactile sensing has been a promising technology since the origins of tactile robotics. The first tactile sensing system was vision-based~(\cite{strickler_development_1966,strickler_design_1966-1}), using a fiber-optic bundle to transmit an internal view of a deformable skin surface to a TV camera (shown in Figure~\ref{fig:1}). By the mid-1980s, many different optoelectronic technologies for tactile sensing were being explored, including fiber optics, photo detector/LED pairs, and the integration of a high-resolution CCD camera within the fingertip of a gripper (see Figure~\ref{fig:6} and Table~\ref{table:2}). This latter technology, by Mott, Lee, and Nicholls from Aberystwyth, Wales, pioneered vision-based tactile sensing 30 years before it would become a main theme of tactile robotics~(\cite{mott_experimental_1984}).

In the mid-2010s, vision-based tactile sensing began to expand rapidly as a research field. For example, the first two dedicated reviews, on ``GelSight: High-resolution tactile sensors for estimating geometry and force''~(\cite{yuan_gelsight_2017-1}) and ``TacTip family: Soft optical tactile sensors with 3D-printed biomimetic morphologies''~(\cite{ward-cherrier_tactip_2018-2}), now count among the most cited reviews in that generation of tactile robotics (Table~\ref{tab:1}). Each of them focused on a specific type of vision-based tactile sensor design from the teams developing these technologies. The GelSight uses graded light-intensities, like~\cite{mott_experimental_1984}, whereas the TacTip uses discrete markers. Both types of design can be effective for robot dexterity: two examples of state-of-the-art manipulation in Figure~\ref{fig:15} use GelSight-type sensors (top and bottom left); another example uses TacTips (bottom right).

% For the first time, vs, not unlike~\cite{strickler_design_1966} but with an array of smaller discrete features

%Those initial reviews focused on specific types of vision-based tactile sensor: the GelSight~(\cite{yuan_gelsight_2017-1}) and the TacTip~(\cite{ward-cherrier_tactip_2018-2}), from the teams developing these technologies. Likewise, the focus of each review was distinguished by the properties of each sensor. The GelSight is reflection-based, and the camera image captures variations in light intensity to represent surface indentation, giving a focus on ``High-resolution tactile sensors for estimating geometry and force''~(Figure~\ref{fig:13}). The TacTip is marker-based, and the camera image captures the motion of discrete marker-tipped pins analogous to human skin structure, giving a focus on ``Soft optical tactile sensors with 3D-printed biomimetic morphologies''~(Figure~\ref{fig:14}). 

%These reviews categorized vision-based tactile sensors according to how they transduce contact into tactile images captured by an internal camera, with two main categories: {\em marker-based}, where changes in discrete markers such as displacement, size or shape indicate aspects of skin deformation; and {\em reflective membrane-based}, where the surface gradient of reflected light maps skin indentation. 

As interest in these technologies grew, there followed many reviews of vision-based tactile sensing~({\em e.g.,} \cite{shimonomura_tactile_2019,abad_visuotactile_2020,shah_design_2021,lepora_soft_2021-2, zhang_hardware_2022,li_marker_2023}) that covered a diversity of vision-based tactile sensor variants. Examples include ChromoTouch, DIGIT, DigiTac, FingerVision, F-Touch, GelForce, GelTip, GelSlim, MultiTip, NeuroTac, OmniTact, Soft-bubble, and Tac3D (references in the reviews). {All of these tactile sensors use camera images to represent contact information~(\cite{li_classification_2025}) in one of two ways:} {\em intensity-based}, where continuous variations in light intensity show the skin indentation, or {\em marker-based}, which images discrete features coupled to the skin deformation. These mechanisms can be further subdivided: intensity-based may use reflected light (as in the GelSight) or refracted light (\cite{mott_experimental_1984}); marker-based may use simple markers inside the skin, or morphological structures that transform the skin deformation (as in the TacTip). Other options ({\em e.g.,} transparency) and combinations also exist.

\begin{table*}[b!]
\resizebox{\textwidth}{!}{%
	\renewcommand{\arraystretch}{1}
	\centering
	\begin{tabular}{@{}ccc@{}} 
		{\bf Neurophysiology} & {\bf Function} & {\bf Biomimetic counterpart} \\
		\hline
		epidermal ridges and dermal papillae & transmits and amplifies deformation of surface to mechanoreceptors & morphological structures, {\em e.g.,} TacTip pins \\
		reticular dermis and subcutaneous fat & soft structure and compliance & soft elastomeric layer beneath outer skin \\
		SA-I mechanoreceptors (Merkel cells) & sense sustained skin deformation; perception of shape and edges & taxel or marker displacements \\ 
        SA-II mechanoreceptors (Ruffel) & sense sustained skin stretch; related to proprioception & not typically considered \\ 
		RA-I mechanoreceptors (Meissner corpuscles) & sense transient skin movement; perception of flutter and surface slip & taxel changes or marker velocities \\
		RA-II mechanoreceptors (Pacinian corpuscles) & vibration sensing; perception of surface texture & embedded microphones or vibration detectors  \\
		nociceptors (free nerve endings) & noxious touch & not typically considered \\
		thermoceptors (free nerve endings) & temperature difference sensing & thermal sensors; thermochromatic paint (vision-based)  \\
		overlapping sensitive receptive fields & hyperacuity; spatial sensitivity & super-resolution; spatial representations \\
	    epidermal ridges (fingerprint) & friction and improved transduction; induces incipient slip & papillary array or artificial fingerprint \\
		neural spiking & efficient signal encoding & event-based transduction  \\
	\end{tabular}}
	\caption{Some aspects of human skin physiology that have biomimetic tactile sensor counterparts (see~\cite{lepora_soft_2021-2}, Table 1).}
	\label{table}
\end{table*}

Why this growing interest? At heart, it is because vision-based tactile sensing is proving to be effective~in~enabling robot dexterity; {\em e.g.,} in-hand manipulation~(Figure~\ref{fig:15}). This appears due to an alignment of several~technologies: (1)~miniature digital cameras, first from webcams for cheap and easy construction of test sensors, then smartphone cameras have led to smaller designs suitable for fingertips; (2)~3D-printers, first for making the sensor body and molds for skin, then with multi-material printing for rapid design and fabrication of the skin and other non-electronic~parts; (3)~neural network software libraries, which can be used to extract useful information from high-resolution images, specifically contact-related features from tactile images.  

Tactile robotics is still in a phase where these constituent technologies are improving, so vision-based tactile sensing will also continue to improve ({\em e.g.,} use of multiple cameras). However, there is potential to make further progress by being integrative with different themes of tactile robotics, such as with complementary aspects of e-skin technology.

% (examples in Figure~\ref{fig:12}, bottom row). 

%A third category of {\em light-conductive plate} was introduced, where markers become visible when pressed against a plate. this mechanism could  which could be considered as a sub-category of a marker-based principle. 

\subsection{Soft and biomimetic technologies}
\label{sec:soft}

%  % As such, it eschews traditional `hard' industrial robotics with single-axis joints and stiff links. 
% Since the late 2000s, the relatively new field of `soft robotics' has emphasized the capabilities arising from novel material properties and body morphologies~(\cite{pfeifer_self-organization_2007,trivedi_soft_2008,kim_soft_2013}). Soft robotics often draws on nature for inspiration, given the vast diversity of evolved body structures to mimic, from trunks and tentacles to paws and hands. This leads to a natural synergy between soft and tactile robotics, as skin is a soft biological structure. 

Soft and biomimetic tactile sensing has a long history of inspiring robotic touch, dating back to the beginnings of the research field. \cite{harmon_tactile_1984} first gave insight into how human touch may guide robot touch, spanning from mechanoreceptor transduction to active control of touch, from the perspective that ``natural systems provide existence proofs of engineering success.'' This theme of taking inspiration from the biology of human skin to design artificial sensors has continued over the history of tactile robotics~(\cite{dario_tactile_1985,jayawant_tactile_1989,howe_dynamic_1993,lee_review_1999,dargahi_human_2004,cutkosky_force_2008,dahiya_tactile_2010,yousef_tactile_2011}). Meanwhile, soft robotics often draws on nature for inspiration of novel material properties and body morphologies~(\cite{pfeifer_self-organization_2007,kim_soft_2013}). This leads to a natural synergy between soft and tactile robotics, as skin is a soft biological structure. 

%Soon after, \cite{dario_tactile_1985} also said that researchers ``tentatively are using the human skin as a model'' to specify artificial transduction performance. They considered piezoelectric tactile sensors as biomimetic models of human skin receptors (see Section~\ref{sec:foundations}, Figure~\ref{fig:7}).
% how the structure of tactile bumps around a crocodile's mouth may impart sensitivity to small pressure changes, and

A core idea guiding soft robotics research is to embody morphological computation~(\cite{pfeifer_morphological_2009}), or ``offloading computation from the brain to the body"~(\cite{muller_what_2017}). In their article ``Morphological computation in haptic sensation and interaction,''~\cite{bernth_morphological_2018} considered examples from nature, including that the wrinkling of human fingertips when wet may help feel vibrations indicating slip. Aspects of skin morphology were also central to ``Soft optical tactile sensors with 3D-printed biomimetic
morphologies''~(\cite{ward-cherrier_tactip_2018-2}), where the TacTip design mimics dermal papillae and intermediate ridge structures in human skin to act as mechanical amplifiers of surface deformation (see Figure~\ref{fig:9A} for human skin). Various other biomimetic counterparts have been considered in artificial tactile sensors (Table~\ref{table}).  

Embodied intelligence combines aspects of intelligence from morphology with those from traditional computation. The physical interaction between a robot and the environment is the central consideration~(\cite{liu_embodied_2020,li_review_2020}), such as for tactile object recognition~(\cite{luo_robotic_2017,liu_recent_2017}). These topics and their origins in active touch are understudied but core to a future AI of touch. %Active touch can also be co, which has been revisited recently relates back to   

This combination of the computational and morphological aspects of tactile sensing is central to many works. In ``Tactile sensors for friction estimation and incipient slip detection'',~\cite{chen_tactile_2018} considered how the tactile skin surface facilitates grip control. Human fingertips have papillary ridges (fingerprints), one function of which is to help signal a loss of grip before an object slips. This understanding has inspired the design of some tactile sensors, such as the PapillArray covered in that review. Generally, there has been a sustained interest in tactile slip detection, being fundamental to secure grasping and manipulation~(\cite{francomano_artificial_2013}).

 %\cite{lucarotti_synthetic_2013} reviewed how this biomimicry could become closer to biology by creating bio-artificial tactile skins, such as bio-hybrid and fully-biological tactile systems for neuroprosthetic and medical applications. Another way to enhance biomimicy is to adopt neural coding methods using trains of discrete `spike' events whose timing encodes tactile information, which \cite{yi_biomimetic_2018} reviewed along with the use of models of mechanoreceptor transduction for artificial tactile processing. The benefits of neuromorphic computing for tactile sensing were also reviewed by~\cite{liu_embodied_2020}, along with the importance of embodiment for `brain-like AI'. 

% Interest has also grown in soft dexterous robots, reflected in influential reviews ``Soft manipulators and grippers'' and  ``Soft robotic grippers''~(\cite{hughes_soft_2016,shintake_soft_2018}). 

In considering ``perceptive soft
robots,''~\cite{wang_toward_2018}, observed that ``a variety of soft sensing technologies are available today, but there is still a gap to effectively utilize them in soft robots for practical applications.'' Furthermore, proprioception is far more challenging for soft than than `hard' robots, ``because they have almost infinite degrees of freedom and can be deformed by both internal driving and external loads.'' Likewise, those challenges are magnified for tactile sensing when the sensor must also be a flexible outer skin for the robot. Consequently, there has been interest in tactile sensors suitable for the outer structure of a soft body, such as piezoresistive, capacitive, piezoelectric, and triboelectric sensing~(\cite{subad_soft_2021,qu_recent_2023}), with an emphasis on 3D-printing~(\cite{zhou_three-dimensional_2021}). These areas are among the most active in tactile robotics.

\subsection{Towards a tactile Internet}
\label{sec:towards}

A new theme in tactile robotics that emerged in the mid-2010s is to envision a tactile Internet: a telecommunications network that connects tactile robots and human operators to relay physical touch experiences and human manipulation skills. In essence, this idea dates back to telepresence and its applications that \cite{minsky_telepresence_1980} popularized (see Section~\ref{sec:beginnings}). However, a prospective tactile Internet opens up many new applications through modern telecommunications ({\em e.g.,}, telemedicine, avatars) and challenges ({\em e.g.,} data security).

The term ``Tactile Internet'' was introduced by \cite{fettweis_tactile_2014} to describe mobile applications that become viable when round-trip latencies drop below 1\,ms, since ``this is the typical interaction latency required for tactile steering and control of real and virtual objects without creating cybersickness''. Because data rates for wireless technologies follow Moore's law of doubling every 18 months, he predicted in 2014 that this progress will soon ``revolutionize education, mobility and traffic, health care, sports, entertainment, gaming and the smart grid''. 5G networks with latencies of 1\,ms have been widely available since 2020, making this now a reality.

In the past decade, there have been many surveys with overlapping content on technical requirements for the tactile Internet and its applications~({\em e.g.,} \cite{maier_tactile_2016,simsek_5g-enabled_2016,aijaz_realizing_2017,dohler_internet_2017,antonakoglou_toward_2018,gupta_tactile_2019}; see Table~\ref{tab:7} for a list). One trend is that the number of new review articles has dropped since 2020, when the citations per year of the early articles peaked, in contrast to other areas of tactile robotics. This may be because 5G is a mature technology and the bottlenecks for a tactile Internet are now a need for improved haptics and tactile robotics.  

This need for improved technologies was recognized by \cite{haddadin_tactile_2019}, who considered ``Tactile robots as a central embodiment of the tactile Internet''. In their view, the tactile robot will constitute the next generation of robots with capabilities beyond those of the current generation whose active compliance stems from knowledge of their kinematics and intrinsic joint forces/torques. However, some `missing technologies' are needed for that new generation: (1)~an affordable, easy-to-use artificial skin that enables tactile sensing for robotic platforms; (2)~wearables that are intuitive to use and seamlessly connect to immerse human use in the tactile avatar; and (3)~exoskeletons as human system interfaces that provide realistic sensation of forces or contact and proprioceptive information. All of these areas are being actively investigated, but the research has not yet escaped the preliminary stage and lacks commercial availability.

% \clearpage
\section{2025-: The future of tactile robotics}
\label{sec:future}

Like societal change, tactile robotics has cycled through a succession of historical generations, here referred to as: 1965-79 (origins), 1980-94  (foundations and growth), 1995-2009 (tactile winter) and 2010-2024 (expansion and diversification). This cycle suggests that tactile robotics is entering a new generation, raising the question: What is the future of tactile robotics from 2025 onward?

Presented below are some views of the future separated into: (i) near-horizon trends that may continue along somewhat predictable paths from recent history; and (ii)~disruptive changes that may unfold unpredictably, where the entire history of the field informs what may occur. Near-horizon trends will be extrapolated from the five emergent themes in \mbox{2010-24} surveyed in Sections~\ref{sec:divergence}-\ref{sec:towards}. Some potentially disruptive changes will then be considered, such as the impact of artificial intelligence on tactile robotics. 

{\em 1) Tactile e-skins:} As research diversified in the recent generation of tactile robotics, an e-skin community has formed around progress in materials science and electronics. Currently, the e-skin community is more active than the original tactile robotics community~({\em c.f.,} Tables \ref{tab:1} and \ref{tab:5}), which will continue for the near future. New skin-like tactile arrays will have novel capabilities, such as low- or self-powered integrated sensing and computation, and novel sensors fabricated with advanced materials such as graphene. %As e-skin capabilities improve and robot manipulators such as dexterous robot hands become more widely used, the e-skin community should engage more with the AI methods now being developed for tactile dexterity with bulkier fingertip-like sensors. %In one scenario, separate tactile fingertip development could become obsolete if the requisite sensing capabilities for human-like manipulation skills are readily available in \mbox{e-skins}.

{\em 2) Tactile robot hands:} Many new grippers and multi-fingered robot hands have emerged in the past decade, and it is now routine to integrate tactile sensors into their fingertips ({\em e.g.,} Figure~\ref{fig:12}). For a long time, a main goal for the field was to reach human-like capabilities, such as in-hand and bi-manual dexterous manipulation. Very recently, the field has undergone a step change with multiple demonstrations of such capabilities (see Figure~\ref{fig:15}), using state-of-the-art AI methods such as sim-to-real deep reinforcement learning applied to tactile robotics. In the future, this dexterity will become routine and other examples of in-hand dexterity will follow, such as tool use or manual assembly.%, bringing skills that can currently only be done by humans within the reach of robots.

{\em 3) Vision-based tactile sensing:} Progress has recently accelerated with the rapid miniaturization and shrinking cost of camera technology and the translation of AI software for computer vision to tactile robotics for dexterous manipulation (Figure~\ref{fig:15}). There will be continued exploration and improvement of the design and fabrication of suitable tactile contact surfaces for imaging. Furthermore, future advances that ease the manufacture and adoption of vision-based tactile sensing will encourage commercialization and transfer to industry. In the longer term, the technology will become ever smaller and easier to integrate, {\em e.g.,} extending from fingertips to the phalanges and palms of robot hands.

{\em 4) Soft and biomimetic tactile robots:} There is a natural and established synergy between tactile sensing, biomimetics, and soft~robotics. Areas of activity include tactile skins designed to instantiate biologically-inspired processing that simplifies downstream perception and control, and attempts to combine tactile sensing with soft robots. In the future, these topics will influence novel e-skins, tactile robot hands, and vision-based tactile sensors. However, major progress is needed to utilize soft robots for practical applications, given its preliminary nature in many areas. Specific aspects such as morphological computing and the use of compliance can be more readily applied to conventional `hard' robots with soft tactile components.%, as significant challenges remain to have soft sensing and actuation suitable for state-of-the-art robot dexterity. %That said, there is huge potential for soft tactile robots, and specific aspects such as morphological computing can be more readily applied to conventional `hard' robots with soft tactile components.

{\em 5) Tactile Internet:} Excitement about a future tactile Internet stemmed from anticipation of telecommunication latencies dropping below a 1\,ms threshold for remote touch. That threshold was passed with 5G starting in 2020, which is now a mature technology. Therefore, the key enabling technologies needed for a tactile Internet are now to have consumer-ready tactile robots and haptic feedback devices. The next generation of tactile robotics could finally reach this maturity to span applications envisaged in industrial robotics (\cite{harmon_automated_1982}) and telepresence (\cite{minsky_telepresence_1980}). %and more recently with the tactile Internet      

However, the above predictions do not say what will truly happen in the next generation of tactile robotics. They only say how this current generation will continue. What could be reasonably expected to be a retrospective on tactile robotics in fifteen years' time, in 2040? 

The 2010s saw unprecedented progress in AI, particularly in computer vision, extending to generative AI and language models in the 2020s. Underpinning that progress has been ever-increasing compute power, which has both become more accessible and an energy burden. In other technologies, automated fabrication has steadily improved, with affordable high-quality 3D-printing, high-end metal printing, and additively-manufactured electronics. The life sciences have advanced in gene-editing and synthetic biology, although they are unlikely to affect tactile robotics apart from being a use case in automated labs. Virtual reality is now an established commercial product that should drive progress in affordable haptic glove technology. Telecommunications continues to advance through 5G to 6G, offering the tactile Internet if haptics and tactile robotics can come together.  

How will technological progress affect tactile robotics? Certainly, no one expects another tactile winter, and we are long past the early optimism of the 1980s. Probably, the most impact is to come from {\em synthesis}: combining domains of technology to create something new. Currently, tactile robotics is fragmented. Artificial skin researchers may look on vision-based tactile sensors as bulky and impractical, while those applying AI to vision-based tactile sensing see years of unmet promises to transform robot dexterity with other sensor types. Likewise, the soft robotics community may eschew traditional electromechanical robots to focus on early-stage sensor and actuator research, while those building highly-dexterous robot hands see an imminent deployment of their technologies. 

The most progress in robot dexterity may come from unifying the diverse themes of tactile robotics. To some extent, this is happening, as vision-based tactile sensing and robotic hand research combine for in-hand and bimanual dexterity~(Figure~\ref{fig:15}). There are other areas of synthesis, such as combining e-skins over large surfaces with vision-based tactile fingertips, or deploying the AI methods used for in-hand and bimanual robot dexterity on soft robots or with artificial skins. Other key areas of synthesis exist, and moving out of siloed approaches will be important.  %The next generation of tactile robotics should bring these and other domains of knowledge together. %sOther less obvious combinations no doubt exist, and identifying those could yield the most interesting and important research.

% In addition to the above near-term themes, some current developments have the potential for disruptive impact on tactile robotics, yet were not covered in dedicated review articles. Of these, two emerging themes are tactile AI and humanoid robots, which are discussed next.

%In particular, robot learning may be necessary for real-world manipulation in varied environments~(\cite{kroemer_review_2021}). 
%~\cite{sunderhauf_limits_2018} treated ``active manipulation as an extension of active vision'' (their Table~2) and

Clearly, advances in AI have the potential for major disruptive impact on all tactile robotics. However, tactile sensing has been a blind spot for AI research. In assessing the ``Potentials and limits for deep learning in robotics,''~\cite{sunderhauf_limits_2018} did not mention tactile sensing. OpenAI's influential work on in-hand manipulation~(\cite{andrychowicz_learning_2020}) used about 20 cameras for a task that would seem more suited to touch. It then took several years for university research laboratories to surpass that dexterity using tactile sensing (see Figure~\ref{fig:15}). Tactile robotics is still a small community. It would be transformative if even a fraction of the community applying AI to vision could change focus to touch. Perhaps it is fanciful, but touch could become as central to future AI research as sound/speech and vision are now. Rather than vague claims about `Artificial General Intelligence', an AI of tactile robotics could give tangible insight into human intelligence through understanding how our nervous systems are wired for dexterity.  

Finally, humanoids will clearly have an important role in the future of tactile robotics. Currently, in the mid-2020s, many tens of billions of dollars are being invested in their development. An output of this investment is a stream of videos showing sleek-looking androids walking or dancing on stage and in other human-centric environments. This is very impressive, but apart from some side applications in entertainment or education, the reason for a humanoid robot is to be a labor-saving device. To perform human labor, the humanoid must have useful hands that function with near-human dexterity. As we have seen, that goal remains some way off and the technology may be harder to develop than a walking robot. That said, a humanoids is well-suited for transporting a pair of dexterous tactile robot hands to where they are needed in human environments. When tactile robotics results in robots with human-like dexterity, humanoids could become commonplace. In the meantime, though, they may not be particularly useful in our daily lives.

To conclude, tactile robotics is entering a new generation where many of its technologies may soon mature to widespread commercial use. Although tactile sensing has been a relatively small and somewhat niche field for much of its history, there has always been a sense that it could have a transformative impact on all robotics. Important applications include those that require autonomous human-like dexterity and immersive teleoperation, spanning a huge range of possible uses from manufacturing and assembly to energy production and health care. As these applications mature to commercial reality in the next generation, they will potentially have a major disruptive impact across all robotics and AI. This could drive rapid and unprecedented progress in tactile robotics as it becomes a pivotal technology for many societally and economically important sectors.

\begin{acks}
This work was supported by an award from the `Robot dexterity' program at the Advanced Research + Innovation Agency (ARIA).
\end{acks}

\vfill
\pagebreak

\appendix

\begin{table*}[p]
\resizebox{\textwidth}{!}{%
	\renewcommand{\arraystretch}{1}
	\centering
	\begin{tabular}{@{}l@{}c@{}c@{}c@{}c@{}}	
		\multirow{2}*{\textbf{Year}} & \textbf{Title} & \textbf{Journal} & \textbf{Citations} & \multirow{2}*{\textbf{Reference}} \\
		 & \textbf{(cropped when necessary)} & \textbf{(abbreviated)}  & \textbf{to 2024 (in 2024)} &  \\
		\hline
        \multicolumn{5}{c}{\textbf{1980-1994: Foundations and Growth of Tactile Robotics (Sections~\ref{sec:beginnings} and~\ref{sec:foundations})}}\\
        \hline
		1980 & Touch-sensing Technology: A Review & Book/report & 85 (0) & \cite{harmon_touch-sensing_1980} \\
		1981 & A Sense of Touch Begins to Gather Momentum & \textit{Sensor Review} & 19 (0) & \cite{harmon_sense_1981} \\
		1982 & Automated Tactile Sensing & \textit{IJRR} & 524 (7) & \cite{harmon_automated_1982} \\
		1984 & Tactile Sensing for Robots & Book chapter & 85 (0) & \cite{harmon_tactile_1984} \\
		\hline 
		1985 & Tactile Sensors and the Gripping Challenge & \textit{IEEE Spectrum} & 318 (3) & \cite{dario_tactile_1985} \\
		1986 & Robotic Tactile Sensing & \textit{Byte Magazine} & 147 (0) & \cite{pennywitt_robotic_1986} \\
		1986 & An Overview of Tactile Sensing & Book/report & 80 (0) & \cite{agrawal_overview_1986} \\
		1988 & Force and Tactile Sensing for Robots & Book chapter & 22 (0) & \cite{dario_force_1988} \\
		1989 & Tactile Sensing in Robotics & \textit{J. Physics E} & 43 (0) & \cite{jayawant_tactile_1989} \\
		1989 & A Survey of Robot Tactile Sensing Technology & \textit{IJRR} & {526 (4)} & \cite{nicholls_survey_1989} \\
		\hline 
        1990 & Robot Tactile Sensing & Book &  175 (0) & \cite{russell_robot_1990} \\
        1991 & Tactile Sensing: Technology and Applications & \textit{Sens. Actuat. A-Phys} & 117 (1) & \cite{dario_tactile_1991} \\
        1991 & Artificial Tactile Sensing and Haptic Perception & \textit{Meas. Sci. \& Tech.} & 57 (0) & \cite{de_rossi_artificial_1991} \\
        1992 & Advanced Tactile Sensing for Robotics & Book &  89 (0) & \cite{nicholls_advanced_1992} \\
        1992 & Lessons From the Study of Biological Touch for Robotic Tactile Sensing & Book chapter &  54 (0) & \cite{lederman_lessons_1992} \\
        1992 & Touch Sensing for Robotic Manipulation and Recognition & Book chapter & 76 (0) & \cite{howe_touch_1992} \\
		1993 & Tactile Sensing and Control of Robotic Manipulation & \textit{Adv. Robotics} & 423 (15) & \cite{howe_tactile_1993-1} \\
		\hline %17
        \multicolumn{5}{c}{\textbf{1995-2009: Tactile Robotics Winter (Section~\ref{sec:pessimism})}}\\
        \hline
		1999 & Tactile Sensing for Mechatronics -- A State of the Art Survey & \textit{Mechatronics} & 983 (18) & \cite{lee_review_1999} \\
  	\hline
		2000 & Tactile Sensing: New Directions, New Challenges & \textit{IJRR} & 288 (8) & \cite{lee_tactile_2000} \\
        2003 & Tactile Sensing Technology for Minimal Access Surgery -- A Review & \textit{Mechatronics} & 289 (6) & \cite{eltaib_tactile_2003} \\
        2004 & Human Tactile Perception as a Standard for Artificial Tactile Sensing -- A Review & \textit{Int. J. Medical Rob.} & 465 (54) & \cite{dargahi_human_2004} \\
		\hline
		2005 & Tactile Sensing in Intelligent Robotic Manipulation -- A Review & \textit{Industrial Robot} & 377 (14) & \cite{tegin_tactile_2005} \\
        2005 & Advances in Tactile Sensors Design/Manufacturing and its Impact on Robotics Applications... & \textit{Industrial Robot} & 162 (4) & \cite{dargahi_advances_2005} \\
		2008 & Force and Tactile Sensing & Book chapter & 213 (5) & \cite{cutkosky_force_2008} \\
    	\hline %7
        \multicolumn{5}{c}{\textbf{2010-2024: Expansion and Diversification (Section~\ref{sec:renewal})}}\\
        \hline 
		{\bf 2010} & {\bf Tactile Sensing -- From Humans to Humanoids} & {\bf \textit{IEEE Trans. Rob.}} & {\bf 1991 (196)} & \cite{dahiya_tactile_2010} \\
		2010 & A Survey of Tactile Human–Robot Interactions & \textit{Rob. Aut. Sys.} & 461 (21) & \cite{argall_survey_2010} \\
		2011 & Tactile Sensing for Dexterous In-hand Manipulation in Robotics -- A Review & \textit{Sens. Actuat. A-Phys} & {934 (70)} & \cite{yousef_tactile_2011} \\
		2012 & A Review of Tactile Sensing Technologies with Applications in Biomedical Engineering & \textit{Sens. Actuat. A-Phys} & {887 (53)} & \cite{tiwana_review_2012} \\
		% 2013 & Directions Toward Effective Utilization of Tactile Skin: A Review & \textit{IEEE Sensors} & 497 (31) & \cite{dahiya_directions_2013} \\
  	2013 & {Robotic Tactile Sensing: Technologies \& System} & Book & {591 (40)} & \cite{dahiya_robotic_2013} \\
		2013 & Artificial Sense of Slip -- A Review & \textit{IEEE Sensors} & 127 (12) & \cite{francomano_artificial_2013} \\
		2013 & Tactile Sensors for Robotic Applications & \textit{Measurement} & 260 (27) & \cite{girao_tactile_2013} \\
		% {\bf 2013} & {\bf The Evolution of E-Skin: A Brief History, Design Considerations, and Recent Progress} & {\bf \textit{Adv. Mater.}} & {\bf 2556 (215)} & \cite{hammock_25th_2013} \\
		2013 & Synthetic and Bio-Artificial Tactile Sensing: A Review & \textit{Sensors} & 192 (12) & \cite{lucarotti_synthetic_2013} \\
		2014 & ... Tactile Sensing for Palpation in Robot-Assisted Minimally Invasive Surgery: A Review & \textit{IEEE Sensors} & 166 (14) & \cite{konstantinova_implementation_2014} \\
    	\hline %9
		{\bf 2015} & {\bf Tactile Sensing in Dexterous Robot Hands -- Review} & \textit{\bf Rob. Aut. Sys.} & {\bf 858 (101)} & \cite{kappassov_tactile_2015-1} \\
		2015 & Sensors for Robotic Hands: A Survey of State of the Art & \textit{IEEE Access} & 123 (11) & \cite{saudabayev_sensors_2015} \\
		2017 & Recent Progress on Tactile Object Recognition & \textit{Int J. Adv. Rob. Sys.} & 84 (13) & \cite{liu_recent_2017} \\
		2017 & Robotic Tactile Perception of Object Properties: A Review & \textit{Mechatronics} & 453 (90) & \cite{luo_robotic_2017} \\
		{\bf 2017} & {\bf GelSight: High-Resolution Robot Tactile Sensors for Estimating Geometry and Force} & {\bf \textit{Sensors}} & {\bf 1034 (357)} & \cite{yuan_gelsight_2017-1} \\
		{\bf 2018} & \ {\bf The TacTip Family: Soft Optical Tactile Sensors with 3D-Printed Biomimetic Morphologies} & {\bf \textit{Soft Robotics}} & {\bf 524 (131)} & \cite{ward-cherrier_tactip_2018-2} \\
        {2018} & Morphological Computation in Haptic Sensation and Interaction: From Nature to Robotics & \textit{Adv. Robotics} & {33 (6)} & \cite{bernth_morphological_2018} \\
        2018 & Tactile Sensors for Friction Estimation and Incipient Slip Detection...: A Review & \textit{IEEE Sensors} & 221 (36) & \cite{chen_tactile_2018} \\
		2018 & Biomimetic Tactile Sensors and Signal Processing with Spike Trains: A Review & \textit{Sens. Actuat. A-Phys} & 64 (11) & \cite{yi_biomimetic_2018} \\
        % {\bf 2018} & {\bf Toward Perceptive Soft Robots: Progress and Challenges} & {\bf\textit{Adv. Sci.}} & {\bf 690 (155)} & \cite{wang_toward_2018} \\
        2019 & Tactile Robots as a Central Embodiment of the Tactile Internet & \textit{Procs IEEE} & 94 (16) & \cite{haddadin_tactile_2019} \\
        2019 & Tactile Image Sensors Employing Camera: A Review & \textit{Sensors} & 172 (43) & \cite{shimonomura_tactile_2019} \\
		2019 & Recent progress in Tactile Sensing...: Can We Turn Tactile Sensing into Vision? & \textit{Adv. Robotics} & 140 (38) & \cite{yamaguchi_recent_2019} \\
    	\hline % 12
		%3
        2020 & Visuotactile Sensors With Emphasis on GelSight Sensor: A Review & \textit{IEEE Sensors} & 125 (47) & \cite{abad_visuotactile_2020} \\
		2020 & A Review of Tactile Information: Perception and Action Through Touch & \textit{IEEE Trans. Rob.} & 259 (86) & \cite{li_review_2020} \\
		2020 & Embodied Tactile Perception and Learning & \textit{Brain Science Adv.} & 18 (8) & \cite{liu_embodied_2020} \\
		%4
        2021 & Soft Biomimetic Optical Tactile Sensing with the TacTip: A Review & \textit{IEEE Sensors} & 132 (61) & \cite{lepora_soft_2021-2} \\
		2021 & On the Design and Development of Vision-based Tactile Sensors & \textit{J. Intell. Rob. Sys.} & 48 (27) & \cite{shah_design_2021} \\
		2021 & Soft Robotic Hands and Tactile Sensors for Underwater Robotics & \textit{Applied Mechanics} & 38 (14) & \cite{subad_soft_2021} \\
		2021 & Three-dimensional Printing of Tactile Sensors for Soft Robotics & \textit{MRS Bulletin} & 15 (4) & \cite{zhou_three-dimensional_2021} \\  
		%6
        2022 & Tactile and Vision Perception for Intelligent Humanoids & \textit{Adv. Intell. Sys.} & 33 (14) & \cite{gao_tactile_2022} \\
        2022 & Review of Bioinspired Vision-Tactile Fusion Perception (VTFP): From Humans to Humanoids & \textit{IEEE Trans. Med. Robot.} & 14 (8) & \cite{he_review_2022} \\ % Bionics
		2022 & Recent Progress of Biomimetic Tactile Sensing Technology Based on Magnetic Sensors & \textit{Biosensors} & 32 (23) & \cite{man_recent_2022} \\
		2022 & Tactile Sensing for Minimally Invasive Surgery... & \textit{Front. Robot. AI} & 66 (35) & \cite{othman_tactile_2022} \\
		2022 & A Review on Sensory Perception for Dexterous Robotic Manipulation & \textit{Int J. Adv. Rob. Sys.} & 41 (21) & \cite{xia_review_2022} \\
		2022 & Hardware Technology of Vision-Based Tactile Sensor: A Review & \textit{IEEE Sensors} & 102 (74) & \cite{zhang_hardware_2022} \\
		%5
        2023 & Machine Learning for Tactile Perception: Advancements, Challenges, and Opportunities & \textit{Adv. Intell. Syst.} & 23 (17) & \cite{hu_machine_2023} \\
		2023 & Marker Displacement Method Used in Vision-Based Tactile Sensors -- From 2-D to 3-D... & \textit{IEEE Sensors} & 30 (26) & \cite{li_marker_2023} \\
		2023 & Tactile-Sensing Technologies: Trends, Challenges and Outlook in Agri-Food Manipulation & \textit{Sensors} & 38 (34) & \cite{mandil_tactile-sensing_2023} \\
		{\bf 2023} & {\bf Recent Progress in Advanced Tactile Sensing Technologies for Soft Grippers} & {\bf \textit{Adv. Func. Mater.}} & {\bf 111 (108)} & \cite{qu_recent_2023} \\
		2023 & Recent Progress of Tactile and Force Sensors for Human–Machine Interaction & \textit{Sensors} & 38 (20) & \cite{xu_recent_2023} \\
        %6
		2024 & The Future Lies in a Pair of Tactile Hands & \textit{Science Robotics} & -- & \cite{lepora_future_2024-1} \\
		2024 & When Vision Meets Touch: A Contemporary Review for Visuotactile Sensors... & \textit{IEEE J-STSP} & -- & \cite{li_when_2024} \\
		2024 & A Comprehensive Review of Robot Intelligent Grasping Based on Tactile Perception & \textit{Robot. \& CIM} & -- &\cite{li_comprehensive_2024} \\
		2024 & Towards High Performance and Durable Soft Tactile Actuators & \textit{Chem. Soc. Rev.} & -- & \cite{tan_towards_2024} \\
		2024 & Recent Progress of Optical Tactile Sensors: A Review & \textit{Opt. Laser Technol.} & -- & \cite{yao_recent_2024-1} \\
		2024 & Optical Micro/Nanofiber Enabled Tactile Sensors and Soft Actuators: A Review & \textit{OES} & -- & \cite{zhang_optical_2024} \\
        \hline %24
	\end{tabular}}
    \vspace{1em}
	\caption{Review articles on tactile sensing and tactile robotics. Citations counts from papers published in 2024 are not given, to give reliable counts at the time of writing. Pre-2024 papers with $<$10 citations have been omitted as they have not impacted the field. Very long titles are cropped. Articles were identified manually using Google Scholar with combinations of search terms such as `tactile', `robotics', `review' and `survey', and also by checking the bibliographies of other reviews. Those articles that are currently being cited most ($>$100 citations in 2024) are emphasized in bold, comprising 6 articles from 74 total. Several trends are visible in this Table. Influential pre-2010 articles now seem mostly forgotten, particularly the foundational articles in the 1980s and early 1990s. Also, as observed in the introduction, some generational changes in behavior are apparent: steady growth in the number of tactile reviews from 1980 to the mid-1990s, then a drop in reviewing for about 15 years, and accelerating growth since. }
    \vspace{0em}% Some other trends and comments on this methodology of cataloging review articles are given in Appendix~\ref{sec:comments}.
	\label{tab:1}
\end{table*}

\begin{table*}[b!]
\resizebox{\textwidth}{!}{%
	\renewcommand{\arraystretch}{1}
	\centering
	\begin{tabular}{@{}cc@{}c@{}c@{}c@{}}	
		\multirow{2}*{\textbf{Year}} & \textbf{Title} & \textbf{Journal} & \textbf{Citations} & \multirow{2}*{\textbf{Reference}} \\
		& \textbf{(cropped when necessary)} & \textbf{(abbreviated)}  & \textbf{to 2024 (in 2024)} &  \\
    	\hline % 5
		{\bf 2010} & {\bf Tactile Sensing -- From Humans to Humanoids} & {\bf \textit{IEEE Trans. Rob.}} & {\bf 1991 (196)} & \cite{dahiya_tactile_2010} \\
    	{2013} & Directions {{Toward Effective Utilization}} of {{Tactile Skin}}: {{A Review}} & {\textit{IEEE       
        Sensors}} & {499 (31)} & \cite{dahiya_directions_2013} \\
  	{2013} & Robotic {{Tactile Sensing}}: Technologies and System & Book & {590 (40)} & \cite{dahiya_robotic_2013} \\
		{\bf 2013} & {\bf The Evolution of E-Skin: A Brief History, Design Considerations, and Recent Progress} & {\bf \textit{Adv. Mater.}} & {\bf 2539 (215)} & \cite{hammock_25th_2013} \\
        {2014} & Flexible {{Tactile Sensing Based}} on {{Piezoresistive Composites}}: {{A Review}} &        
        {\textit{Sensors}} & {506 (43)} & \cite{stassi_flexible_2014} \\
    	\hline % 14
        {\bf 2015} & {\bf Technologies for {{Printing Sensors}} and {{Electronics Over Large Flexible Substrates}}: {{A Review}}} & {\bf \textit{IEEE Sensors}} & {\bf 1407 (113)} & \cite{khan_technologies_2015} \\
        {2015} & Artificial Skin and Tactile Sensing for Socially Interactive Robots: {{A}} Review & {\textit{Rob. Aut. Systems}} & {276 (31)} & \cite{silvera-tawil_artificial_2015} \\
        {\bf 2015} & {\bf Recent {{Progress}} in {{Electronic Skin}}} & {\bf \textit{Adv. Sci.}} & {\bf 974 (102)} & \cite{wang_recent_2015} \\
		2016 & Robots with a Sense of Touch & \textit{Nat. Mater.} & 302 (37) & \cite{bartolozzi_robots_2016} \\
        {\bf 2016} & {\bf Pursuing Prosthetic Electronic Skin} & {\bf \textit{Nat. Mater.}} & {\bf 2268 (323)} & \cite{chortos_pursuing_2016} \\
        {2017} & Recent Progresses on Flexible Tactile Sensors & {\textit{Mater. Today Phys.}} & {304 (39)} & \cite{wan_recent_2017} \\
        {2017} & Novel {{Tactile Sensor Technology}} and {{Smart Tactile Sensing Systems}}: {{A Review}} & {\textit{Sensors}} & {293 (41)} & \cite{zou_novel_2017} \\
        {2018} & Recent {{Advances}} in {{Tactile Sensing Technology}} & {\textit{Micromachines}} & {97 (13)} & \cite{park_recent_2018} \\
        {2019} & {E-{{Skin}}: {{From Humanoids}} to {{Humans}} [{{Point}} of {{View}}]} & {\textit{Procs of the IEEE}} & {185 (22)} & \cite{dahiya_e-skin_2019} \\
        {2019} & {Large-{{Area Soft}} e-{{Skin}}: {{The Challenges Beyond Sensor Designs}}} & {\textit{Procs of the IEEE}} & {300 (62)} & \cite{dahiya_large-area_2019} \\
        {2019} & {A {{Review}} of {{Printable Flexible}} and {{Stretchable Tactile Sensors}}} & {\textit{Research}} & {172 (28)} & \cite{senthil_kumar_review_2019} \\
        {2019} & {Self-{{Powered Tactile Sensor Array Systems Based}} on the {{Triboelectric Effect}}} & {\textit{Adv. Func. Mater.}} & {175 (35)} & \cite{tao_selfpowered_2019} \\
        {2019} & {Tactile {{Sensors}} for {{Advanced Intelligent Systems}}} & {\textit{Adv. Intell. Syst.}} & {125 (37)} & \cite{wang_tactile_2019} \\
        {\bf 2019} & {\bf Electronic {{Skin}}: ... {{Skin}}-{{Attachable Devices}} for {{Health Monitoring}}, {{Robotics}}, and {{Prosthetics}}} & {\bf\textit{Adv. Mater.}} & {\bf 1451 (362)} & \cite{yang_electronic_2019} \\
    	\hline % 35
        {2020} & {A {{Survey}} of {{Tactile-Sensing Systems}} and {{Their Applications}} in {{Biomedical Engineering}}} & {\textit{Adv. Mat. Sci. Eng.}} & {75 (16)} & \cite{al-handarish_survey_2020} \\
        {2020} & {Ionic {{Tactile Sensors}} for {{Emerging Human}}-{{Interactive Technologies}}: {{A Review}} of {{Recent Progress}}} & {\textit{Adv. Func. Mater.}} & {168 (43)} & \cite{amoli_ionic_2020} \\
        {\bf 2020} & {\bf ...High-Performance Piezoresistive and Capacitive Flexible Pressure Sensors: {{A}} Review} & {\bf\textit{J. Mat. Sci. Tech.}} & {\bf 326 (110)} & \cite{chen_progress_2020} \\
        {2020} & {Biomimetic {{Tactile Sensors Based}} on {{Nanomaterials}}} & {\textit{ACS Nano}} & {68 (16)} & \cite{lee_biomimetic_2020} \\
        {2020} & {Recent Progress in Tactile Sensors and Their Applications in Intelligent Systems} & {\textit{Science Bulletin}} & {179 (36)} & \cite{liu_recent_2020} \\
        {\bf 2020} & {\bf Electronic Skins and Machine Learning for Intelligent Soft Robots} & {\bf\textit{Science Robotics}} & {\bf 536 (174)} & \cite{shih_electronic_2020} \\
        {2020} & {Soft {{eSkin}}: Distributed Touch Sensing with Harmonized Energy and Computing} & {\textit{Phil. Trans Roy. Soc. A}} & {93 (9)} & \cite{soni_soft_2020} \\
        {2020} & {...{{From}} Machine Learning-Enhanced Tactile Sensing to Neuromorphic Sensory Systems} & {\textit{App. Physics Rev.}} & {240 (51)} & \cite{zhu_technologies_2020} \\
        {\bf 2021} & {\bf ... {{Flexible Tactile Sensors}} for {{Human}}-{{Interactive Systems}}: {{From Sensors}} to {{Advanced Applications}}} & {\bf\textit{Adv. Mater.}} & {\bf 446 (216)} & \cite{pyo_recent_2021} \\
        {2022} & Recent Advances in Touch Sensors for Flexible Wearable Devices & {\textit{Sensors}} & {74 (29)} & \cite{anwer_recent_2022} \\
        {2022} & {An {{Atlas}} for the {{Inkjet Printing}} of {{Large-Area Tactile Sensors}}} & {\textit{Sensors}} & {13 (5)} & \cite{baldini_atlas_2022} \\
        {2022} & {{{MEMS-Based Tactile Sensors}}: {{Materials}}, {{Processes}} and {{Applications}} in {{Robotics}}} & {\textit{Micromachines}} & {28 (16)} & \cite{bayer_mems-based_2022} \\
        {2022} & Conducting Polymers for the Design of Tactile Sensors & {\textit{Polymers}} & {33 (19)} & \cite{bubniene_conducting_2022} \\
        {2022} & Flexible Tactile Sensors based on Silver Nanowires: Material Synthesis, Microstructuring, Assembly... & {\textit{Emergent Mater.}} & {14 (6)} & \cite{cuasay_flexible_2022} \\
        {2022} & Recent Advances in Resistive Sensor Technology for Tactile Perception: A Review & {\textit{IEEE Sensors}} & {30 (18)} & \cite{zhu_recent_2022} \\
        {2023} & ...Flexible Tactile Sensors in Robotic Applications on Objects Properties Recognition, Manipulation... & {\textit{Soft Science}} & {16 (14)} & \cite{jin_progress_2023} \\
        {2023} & Emerging MXene-Based Flexible Tactile Sensors for Health Monitoring and Haptic Perception & {\textit{Small}} & {105 (88)} & \cite{lai_emerging_2023} \\
        % {2023} & Emerging Functional Polymer Composites for Tactile Sensing & {\textit{Materials}} & {4 (4)} & \cite{lian_emerging_2023} \\
        % {2023} & Progress and Prospects in Flexible Tactile Sensors & {\textit{Front. Bioeng. Biotechnol.}} & {7 (7)} & \cite{liu_progress_2023} \\
        {2023} & PVDF-Based Flexible Piezoelectric Tactile Sensors: Review & {\textit{Cryst. Res. Technol.}} & {21 (20)} & \cite{qi_pvdf-based_2023} \\
        {2023} & Self-powered Ionic Tactile Sensors & {\textit{J. Mater. Chem. C}} & {13 (7)} & \cite{saha_self-powered_2023} \\
        {2023} & Tactile Sensing Technology in Bionic skin: A Review & {\textit{Biosens. Bioelectron.}} & {67 (49)} & \cite{wang_tactile_2023} \\
        {2023} & Field Effect Transistor-based Tactile sensors: From Sensor Configurations to Advanced Applications &  {\textit{InfoMat}} & {49 (29)} & \cite{wang_field_2023} \\
        {2023} & Recent Progress in High-resolution Tactile Sensor Array: From Sensor Fabrication to Advanced Applications & {\textit{Prog. Nat. Sci. Mat. Int.}} & {18 (14)} & \cite{wang_recent_2023} \\
        % {2023} & Review: Progress on 3D-Printing Technology in the Preparation of Flexible Tactile Sensors & {\textit{J. Mater. Sci.}} & { ()} & \cite{xu_review_2023} \\
        {2023} & Recent Advances in Flexible Piezoresistive Arrays: Materials, Design, and Applications & {\textit{Polymers}} & {30 (26)} &  \cite{xu_recent_2023-1} \\
        {2023} & Advanced Polymer Materials-based Electronic Skins for Tactile and Non-contact Sensing Applications &  {\textit{InfoMat}} & {23 (23)} & \cite{yin_advanced_2023} \\
        {2023} & Flexible Tactile Sensors with Biomimetic Microstructures: Mechanisms, Fabrication, and Applications & {\textit{Adv. Colloid Interface Sci.}} & {17 (17)} & \cite{zhang_flexible_2023} \\
        {2023} & Recent Development of Self-Powered Tactile Sensors Based on Ionic Hydrogels & {\textit{Gels}} & 14 (7) & \cite{zhao_recent_2023} \\
		2024 & The role of bio-inspired micro-/nano-structures in flexible tactile sensors & \textit{J. Mater. Chem. C} & -- & \cite{fu_role_2024} \\
		2024 & A review on graphene-based sensors for tactile applications & \textit{Sensor Actuat. A-Phys.} & -- & \cite{he_review_2024} \\
		2024 & Recent advances in multimodal sensing integration and decoupling strategies for tactile perception & \textit{Materials Futures} & -- & \cite{kong_recent_2024} \\
		2024 & Recent Advances in Self-Powered Tactile Sensing for Wearable Electronics & \textit{Materials} & -- & \cite{liu_recent_2024} \\
		2024 & Tactile sensors: A review & \textit{Measurement} & -- & \cite{meribout_tactile_2024} \\
		2024 & Recent advances in triboelectric tactile sensors for robot hand & \textit{Mater. Today Phys.} & -- & \cite{noor_recent_2024} \\
		2024 & Functional Tactile Sensor Based on Arrayed Triboelectric Nanogenerators & \textit{Adv. Energy Mater.} & -- & \cite{peng_functional_2024} \\
		2024 & Recent developments in stretchable and flexible tactile sensors towards piezoresistive systems: A review & \textit{Polym. Adv. Technol.} & -- & \cite{saxena_recent_2024} \\
		2024 & Flexible resistive tactile pressure sensors & \textit{J. Mater. Chem. A} & -- & \cite{shu_flexible_2024} \\
        \hline
	\end{tabular}}
    \vspace{1em}
	\caption{Review articles on e-skins. Citations counts from papers published in 2024 are not given, to give reliable counts at the time of writing. Pre-2024 papers with $<$10 citations have been omitted as they have not impacted the field. Very long titles are cropped. Articles were identified manually using Google Scholar. Those articles that are currently being cited most ($>$100 citations in 2024) are emphasized in bold, comprising 8 articles from 54 total.}
	\label{tab:5}
\vspace{3em}
\end{table*}

\begin{table*}[b!]
\resizebox{\textwidth}{!}{%
	\renewcommand{\arraystretch}{1}
	\centering
	\begin{tabular}{@{}cc@{}c@{}c@{}c@{}}	
		\multirow{2}*{\textbf{Year}} & \textbf{Title} & \textbf{Journal} & \textbf{Citations} & \multirow{2}*{\textbf{Reference}} \\
		\textbf{} & \textbf{(cropped when necessary)} & \textbf{(abbreviated)}  & \textbf{to 2024 (in 2024)} & \\
		\hline
        2014 & The tactile Internet: Applications and Challenges & \textit{IEEE Veh. Tech. Mag.} & 1253 (50) & \cite{fettweis_tactile_2014} \\
        \hline
        2016 & The tactile Internet:  Vision, Recent Progress, and Open Challenges & \textit{IEEE Comms Mag.} & 383 (21) & \cite{maier_tactile_2016} \\
        2016 & 5G-Enabled tactile Internet & \textit{IEEE J. Sel. Areas Comms} & 1022 (59) & \cite{simsek_5g-enabled_2016} \\
        2017 & Realizing the tactile Internet: Haptic Communications over Next Generation 5G Cellular Networks & \textit{IEEE Wireless Comms} & 394 (12) & \cite{aijaz_realizing_2017} \\
        2017 & Internet of skills, where robotics meets AI, 5G and the tactile Internet & \textit{EuCNC} & 168 (6) & \cite{dohler_internet_2017} \\
        2017 & Challenges in Haptic Communications Over the tactile Internet & \textit{IEEE Access} & 145 (10) & \cite{van_den_berg_challenges_2017} \\
        2018 & Toward Haptic Communications Over the 5G tactile Internet & \textit{IEEE Comms Surveys} & 276 (29) & \cite{antonakoglou_toward_2018} \\
        2019 & The tactile Internet for Industries: A Review & \textit{Procs IEEE} & 184 (21) & \cite{aijaz_tactile_2019} \\
		2019 & Tactile Internet and its Applications in 5G Era: A Comprehensive Review & \textit{Int. J. Commun. Syst.} & 158 (16) & \cite{gupta_tactile_2019} \\     	
        2019 & Tactile Robots as a Central Embodiment of the tactile Internet & \textit{Procs IEEE} & 94 (16) & \cite{haddadin_tactile_2019} \\
        \hline
        2020 & Toward tactile Internet in Beyond 5G Era: Recent Advances, Current Issues, and Future Directions & \textit{IEEE Access} & 203 (36) & \cite{sharma_toward_2020} \\
  	2021 & A Comprehensive Survey of the tactile Internet: State-of-the-Art and Research Directions & \textit{IEEE Commun. Tut.} & 137 (42) & \cite{promwongsa_comprehensive_2021} \\
		2022 & TDM-PON-Based Optical Access Network for tactile Internet, 5G, and Beyond & \textit{IEEE Network} & 31 (12) & \cite{chung_tdm-pon-based_2022} \\        
		% 2022 & A Review on Tactile IoT: Architecture, Requirements, Prospects, and Future Directions & \textit{Trans Emerging Tel Tech} & 6 (2) & \cite{ray_review_2022} \\
		2023 & Fog Computing for 5G-Enabled tactile Internet: Research Issues, Challenges, and Future Research Directions & \textit{Mobile Netw. Appl.} & 46 (6) & \cite{aggarwal_fog_2023} \\
		% 2023 & Tactile IoT and 5G \& Beyond Schemes as Key Enabling Technologies for the Future Metaverse & \textit{Telecommun. Syst.} & 6 (5) & \cite{tychola_tactile_2023} \\
        % \hline
	\end{tabular}}
    \vspace{1em}
	\caption{Review articles on the tactile Internet. Papers with $<$10 citations have been omitted as they have not impacted the field. Very long titles are cropped. Articles were identified manually using Google Scholar. Total of 14 articles.}
	\label{tab:7}
    \vspace{2em}
\end{table*}

\clearpage
\vfill
\pagebreak

\bibliographystyle{SageH}
\bibliography{Tactile_robotics,Telerobotics,Robot_manipulation,History_of_robotics,Tactile_robotics_supplemental,Artificial_intelligence,Tactile_eskins,Robot_learning
,Soft_robotics,Tactile_internet,Robot_hands}

\end{document}